 \documentclass[final,3p,times]{elsarticle}

\usepackage{lineno,hyperref}
\usepackage{graphics}
\usepackage{graphicx}
\usepackage{epsfig}
\usepackage{subfigure}
\usepackage{amsfonts}
\usepackage{mathrsfs}
\usepackage{epstopdf}
\usepackage{amssymb}
\usepackage{rotating}
\usepackage{lscape}
\usepackage{natbib}
\usepackage{amsmath}
\usepackage{amsthm}

\usepackage{algorithm}
\usepackage{algorithmicx}
\usepackage{algpseudocode}

\usepackage{nicefrac}       
\usepackage{microtype}      
\usepackage{color}
\usepackage{multirow}
\usepackage{booktabs}       

\floatname{algorithm}{Algorithm}

\newtheorem{definition}{Definition}

\usepackage{caption2}

\begin{document}

\begin{frontmatter}

\title{Logic could be learned from images}

\author[mymainaddress0,mymainaddress2]{Qian Guo} \ead{czguoqian@163.com}\vskip 2mm
\author[mymainaddress0,mymainaddress1,mymainaddress2]{Yuhua Qian \footnotemark[1]}\ead{jinchengqyh@126.com}\vskip 2mm
\author[mymainaddress0,mymainaddress2]{Xinyan Liang}\ead{liangxinyan48@163.com}\vskip 2mm
\author[mymainaddress3]{Yanhong She}\ead{yanhongshe@gmail.com,yanhongshe@xsyu.edu.cn}\vskip 2mm
\author[mymainaddress1,mymainaddress2]{Deyu Li}\ead{lidy@sxu.edu.cn}\vskip 2mm
\author[mymainaddress1,mymainaddress2]{Jiye Liang}\ead{ljy@sxu.edu.cn}\vskip 2mm

\renewcommand{\thefootnote}{\fnsymbol{footnote}}
\footnotetext[1]{Corresponding author.}
\address[mymainaddress0]{Institute of Big Data Science and Industry, Shanxi University, Taiyuan 030006, Shanxi, China}
\address[mymainaddress1]{Key Laboratory of Computational Intelligence and Chinese Information Processing of Ministry of Education, Shanxi University, Taiyuan 030006, Shanxi, China}
\address[mymainaddress2]{School of Computer and Information Technology, Shanxi University, Taiyuan 030006, Shanxi, China}
\address[mymainaddress3]{College of Science, Xi'an Shiyou University, Xi'an 710065, Shaan'xi, China }

\begin{abstract}
  Logic reasoning is a significant ability of human intelligence
  and also an important task in artificial intelligence.
  The existing logic reasoning methods, quite often,
  need to design some reasoning patterns beforehand.
  This has led to an interesting question:
  can logic reasoning patterns be directly learned from given data?
  The problem is termed as a data concept logic.
  In this study, a learning logic task from images,
  called a LiLi task, first is proposed.
  This task is to learn and reason the logic relation from images,
  without presetting any reasoning patterns.
  As a preliminary exploration,
  we design six LiLi data sets (Bitwise And, Bitwise Or, Bitwise Xor, Addition,
  Subtraction and Multiplication),
  in which each image is embedded with a n-digit number.
  It is worth noting that a learning model beforehand does not know
  the meaning of the n-digit numbers embedded in images
  and the relation between the input images and the output image.
  In order to tackle the task, in this work
  we use many typical neural network models and produce fruitful results.
  However, these models have the poor performances on the difficult logic task.
  For furthermore addressing this task,
  a novel network framework called a divide and conquer model
  by adding some label information is designed,
  achieving a high testing accuracy.
\end{abstract}


\end{frontmatter}


\section{Introduction}
Human intelligence integrates cognitive functions
such as perception, learning, memory, problem solving
and logic reasoning \citep{colom2010human}.
Among them, logic reasoning is a significant ability of human intelligence.
Applying the reasoning, humans obtain some rules hidden in complex phenomenon,
and even forecast the unknown events.
One of the goals of artificial intelligence is to mimic human cognitive functions to the utmost.
As a part of cognitive functions,
logic reasoning is also an important task
in artificial intelligence \cite{Johnson2016CLEVR}.

Many logic reasoning methods such as
fuzzy reasoning \cite{Wang1997Fuzzy, Mizumoto1982Comparison, Yen1999Fuzzy, Pei2004On},
FCA \cite{wille1982restructuring, Tadrat2012A, Golinskapilarek2007Relational, Shao2020The},
probabilistic reasoning \cite{nilsson1986probabilistic,
nilsson1993probabilistic, She2018A, Li2020A},
evidential reasoning \cite{pearl1987evidential, chen2016fuzzy},
Bayesian reasoning \cite{yang2008fuzzy, tenenbaum2006theory} and
rough reasoning \cite{Qian2018Local, sheroughlogic, Lin2019Granular, Li2019Double},
have been proposed.
However, quite often, these methods need to
design some reasoning patterns beforehand.
For example, in the FCA, one first obtains a formal context
applying the domain expert knowledge,
then computes the concept lattice from the formal context,
and finally achieves knowledge reasoning
using the disjunction and conjunction operations.
This process not only costs a large amount of time,
but also heavily depends on the domain expert experience.
But, without mastering special domain knowledge beforehand,
human still can directly reason from given data.
For example, without mastering knowledge of 3D reconstruction beforehand,
people can reconstruct 3D model of an unseen 2D image in his mind
through observing and reasoning many 2D images and corresponding 3D scenes in real world.
This has led to an interesting research
topic: can machine directly learn logic reasoning patterns from
given data? And these logical patterns are termed as the data concept logic (DCL).

\begin{figure*}[t]
\begin{center}
    \includegraphics[width=0.9\textwidth]{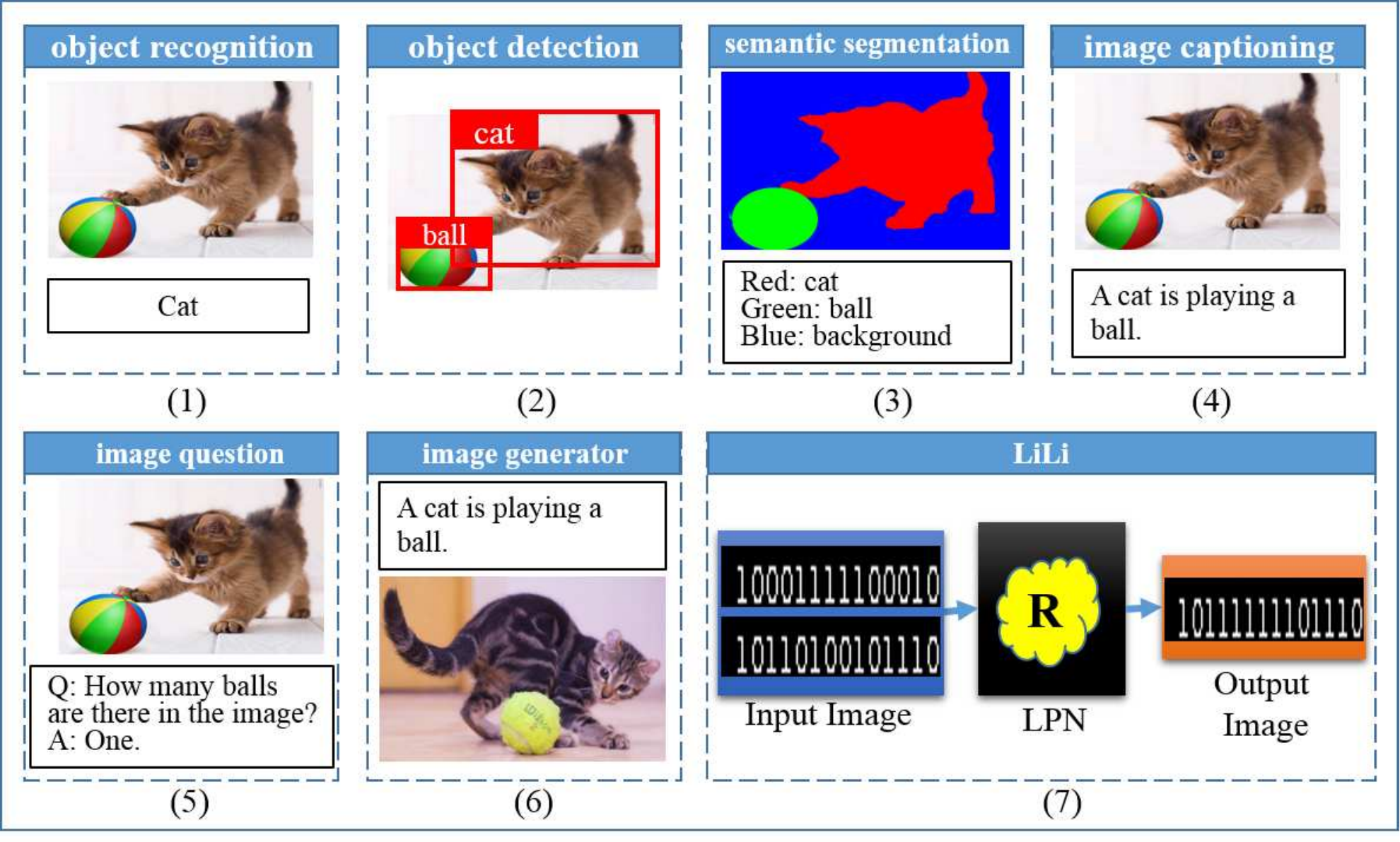}
\end{center}
    \caption{ The differences among these popular computer vision tasks
    (1) Object recognition (sometimes object classification) is to classify individual objects.
    (2) Object detection is to classify individual objects and localize each using a bounding box.
    (3) Semantic segmentation is to classify each pixel into a fixed set of categories without differentiating object instances.
    (4) Image captioning is to describe the content of an image by using reasonably formed natural sentences.
    (5) Visual question answering (VQA) is to automatically answer natural language questions according to related the image content.
    (6) Image generator is to generate images according to images or text description.
    (7) Data concept logic is to learn to obtain logic concepts from a given data set.}
\label{Fig:1}
\end{figure*}

As a preliminary exploration, in this study,
we design a task of the DCL
which is called learning logic task from images,
just a LiLi task shown in Fig. \ref{Fig:1}(7).
Unlike the logical operation defined by human (LOH)
using some domain expert knowledge,
a LiLi task is to learn and reason the relation
between two input images and one output image
without any reasoning patterns beforehand,
i.e. LiLi does not know any reasoning patterns about $R$.
In summary, there are some differences below between a LiLi task
and a LOH.

\begin{itemize}
  \item[$\bullet$] For LiLi, one does not know
  any reasoning patterns about $R$ except for giving a data set,
  while for LOH whose focus is that how to define a reasonable logical operation,
  one always possesses lots of domain knowledge about $R$.
  \item[$\bullet$] LPN induced by a LiLi  models an abstract or low level logical relation
  in term of the pixel values.
  However, the existing logical operation
  models a semantic or high level logical relation
  in term of the numbers or symbols.
  \item[$\bullet$] LPN induced by a LiLi is a data-driven method to model the logical relation,
  while LOH is an expert-driven method.
\end{itemize}

Learning logic task from images (LiLi task)
is also a very important computer vision task.
Unfortunately, to the best of our knowledge,
there are only a bit of work on the LiLi task
shown in Fig. \ref{Fig:1}(7).
\cite{Guo2019Mining} mined the logical patterns from Fashion-Logic data sets in a data-driven way.
Zhou et al. \cite{Dai2019Bridging} proposed abductive learning framework
which can learn perception and reasoning modules concurrently.
In contrast, a variety of models based on deep convolutional neural networks (CNNs)
have achieved the state-of-the-art performances,
even super-human on some tasks
for the common computer vision tasks such as
object recognition \cite{huang2016densely, He2016Deep, Liang2021Evolutionary},
object detection \cite{Ren2015Faster, He2017Mask},
semantic segmentation \cite{Shelhamer2014Fully, Chen2016DeepLab},
image captioning \cite{Vinyals2016Show, Johnson2015DenseCap},
visual question answering (VQA)\cite{Yang2016Stacked, Wu2017Image},
image generator \cite{Goodfellow2014Generative, Reed2016Generative}
(see Fig.~\ref{Fig:1}).
It is well known that
the logic reasoning is one of the abilities that
the general/strong artificial intelligence has to possess.
In the existing computer vision tasks,
image captioning and visual question answering seem to need some reasoning abilities,
especially VQA (indeed VQA performs need more knowledge: image itself, common sense, domain knowledge, and so on).
In fact, because of some shortcomings of existing benchmark data sets (described in Sect. \ref{sec:dataset1}),
the systems can correctly answer the questions without reasoning
\cite{Johnson2016CLEVR, hu2017learning, Zhang_2016_CVPR}.
Hence, it is desired to provide a new task,
such as the LiLi task,
to test the reasoning ability of models.

Our contributions are as follows:
\begin{enumerate}[(1)]
\item The data concept logic (DCL) is proposed to directly
 learn the concept logical patterns from the given data.
\item
  We propose a LiLi task where the abstract or low logical relation
  between two input images and one output image
  needs to be learned and reasoned
  without any reasoning patterns beforehand.
\item
  We provide an inference form of the LiLi task
  that is the consistent with classical propositional calculus form.
\item
  Six LiLi data sets with three difficulty levels:
  Bitwise And, Bitwise Or, Bitwise Xor, Addition, Subtraction and Multiplication,
  are provided.
\item
  Unlike a semantic or high level logical relation defined by human,
  an abstract or low level logical relation is expressed
  by a novel data-driven method called as LPN.
\item
  The performances of these typical neural networks:
  CNN-LSTM, MLP, CNN-MLP, Autoencoder and ResNets,
  are tested on six LiLi data sets.
  \item The divide and conquer model (DCM) is proposed using a decomposing strategy
  to solve the difficult task Multiplication,
  achieving a better performance than the typical neural networks used in this paper.
\end{enumerate}

The remainder of this paper is organized as follows:
Sect.~\ref{sec:DCL} proposes the DCL.
Sect.~\ref{sec:LiLi} proposes six LiLi data sets,
the LiLi task and its inference form.
Sect.~\ref{sec:experiment} presents the performance
evaluation of the typical neural networks
on six LiLi data sets.
In Sect.~\ref{sec:DCM}, the DCM is devised
to solve the difficult logic task Multiplication.
Finally, we draw conclusions in Sect.~\ref{sec:conclusion}.

\section{DCL}\label{sec:DCL}

In this section, we first
detail the DCL proposed in this paper,
and then provide an inference form of DCL.
\subsection{DCL}

Data concept logic (DCL)
is a data-driven tool for learning to obtain logic concepts
from a given data set directly.
Applying the learned concepts,
it can output the logical relations among the input data.
It is noted that DCL merely uses pure original data cues,
and can not know other information
such as the meaning of symbols/numbers in data in advance.
The DCL can be formalized as follows.

\begin{definition}
A data concept logic is termed as a triple $\mathcal{R}=(I,R,O)$,
where $I=\{x_{i}~|~ x_{i}=(x_{i}^{1},x_{i}^{2},\ldots,x_{i}^{m_{I}}),i=1,2,\ldots,N\}$
is an input sequence with the length $m_{I}$,
$O=\{y_{i}~|~y_{i}=(y_{i}^{1},y_{i}^{2},\ldots,y_{i}^{m_{O}}),i=1,2,\ldots,N\}$
is an output sequence with the length $m_{O}$,
$R:I\rightarrow O$ is a reasoning pattern (relation mapping) from the input $I$ to the output $O$.
\end{definition}

The aim of DCL is to learn the $R$ from the input $I$
to the output $O$. In this paper,
we propose a deep learning network framework: Logical Pattern Network ($LPN$)
parameterized by $W$ to learn the $R$.
This model can be learned by solving the following
optimization problem.

\begin{equation}\label{a}
\begin{aligned}
   W^{*} & = \arg \min_{W} \mathcal{L}(LPN_{W}(I),O)\\
         & = \arg \min_{W} \frac{1}{N}\sum_{i=1}^{N}{\mathcal{L}(LPN_{W}(x_{i}),y_{i})},
\end{aligned}
\end{equation}
where $\mathcal{L}$ is a loss function,
and $N$ is the number of the training samples.

The universal approximation theorem tells us that neural networks are able to approximate any measurable function with any precision \cite{Hornik1989Multilayer}.
Theoretically, the logical pattern $R$ can be represented by one neural network.
In the DCL, $R$ is hidden in the LPN,
and mining $R$ from data can be regarded as the iterative optimization process of parameter $W$ of LPN.
At each iteration, the value of $W$ changes in the direction that the loss $\mathcal{L}$ becomes smaller.
When the loss is small enough, the iteration stops and $R$ is obtained.

\begin{figure}[t]
\begin{center}
    \includegraphics[width=0.49\textwidth]{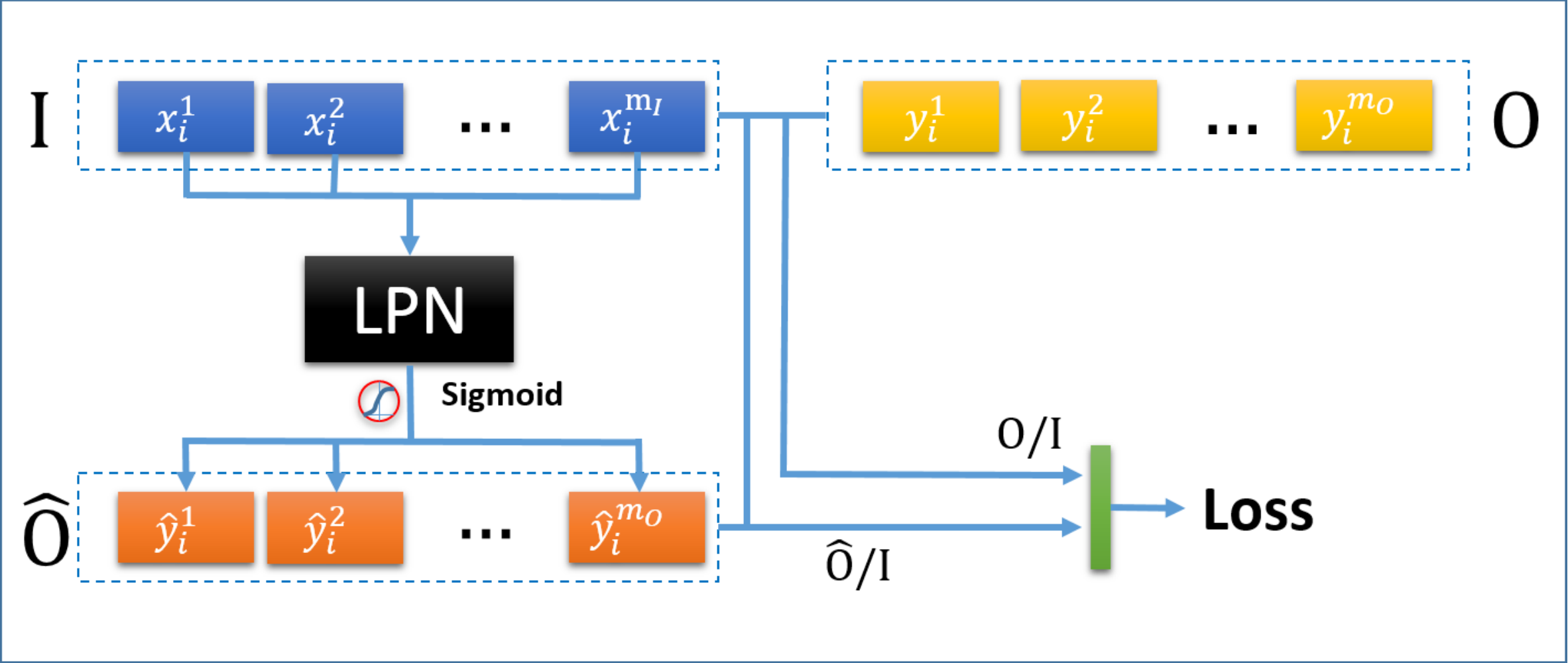}
\end{center}
    \caption{The workflow of a DCL task.}
\label{Fig:2}
\end{figure}

The workflow of a DCL task is illustrated in Fig. \ref{Fig:2},
where $I$ is the set of input data,
$O$ is the set of ground-truth output data,
$\hat{O}$ indicates the set of logical relation patterns reasoned
by $f(LPN_{W}(x_{i}^{1},x_{i}^{2},\ldots,x_{i}^{m_{I}}))$,
$O/I$ is the ground-truth logical relation set for a given input set $I$,
$\hat{O}/I$ is the prediction logical relation set for a given input set $I$ using $LPN$,
Loss is used to evaluate the difference between $O/I$ and $\hat{O}/I$.
$LPN$ indicates the logical pattern network.

\subsection{Inference form of DCL}\label{sec:DCLform}

Human, in our daily life, often makes inferences
using some known antecedents.
And this process can be formalized as the following form \cite{Mizumoto1982Comparison}.

\begin{equation}\label{DCLFMP}
  \begin{array}{lcr}
  Antecedent\ 1:       & A_{1}   &  \longrightarrow B_{1} \\
  Antecedent\ 2:       & A_{2}   &  \longrightarrow B_{2} \\
  ~~~~~~~~~~~~\vdots   & \vdots  & \\
  Antecedent\ n:       & A_{n}   &  \longrightarrow B_{n} \\
  Antecedent\ *:       & A_{*}\  & \\
  \hline
  Consequence:         &  &  B_{*}, \\
\end{array}
\end{equation}

Formula \ref{DCLFMP} exactly is also the mathematical model
of the classical propositional calculus \cite{Mizumoto1982Comparison}
where the consequence of the antecedent $*$ is inferred
using the known $n$ antecedents.
There exist many methods addressing the task.
For example, Zadeh \cite{Zadeh1973Outline} provided an inference rule
called `compositional rule of inference' (CRI)
to make such an inference whose antecedents and consequences contain fuzzy concepts.
Specially, an implication $A \rightarrow B$ first is translated into
a fuzzy relation $R_{z}(A, B)$ from $A$ to $B$.
And then, $B_{*}$ can be inferred by the composition of
$R_{z}$ and $A_{*}$ by the following formula.

\begin{equation}\label{F:DCLcomposition}
  B_{*} = R_{z}(A, B) \circ A_{*},
\end{equation}
where $R_{z}: [0,1]^{2} \rightarrow [0,1]$ defined beforehand by the human experts
is a duality function.
$\circ$ denotes a composition operator.

Inspired by fuzzy reasoning \cite{Mizumoto1982Comparison},
a DCL task can be written as the following inference form based on the IF THEN rule.

\begin{equation}\label{F:DCLGENERALFORM}
\begin{array}{ll}
  Antecedent\ 1:  & If\ the\ input\ is\ x_{1}\ then\ the\ output\ is\ y_{1} \\
  Antecedent\ 2:  & If\ the\ input\ is\ x_{2}\ then\ the\ output\ is\ y_{2} \\
  ~~~~~~~~~~~~\vdots &  ~~~~~~~~~~~~~~~~~~~~~~~~~~~~ \vdots \\
  Antecedent\ n:  & If\ the\ input\ is\ x_{n}\ then\ the\ output\ is\ y_{n} \\
  Antecedent\ n+1: & If\ the\ input\ is\ x_{n+1} \\
  Antecedent\ n+2: & If\ the\ input\ is\ x_{n+2} \\
  ~~~~~~~~~~~~\vdots &  ~~~~~~~~~~~~~ \vdots \\
  Antecedent\ n+m: & If\ the\ input\ is\ x_{n+m} \\\\
  \hline
  Consequence\ n+1:   &  The\ output\ is\ y_{n+1} \\
  Consequence\ n+2:   &  The\ output\ is\ y_{n+2} \\
  ~~~~~~~~~~~~\vdots  &  ~~~~~~~~~~~~~\vdots \\
  Consequence\ n+m:   &  The\ output\ is\ y_{n+m}, \\
\end{array}
\end{equation}
where $x_{i}$ is the input of the LPN, $y_{i}$ is the output of the LPN.

It should be noted that $x_{i}$ and $y_{i}$ can be many kinds of objects in LPN.
For example, antecedents and consequences contain fuzzy concepts as shown in Formula \ref{DCLFMP}.

In this paper, $x_{i}$ and $y_{i}$ are images.
Specifically, $x_{i}$ is an input sequence with the length $m_{I}$,
$y_{i}$ is an output sequence with the length $m_{O}$.
Based on this, Formula \ref{F:DCLGENERALFORM} can
be written as the following form.

\begin{equation}\label{F:DCLLiLiIFTHEN}
\begin{array}{lll}
  Antecedent\ 1:  & If\ the\ input\ squence\ is\ (x_{1}^{1}\ , x_{1}^{2}\ , \ldots, x_{1}^{m_{I}})\ \\
                  & then\ the\ output\ squence\ is\ (y_{1}^{1}\ , y_{1}^{2}\ , \ldots, y_{1}^{m_{O}}) \\
  Antecedent\ 2:  & If\ the\ input\ squence\ is\ (x_{2}^{1}\ , x_{2}^{2}\ , \ldots, x_{2}^{m_{I}})\ \\
                  & then\ the\ output\ squence\ is\ (y_{2}^{1}\ , y_{2}^{2}\ , \ldots, y_{2}^{m_{O}}) \\
  ~~~~~~~~~~~~\vdots       &   ~~~~~~~~~~~~~~~~~~~~~~~~~~~~~~~~~~~~ \vdots \\
  Antecedent\ n:  & If\ the\ input\ squence\ is\ (x_{n}^{1}\ , x_{n}^{2}\ , \ldots, x_{n}^{m_{I}})\ \\
                  & then\ the\ output\ squence\ is\ (y_{n}^{1}\ , y_{n}^{2}\ , \ldots, y_{n}^{m_{O}}) \\
  Antecedent\ n+1: & If\ the\ input\ squence\ is\ (x_{n+1}^{1}\ , x_{n+1}^{2}\ , \ldots, x_{n+1}^{m_{I}}) \\
  Antecedent\ n+2: & If\ the\ input\ squence\ is\ (x_{n+2}^{1}\ , x_{n+2}^{2}\ , \ldots, x_{n+2}^{m_{I}}) \\
  ~~~~~~~~~~~~\vdots &   ~~~~~~~~~~~~~~~~~~~~~~~~~~~~~~~~~~~~ \vdots \\
  Antecedent\ n+m: & If\ the\ input\ squence\ is\ (x_{n+m}^{1}\ , x_{n+m}^{2}\ , \ldots, x_{n+m}^{m_{I}}) \\\\
  \hline
  Consequence\ n+1:   &  The\ output\ squence\ is\ (y_{n+1}^{1}\ , y_{n+1}^{2}\ , \ldots, y_{n+1}^{m_{O}}) \\
  Consequence\ n+2:   &  The\ output\ squence\ is\ (y_{n+2}^{1}\ , y_{n+2}^{2}\ , \ldots, y_{n+2}^{m_{O}}) \\
  ~~~~~~~~~~~~\vdots  &  ~~~~~~~~~~~~~\vdots \\
  Consequence\ n+m:   &  The\ output\ squence\ is\ (y_{n+m}^{1}\ , y_{n+m}^{2}\ , \ldots, y_{n+m}^{m_{O}}), \\
\end{array}
\end{equation}
where $(x_{i}^{1}, x_{i}^{2}, \ldots$, $x_{i}^{m_{I}})$
is the input data fed into the LPN,
$(y_{i}^{1}, y_{i}^{2}, \ldots$, $y_{i}^{m_{O}})$
is the output data expressing the relation of the input data.

In Formula \ref{F:DCLLiLiIFTHEN},
the $n$ antecedents from 1 to $n$ constituting the training set
are used to train the LPN inference model.
And the $m$ antecedents from $n+1$ to $n+m$ constituting the testing set
are used to test the inference ability of LPN.
Based on this, Formula \ref{F:DCLLiLiIFTHEN} can
be further simplified as the following form.

\begin{equation}\label{DCLLiLi1}
\begin{array}{lcc}
  Training\ antecedent:\  & (x_{1}^{1},x_{1}^{2},\ldots,x_{1}^{m_{I}}) & \longrightarrow (y_{1}^{1}, y_{1}^{2} , \ldots, y_{1}^{m_{O}}) \\
                          & (x_{2}^{1},x_{2}^{2},\ldots,x_{2}^{m_{I}}) & \longrightarrow (y_{2}^{1}, y_{2}^{2} , \ldots, y_{2}^{m_{O}}) \\
                          &  \vdots  &  \vdots \\
                          & (x_{n}^{1},x_{n}^{2},\ldots,x_{n}^{m_{I}}) & \longrightarrow (y_{n}^{1}, y_{n}^{2} , \ldots, y_{n}^{m_{O}}) \\
  Testing\ antecedent:    & (x_{n+1}^{1},x_{n+1}^{2},\ldots,x_{n+1}^{m_{I}}) &  \\
                          & (x_{n+2}^{1},x_{n+2}^{2},\ldots,x_{n+2}^{m_{I}}) &  \\
                          &  \vdots \\
                          & (x_{n+m}^{1},x_{n+m}^{2},\ldots,x_{n+m}^{m_{I}}) &  \\
  \hline
   Consequence:   & & (y_{n+1}^{1}, y_{n+1}^{2} , \ldots, y_{n+1}^{m_{O}}) \\
                  & & (y_{n+2}^{1}, y_{n+2}^{2} , \ldots, y_{n+2}^{m_{O}}) \\
                  & & ~~~\vdots \\
                  & & (y_{n+m}^{1}, y_{n+m}^{2} , \ldots, y_{n+m}^{m_{O}}), \\
\end{array}
\end{equation}

Formula \ref{DCLLiLi1} can be further simplified as the following form
by $I_{train}$ $=$ $\{(x_{1}^{1},x_{1}^{2},$$\ldots,$$x_{1}^{m_{I}}),$
$(x_{2}^{1},x_{2}^{2},$$\ldots,$$x_{2}^{m_{I}}),$ $\ldots,$
$(x_{n}^{1},x_{n}^{2},$ $\ldots,x_{n}^{m_{I}})\},$
$O_{train}$ $= $$\{(y_{1}^{1},$$y_{1}^{2},$$\ldots,$$y_{1}^{m_{O}}),$
$(y_{2}^{1},$$y_{2}^{2},$$\ldots,$$y_{2}^{m_{O}}),$ $\ldots,$
$(y_{n}^{1},$$y_{n}^{2},$$\ldots,$$y_{n}^{m_{O}})\},$
$I_{test}$ $=$ $\{(x_{n+1}^{1},$$x_{n+1}^{2},$$\ldots,$$x_{n+1}^{m_{I}}),$
$(x_{n+2}^{1},$$x_{n+2}^{2},$ $\ldots,$$x_{n+2}^{m_{I}}),$$\ldots, $$(x_{n+m}^{1},x_{n+m}^{2},$$\ldots,$$x_{n+m}^{m_{I}})\},$
and $O_{test}$ $=$ $\{(y_{n+1}^{1},$$y_{n+1}^{2},$$\ldots,$$y_{n+1}^{m_{O}}),$
$(y_{n+2}^{1},$$y_{n+2}^{2},$$\ldots,$$y_{n+2}^{m_{O}}),$ $\ldots,$
$(y_{n+m}^{1},$$y_{n+m}^{2},$$\ldots,$$y_{n+m}^{m_{O}})\}$

\begin{equation}\label{DCLLiLi3}
  \begin{array}{rcc}
  Training\ antecedent\ set:  & I_{train} & \longrightarrow O_{train} \\
  Testing\ antecedent \ set:  & I_{test}  & \\
  \hline
  Consequence \ set:          &              & ~~~~~~O_{test},\\
\end{array}
\end{equation}

In fact, Formula \ref{DCLLiLi3} contains three implications, i.e.
$(I_{train} \rightarrow O_{train})\rightarrow (I_{test}\rightarrow O_{test})$.
One can obtain the consequence $O_{test}$ of the antecedent $I_{test}$
by translating three implications
to the following form.

\begin{equation}\label{F:composition}
  O_{test} = R(I_{train}, O_{train}) \circ I_{test},
\end{equation}
where $R(I_{train}, O_{train})$
learned using a data-driven method is a high-dimension function.

\begin{figure*}[t]
\begin{center}
    \includegraphics[width=0.9\textwidth]{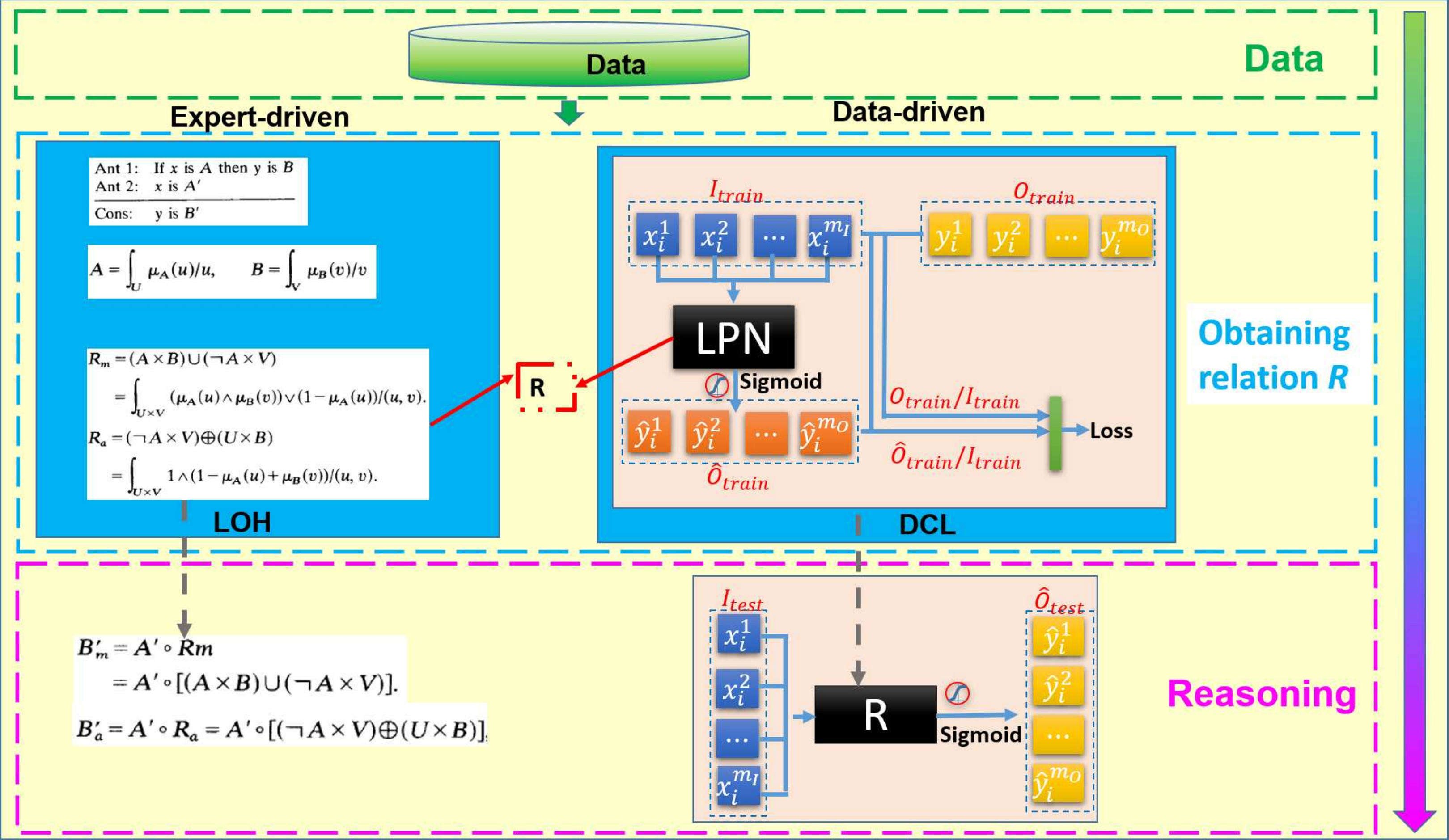}
\end{center}
    \caption{The comparative analysis between the DCL and the LOH.}
\label{Fig:3}
\end{figure*}

From the above analysis, one can find that
the DCL has the consistent inference form
with the classical propositional calculus.
The comparison of the DCL and the LOH is illustrated in Fig. \ref{Fig:3}.
From Fig. \ref{Fig:3}, one can see that one fundamental task of DCL or LOH
is to obtain the relation $R$.
For this task, they have a very obvious difference:
for LOH, $R$ needs to be defined beforehand by the experts, while
for DCL, $R$ is learned from a given data set.

Based on the above analysis,
it is desired to design a human-free and data-driven method
directly learn the reasoning pattern from given data.
In this study, we explore this problem by proposing the LiLi task.
What follows, the LiLi task will be detailed and formalized.

\section{A LiLi task}\label{sec:LiLi}

In this section, we first construct six LiLi data sets,
then detail the LiLi task proposed in this paper,
and finally provide its inference form consistent with
the classical propositional calculus form.

\subsection{LiLi data sets}\label{sec:dataset1}

\begin{figure*}
\begin{center}
	\subfigure[Bitwise And]{\label{fig:4a}\includegraphics[width=0.3\linewidth]{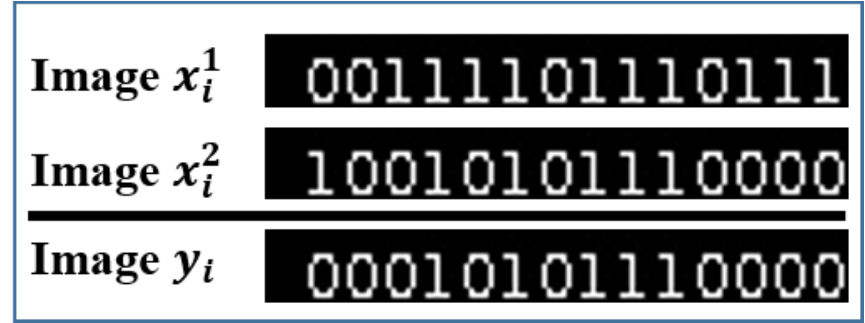}}
	\subfigure[Bitwise Or]{\label{fig:4b}\includegraphics[width=0.3\linewidth]{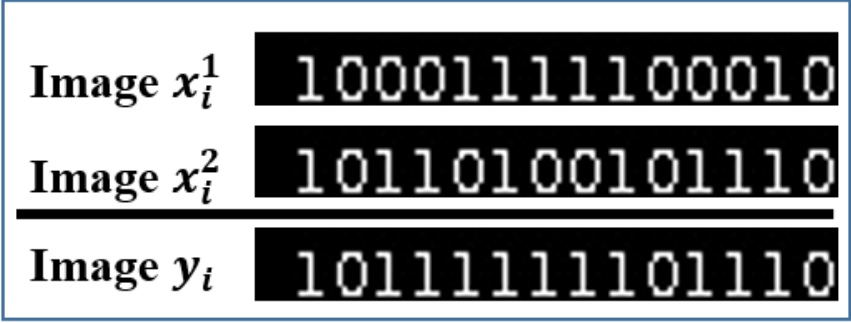}}
	\subfigure[Bitwise Xor]{\label{fig:4c}\includegraphics[width=0.3\linewidth]{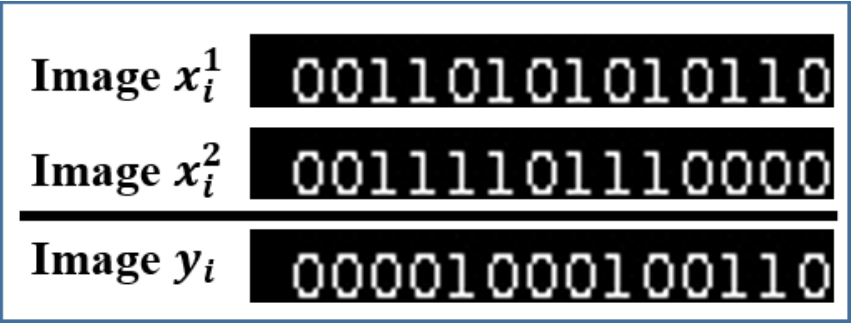}}\\
    \centering
    \subfigure[Addition]{\label{fig:4d}\includegraphics[width=0.3\linewidth]{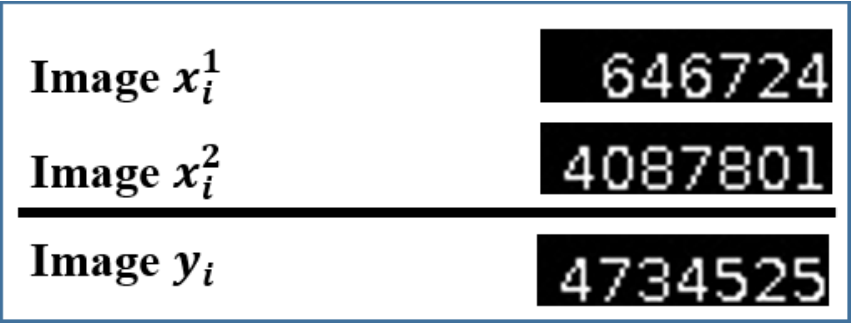}}
	\centering
    \subfigure[Subtraction]{\label{fig:4e}\includegraphics[width=0.3\linewidth]{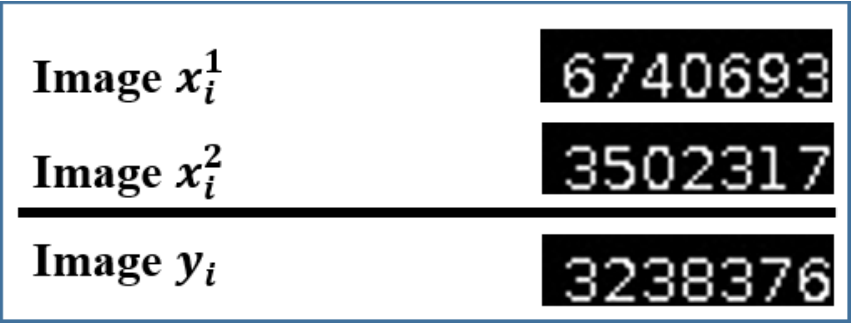}}
	\subfigure[Multiplication]{\label{fig:4f}\includegraphics[width=0.3\linewidth]{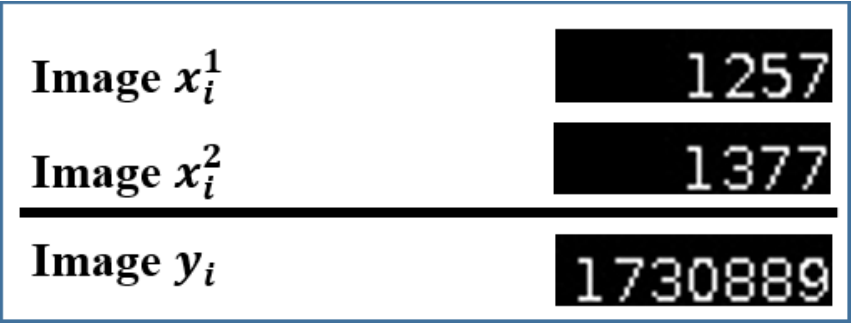}}
\end{center}
	\caption{The samples of six LiLi data sets.}
	\label{fig:4}
\end{figure*}

The existing logic reasoning data sets such as CLEVR \cite{Johnson2016CLEVR} and VQA \cite{Antol2015VQA}
have made outstanding contributions to testing the logic reasoning ability of machines,
but they have also some shortcomings:
(1) Because of biases of the data sets,
some questions can be answered through directly perceiving images rather than reasoning
\cite{Johnson2016CLEVR, hu2017learning, Zhang_2016_CVPR}.
For example, the question is what color is the object in the given image,
and the answer can be obtained directly from the image through perception.
(2) The existing logic reasoning data sets may seem complex,
but the typical neural networks and their results suggest that
the logics that are embedded in these data sets are relatively simple for machines.
More difficult logic data sets should be designed.
(3) Some questions from the existing logic reasoning data sets have multiple answers,
so it is not easy to judge whether the answers of these questions are correct or not.
These shortcomings make it difficult to assess the reasoning abilities of machines
using these data sets.

Therefore, we construct the LiLi data sets to overcome these shortcomings.
In this paper, these logical relations:
Bitwise And, Bitwise Or, Bitwise Xor, Addition, Subtraction and Multiplication
are selected to construct the LiLi data sets.
(1) Questions are able to be answered only if a model has both perception and reasoning abilities.
(2) The typical neural networks are almost powerless for the logic of multiplication (detailed in Sect. \ref{sec:experiment}).
It indicates that the LiLi task is really worth studying.
(3) The construction process of the LiLi data set is controlled by us and
only one correct answer can be obtained from each sample.
Hence, it is easy to evaluate the correctness of the answer.

We construct the LiLi data sets to verity the performance of the proposed LPN model.
It is worth noting that the LPN model does not know
the logical relations hidden in images beforehand.
The bitwise operations are binary numbers and arithmetic operations are decimals.
For Bitwise And, Bitwise Or and Bitwise Xor data sets,
the size of the images is set to 15 $\times$ 120,
so the number embedded in one image is at most a 14-digit number.
For Addition, Subtraction and Multiplication data sets,
the size of the images is set to 15 $\times$ 60,
hence the number embedded in one image is at most a 7-digit number.
This step ensures that the proportion of numbers used for training is a very small fraction of all possible combinations.
Each of these samples consists of two input images each containing an integer number.
The pair of two input images marked
$x_{i}^{1}$ and $x_{i}^{2}$
are then generated from a pre-specified range as detailed below.
The output image marked $y_{i}$ is generated according
to the result of the operation on the two input images.
The numbers embedded in images $x_{i}^{1}$, $x_{i}^{2}$
and $y_{i}$ are \emph{A}, \emph{B} and \emph{E}.

The details about these data sets are here:

\begin{itemize}
  \item[$\bullet$] Bitwise And:
  For per sample, both \emph{A} and \emph{B} have 14 binary digits.
  \emph{E} is the bitwise \emph{and} of \emph{A} and \emph{B}.
  For example, \emph{A} \emph{and} \emph{B}
  are ``\emph{00111101110111}'' and ``\emph{10010101110000}'', respectively.
  So, \emph{E} is ``\emph{00010101110000}''.
  The sample is shown in Fig.~\ref{fig:4a}.

  \item[$\bullet$] Bitwise Or:
  For per sample, both \emph{A} and \emph{B} have 14 binary digits.
  \emph{E} is the bitwise \emph{or} of \emph{A} and \emph{B}.
  For example, \emph{A} and \emph{B}
  are ``\emph{10001111100010}'' and ``\emph{10110100101110}'', respectively.
  So, \emph{E} is ``\emph{10111111101110}''.
  The sample is shown in Fig.~\ref{fig:4b}.

  \item[$\bullet$] Bitwise Xor:
  For per sample, both \emph{A} and \emph{B} have 14 binary digits.
  \emph{E} is the bitwise \emph{xor} of \emph{A} and \emph{B}.
  For example, \emph{A} and \emph{B}
  are ``\emph{00110101010110}'' and ``\emph{00111101110000}'', respectively.
  So, \emph{E} is ``\emph{00001000100110}''.
  The sample is shown in Fig.~\ref{fig:4c}.

  \item[$\bullet$] Addition:
  For per sample, the range of \emph{A} and \emph{B} are 0$\sim$4999999.
  \emph{E} is the sum of \emph{A} and \emph{B}.
  For example, \emph{A} and \emph{B}
  are ``\emph{646724}'' and ``\emph{4087801}'', respectively.
  So, \emph{E} is ``\emph{4734525}''.
  The sample is shown in Fig.~\ref{fig:4d}.

  \item[$\bullet$] Subtraction:
  For per sample, the range of \emph{A} and \emph{B} are 0$\sim$9999999.
  \emph{E} is the difference between \emph{A} and \emph{B}.
  In order to ensure a positive result, \emph{A}
  is chosen to be larger or equal to \emph{B}.
  For example, \emph{A} and \emph{B}
  are ``\emph{6740693}'' and ``\emph{3502317}'', respectively.
  So, \emph{E} is ``\emph{3238376}''.
  The sample is shown in Fig.~\ref{fig:4e}.

  \item[$\bullet$] Multiplication:
  For per sample, the range of \emph{A} and \emph{B} are 0$\sim$3160.
  \emph{E} is the product of \emph{A} and \emph{B}.
  For example, \emph{A} and \emph{B}
  are ``\emph{1257}'' and ``\emph{1377}'', respectively.
  So, \emph{E} is ``\emph{1730889}''.
  The sample is shown in Fig.~\ref{fig:4f}.
\end{itemize}

According to the difficulty of the logical relations embedded in data sets,
these data sets are divided into 3 levels: one-star($\star$, easy), two-star($\star$$\star$, intermediate), and three-star($\star$$\star$$\star$, difficult).

Bitwise And, Bitwise Or and Bitwise Xor data sets ({$\star$}):
(1) The value of each digit of \emph{E} is only determined by
the values at the same position in \emph{A} and \emph{B},
e.g., in Fig.~\ref{fig:4a},
the value at 2$^{th}$ (the rightmost position is 1$^{th}$) position in \emph{E} is only determined by
the values at 2$^{th}$ position in \emph{A} and \emph{B} ,
so the value at 2$^{th}$ position in \emph{E} is ``\emph{0}'' (1\&0=0);
(2) The possible value of each digit of \emph{E} is 0 or 1.

Addition and Subtraction data sets ({$\star$$\star$}):
(1) The value of each digit of \emph{E} is determined by the carry or borrow and
the values at the same position in \emph{A} and \emph{B},
e.g., in Fig.~\ref{fig:4d},
the value at 2$^{th}$ position in \emph{E} is determined by
the carry of the sum of values at 1$^{th}$ position in \emph{A} and \emph{B} and
the values at 2$^{th}$ position in \emph{A} and \emph{B};
(2) The possible value of carry or borrow part is 0 or 1,
so the possible value of each digit (except the rightmost position) of \emph{E} has two possibilities in 0$\sim$9,
we choose one of the two possibilities as the final result based on the carry or borrow case.
E.g., in Fig.~\ref{fig:4d}, the carry of the sum of values at 1$^{th}$ position in \emph{A} and \emph{B} is ``\emph{0}'',
the sum of values at 2$^{th}$ position in \emph{A} and \emph{B} is ``\emph{2}'' (2+0=2),
so the value at 2$^{th}$ position in \emph{E} is ``\emph{2}'' (0+2=2).

Multiplication data set ({$\star$$\star$$\star$}):
(1) The value at a given position in \emph{E} is determined by
the values at the given positions in \emph{A} and \emph{B} and
all positions in \emph{A} and \emph{B} before that given position.
E.g., in Fig.~\ref{fig:4f},
the value at 2$^{th}$ position in \emph{E} is determined by
the values at 1$^{th}$ and 2$^{th}$ positions in \emph{A} and
the values at 1$^{th}$ and 2$^{th}$ positions in \emph{B}.
(2) The number of the possible value of each digit (except the rightmost position) of the \emph{E}
on Multiplication data set is more than that on other LiLi data sets.

\subsection{LiLi task}
In this paper, we focus on the scene
where a model directly learns and reasons
the relation between two input images and one output image,
without any reasoning patterns beforehand.
In this task, we first generate three images,
two for the input and one for the output.
The output image expresses the relation between two input images.
In addition, the n-digit number embedded
in the images are not explicitly introduced,
which means that the meaning of contents embedded in
images and the relation between two input images and one output image
are not known at all.
One example is used to illustrate the LiLi task.
If the n-digit numbers embedded in two input images are
``$\textit{234}$" and``$\textit{432}$",
the output image are ``$\textit{666}$",
the logical relation between two input images and the output image
is addition. It can be formalized as follows.

Given a data concept logic system as a set of triple $\mathcal{R}=(I,R,O)$,
where $I$$=$$\{x_{i}~$$|$$~ x_{i}$$=$$(x_{i}^{1},$$x_{i}^{2}),$$i=1,2,\ldots,N\}$
is an input sequence,
$O=\{y_{i}\}_{i=1}^{N}$ is the output sequence,
where $x_{i}^{1},x_{i}^{2}$ and $y_{i}$ are three images with $K$ pixels
shown in Fig. \ref{fig:4}.
$R$ denotes the logical relation between the pair of images
$x_{i} \in I$ and $y_{i} \in O$.

At the semantic or high level, $R$ is called as
\emph{Bitwise And, Bitwise Or, Bitwise Xor, Addition, Subtraction and Multiplication}
denoted as $\&, |, \wedge, +, -$ or $\times$,
and they are easily understood by human beings.
However, at the abstract or low level,
$R$ may be a high-dimensional mapping
that is extremely difficult to define the mapping by human,
in this paper, $R:[-1,1]^{2K} \rightarrow \{0, 1\}^{K}$.
Hence, it is desired to design a novel method
to express an abstract or low level logical relation.

In this task, given a data set $D=\{(x_{i},y_{i})\}_{i=1}^{N}$,
where $y_{i}$ denotes the logical relation between the pair of images
$x_{i}^{1}$ and $x_{i}^{2}$.
When drawing these images, we use the pixel value 0 for black, the pixel value 1 for white.
For the input images, we scale every pixel value into -1 $\sim$ 1 by subtracting the mean,
so $x_{i}^{1}, x_{i}^{2} \in [-1,1] ^{K}$.
For the output image $y_{i} \in \{0,1\} ^{K}$.
This task can be viewed as finding a mapping
from the input space $I=\{x_{i}\}_{i=1}^{N}$
to the output space $O=\{y_{i}\}_{i=1}^{N}$
by a supervised learning strategy.
In this study, this task can be transformed into a regression problem
with the Mean Square Error (MSE) loss function, i.e. $\mathcal{L}$ is $MSE$.
It can be by solving the following
optimization problem.

\begin{equation}\label{LPNimage}
\begin{aligned}
  W^{*} & = \arg \min_{W} MSE(f(LPN_{W}(I)),O) \\
   & = \arg \min_{W} \frac{1}{N}\sum_{i=1}^{N}{MSE(f(LPN_{W}(x_{i}^{1},x_{i}^{2})),y_{i})}\\
   & = \arg \min_{W} \frac{1}{N}\sum_{i=1}^{N}\sqrt{\sum_{k=1}^{K}{(f(LPN_{W}(x_{i}^{1},x_{i}^{2}))_{k}-{y_{i}}_{k})^{2}}},
\end{aligned}
\end{equation}
where $f$ is a sigmoid function to transform $LPN_{W}(x_{i}^{1},x_{i}^{2})$ to [0,1],
i.e. $f(LPN_{W}$$(x_{i}^{1},x_{i}^{2})$$)$ $\in$ $[0,1]^{K}$,
and LPN is parameterized by $W$.
Formula \ref{LPNimage}
is differentiable with respect to the parameter $W$,
and can be efficiently solved by using the gradient descent method.

\begin{figure}
\begin{center}
    \includegraphics[width=0.49\textwidth]{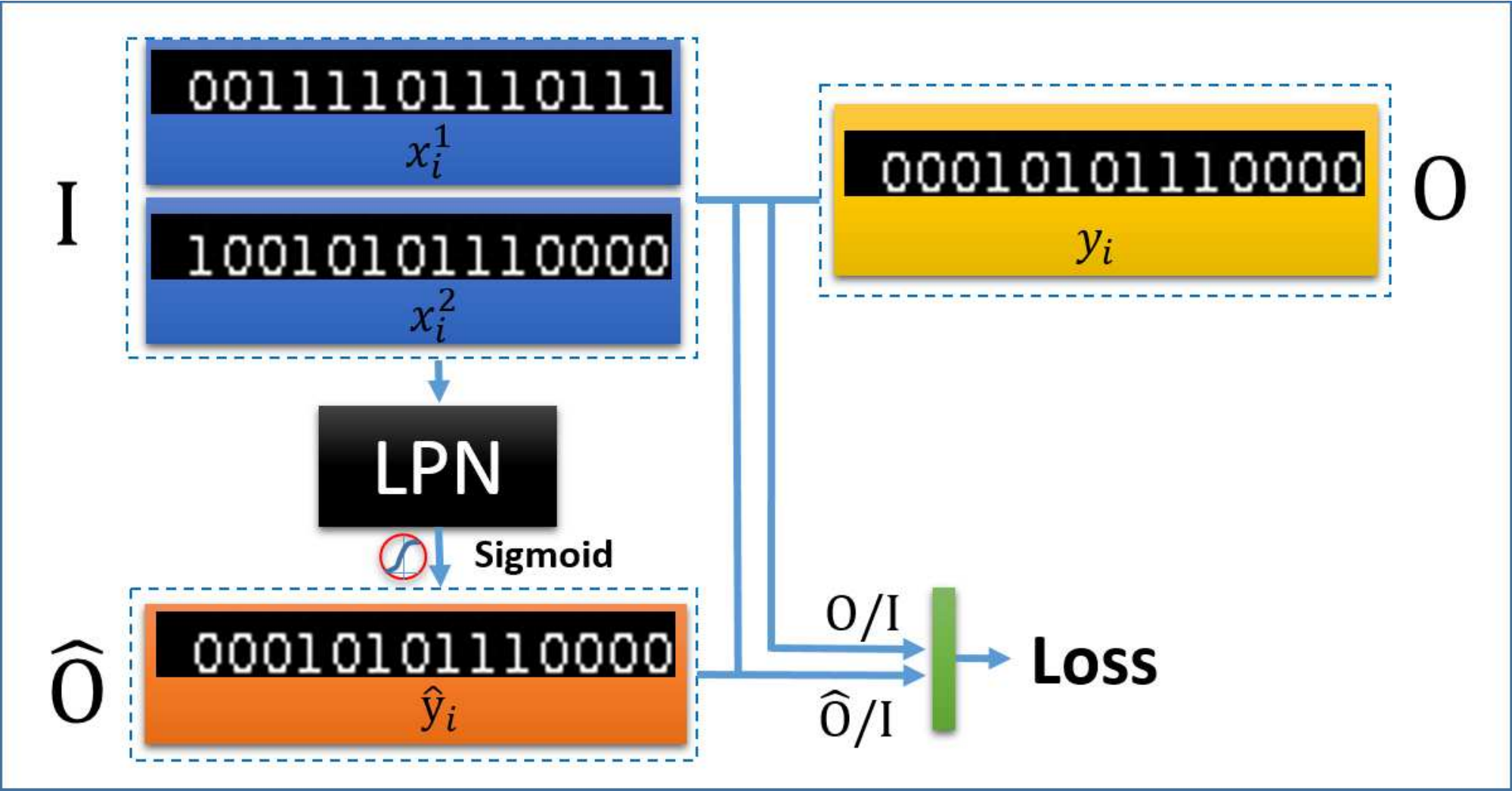}
\end{center}
    \caption{The workflow of a LiLi task.}
\label{fig:5}
\end{figure}

Based on above analysis and discussion,
we illustrate the workflow of the LiLi task shown in Fig. \ref{fig:5},
where $I$ is the set of input image data,
$O$ is the set of ground-truth output image data,
$\hat{O}$ indicates the set of logical relation patterns reasoned
by $f(LPN_{W}(x_{i}^{1},x_{i}^{2}))$,
$O/I$ is the ground-truth logical relation set for a given input image set $I$,
$\hat{O}/I$ is the prediction logical relation set for a given input image set $I$ using $LPN$,
Loss is used to evaluate the difference between $O/I$ and $\hat{O}/I$.
$LPN$ indicates the logical pattern network,
which is implemented in this paper using CNN-LSTM, MLP, Autoencoder,
ResNet18, ResNet50, ResNet152 and DCM, respectively.
More implementation details about LPN see Sects. \ref{set:models} and \ref{sec:DCM}.

From Formula \ref{LPNimage} and Fig. \ref{fig:5}, one observes that
the LPN merely needs to be provided some training data
to automatically learn the logical patterns
between a pair of the given images
without providing any reasoning patterns beforehand.
This is an absolutely data-driven strategy to mine the logical patterns hidden in data.

\subsection{Inference form of a LiLi task}

Based on the inference form of the DCL \ref{sec:DCLform},
a LiLi task can be written as the following inference form based on the IF THEN rule.

\begin{equation}\label{F:LiLiIFTHEN}
\begin{array}{ll}
  Antecedent\ 1:  & If\ two\ input\ images\ are\ x_{1}^{1}\ and\ x_{1}^{2}\ then\ the\ output\ image\ is\ y_{1}\\
  Antecedent\ 2:  & If\ two\ input\ images\ are\ x_{2}^{1}\ and\ x_{2}^{2}\ then\ the\ output\ image\ is\ y_{2}\\
  ~~~~~~~~~~~~\vdots       &   ~~~~~~~~~~~~~~~~~~~~~~~~~~~~~~~~~~~~ \vdots \\
  Antecedent\ n:  & If\ two\ input\ images\ are\ x_{n}^{1}\ and\ x_{n}^{2}\ then\ the\ output\ image\ is\ y_{n}\\
  Antecedent\ n+1: & If\ two\ input\ images\ are\ x_{n+1}^{1}\ and\ x_{n+1}^{2}\ \\
  Antecedent\ n+2: & If\ two\ input\ images\ are\ x_{n+2}^{1}\ and\  x_{n+2}^{2}\\
  ~~~~~~~~~~~~\vdots &   ~~~~~~~~~~~~~~~~~~~~~~~~~~~~~~~~~~~~ \vdots \\
  Antecedent\ n+m: & If\ two\ input\ images\ are\ x_{n+m}^{1}\ and x_{n+m}^{2}\\
  \hline
  Consequence\ n+1:   &  The\ output\ image\ is\ y_{n+1}\\
  Consequence\ n+2:   &  The\ output\ image\ is\ y_{n+2}\\
  ~~~~~~~~~~~~\vdots  &  ~~~~~~~~~~~~~\vdots \\
  Consequence\ n+m:   &  The\ output\ image\ is\ y_{n+m},\\
\end{array}
\end{equation}
where $x_{i}^{1}$ and $x_{i}^{2}$ are the input images,
$y_{i}$ is the output image expressing the relation between two input images.

In Formula \ref{F:LiLiIFTHEN},
the $n$ antecedents from 1 to $n$ constituting the training set
are used to train the LPN inference model.
And the $m$ antecedents from $n+1$ to $n+m$ constituting the testing set
are used to test the inference ability of LPN.
Based on this, Formula \ref{F:LiLiIFTHEN} can
be further simplified as the following form.

\begin{equation}\label{LiLi1}
\begin{array}{lcl}
  Training\ antecedent:\  & (x_{1}^{1},x_{1}^{2}) & \longrightarrow y_{1} \\
                          & (x_{2}^{1},x_{2}^{2}) & \longrightarrow y_{2}\\
                          &  \vdots \\
                          & (x_{n}^{1},x_{n}^{2}) & \longrightarrow y_{n}\\
  Testing\ antecedent:    & (x_{n+1}^{1},x_{n+1}^{2}) &  \\
                          & (x_{n+2}^{1},x_{n+2}^{2}) &  \\
                          &  \vdots \\
                          & (x_{n+m}^{1},x_{n+m}^{2}) &  \\
  \hline
   Consequence:   & & y_{n+1}\\
                  & & y_{n+2}\\
                  & & ~~~\vdots \\
                  & & y_{n+m},\\
\end{array}
\end{equation}

Formula \ref{LiLi1} can be further simplified as the following form
by $I_{train}$ $=$ $\{(x_{1}^{1},$$x_{1}^{2}),$
$(x_{2}^{1},$$x_{2}^{2}),$$\ldots,$
$(x_{n}^{1},$$x_{n}^{2})\}$,
$O_{train}$ $=$ $\{y_{1},$ $y_{2},$$\ldots,$$y_{n}\}$,
$I_{test}$ $=$ $\{(x_{n+1}^{1},$$x_{n+1}^{2}),$
$(x_{n+2}^{1},$$x_{n+2}^{2}),$$\ldots,$
$(x_{n+m}^{1},$$x_{n+m}^{2})\}$,
and
$O_{test}$ $=$ $\{y_{n+1},$$y_{n+2},$$\ldots,$$y_{n+m}\}$.

\begin{equation}\label{LiLi3}
  \begin{array}{rcc}
  Training\ antecedent\ set:  & I_{train} & \longrightarrow O_{train} \\
  Testing\ antecedent \ set:  & I_{test}  & \\
  \hline
  Consequence \ set:          &              & ~~~~~~O_{test},\\
\end{array}
\end{equation}

One can obtain the consequence $O_{test}$ of the antecedent $I_{test}$
by translating three implications
$(I_{train} \rightarrow O_{train})\rightarrow (I_{test}\rightarrow O_{test})$
included by Formula \ref{LiLi3}
to the following form.

\begin{equation}\label{F:composition}
  O_{test} = R(I_{train}, O_{train}) \circ I_{test},
\end{equation}
where $R(I_{train}, O_{train}): [-1,1]^{2K} \rightarrow \{0,1\}^{K}$
learned using a data-driven method is a high-dimension mapping function.

According to the above analysis, one can find that
on the one hand, the LiLi task has the consistent inference form
with the classical propositional calculus,
on the other hand they have some different aspects as follows.
\begin{itemize}
  \item[$\bullet$] $R_{z}: [0,1]^{2} \rightarrow [0,1]$ is a duality function.
  However, $R(I_{train}, O_{train}): [-1,1]^{2K} \rightarrow \{0,1\}^{K}$
  is a complex function with high dimensions ($K$ takes 1800 or 900 in this paper).
  \item[$\bullet$] $R_{z}$ needs to be defined beforehand by the experts, while
  $R$ is learned from a given data set because it is almost impossible
  to be defined the function beforehand by human.
\end{itemize}

In real world, there exist many complex relations
that can not be provided beforehand by human beings.
When facing this situation, the classical propositional calculus
can not work well, even cannot work.
Hence, it is desired to design a human-free and data-driven method
to learn an unknown relation function.
This is the our most main motivation.

\section{Experiments}\label{sec:experiment}

In this section, we compare the performances of
several typical deep neural networks on the six LiLi data sets.
Next, we detail used models and experimental setup.

\subsection{Models and experimental setup}\label{set:models}

For all models, two images as input are fed into the models,
and one image as output is used to compare with the  ground truth image.
These models are trained to produce one output image
in which the correct number is embedded
by optimising a mean square error (mse) loss
and using the ADAM or SGD optimiser.
The early-stopping is used to choose the optimiser and hyper-parameters of
smallest loss estimated on the validation set.
In addition, the batch size is set to 32.
The hyper-parameter settings and further
details on all models see in Table~\ref{tab:model_detail_all}.
Finally, the performance values are reported on the testing set.

\begin{table*}
  \caption{The hyper-parameter settings on all models.}\label{tab:model_detail_all}
  \centering
  \begin{tabular}{l|l}
  \toprule
   Model       & hyper-parameter\\
  \hline
      \multirow{3}{*}{CNN-LSTM}   & Conv(32,(5,5),l2(1.e-4))-$>$BatchNormalization()-$>$MaxPooling((2,2))-$>$ \\
                                  & Conv(64,(3,3),l2(1.e-4))-$>$BatchNormalization()-$>$MaxPooling((2,2))-$>$ \\
                                  & LSTM(1024, dropout=0.5) \\
  \hline
      MLP        & Dense(256)-$>$Dense(256)-$>$Dense(256) \\
  \hline
      \multirow{3}{*}{CNN-MLP}    & Conv(32,(5,5))-$>$BatchNormalization()-$>$MaxPooling((2,2))-$>$ \\
                                  & Conv(64,(3,3))-$>$BatchNormalization()-$>$MaxPooling((2,2))-$>$ \\
                                  & Dense(4096) \\
  \hline

      \multirow{3}{*}{Autoencoder}& Conv(32,(5,5))-$>$MaxPooling((2,2))-$>$Conv(64,(5,5))-$>$MaxPooling((2,2)) \\
                                  & Conv(64,(5,5))-$>$UpSampling((2,2))-$>$Conv(32,(5,5))-$>$UpSampling((2,2)) \\
                                  & Cropping2D(((0,1),(0,0)))-$>$Conv(1,(5,5)) \\
  \bottomrule
\end{tabular}
\end{table*}

\begin{itemize}
  \item[$\bullet$] CNN-LSTM:
  We develop the model using a standard LSTM module \cite{Graves1997Long}.
  Since LSTMs are designed to process inputs sequentially,
  we first pass images sequentially and independently through
  a 2-layer CNN, and the resulting sequence is handed over to the LSTM.
  The final hidden state of the LSTM is passed
  through a fully-connected layer with sigmoid activation function.
  The model is trained using batch normalization after each convolutional layer
  and dropout is applied to the LSTM hidden state.

  \item[$\bullet$] MLP: The MLP is implemented followed by \cite{hoshen2016visual}.
  The model has three hidden layers each with 256
  nodes with ReLU activation functions and one output layer with sigmoid activation.
  All nodes between adjacent layers are fully-connected.

  \item[$\bullet$] CNN-MLP: Inspired by \cite{lecun2015deep},
  we implement a 2-layer CNN with ReLU activation functions and batch normalizations.
  The input images are treated as a set of
  separate greyscale input feature maps for the CNN.
  The convolutional output is passed through two-layer fully-connected layers, in which the first layer using a ReLU activation function
  and the second layer using a sigmoid activation function.

  \item[$\bullet$] Autoencoder: A simple autoencoder network is implemented
  using the idea of \cite{hinton2006reducing}.
  In this model, a 2-layer CNN is used as the encoder network
  and a 2-layer upsampling network as the decoder network.
  At last, a convolutional layer is used as the output layer with a sigmoid activation.

  \item[$\bullet$] ResNet: We use ResNet architecture as described in \cite{He2016Deep} and
  modify the softmax activation function to sigmoid activation function
  on the last layer of the network.
  In this paper, we train ResNet-18, ResNet-50 and ResNet-152
  on all LiLi data sets and get nearly performances.
\end{itemize}

\subsection{Experiments and analysis on LiLi data sets}

\begin{figure*}[t]
\begin{center}
	\subfigure[Bitwise And]{\label{fig:6a}\includegraphics[width=0.3\linewidth]{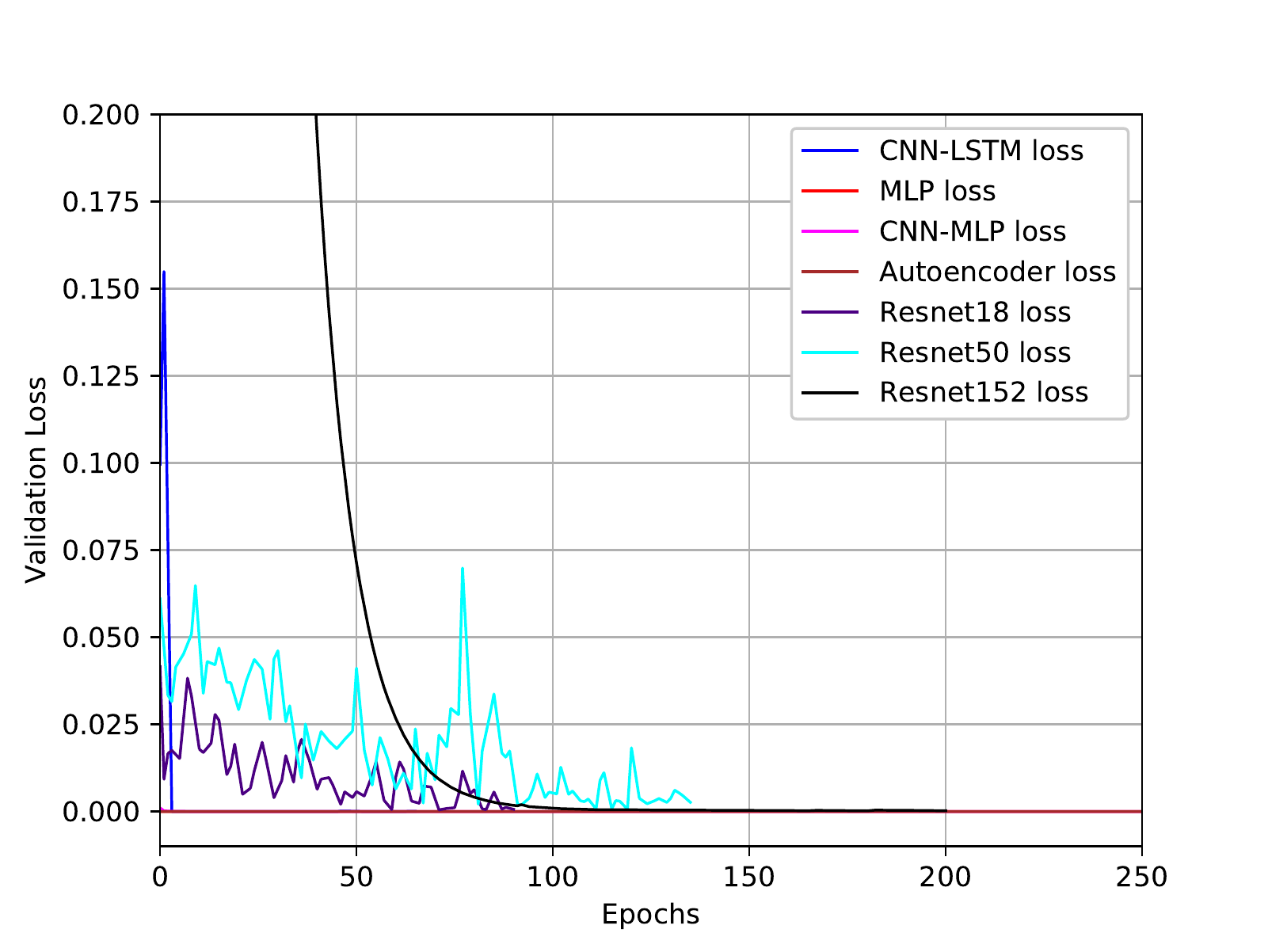}}
	\subfigure[Bitwise Or]{\label{fig:6b}\includegraphics[width=0.3\linewidth]{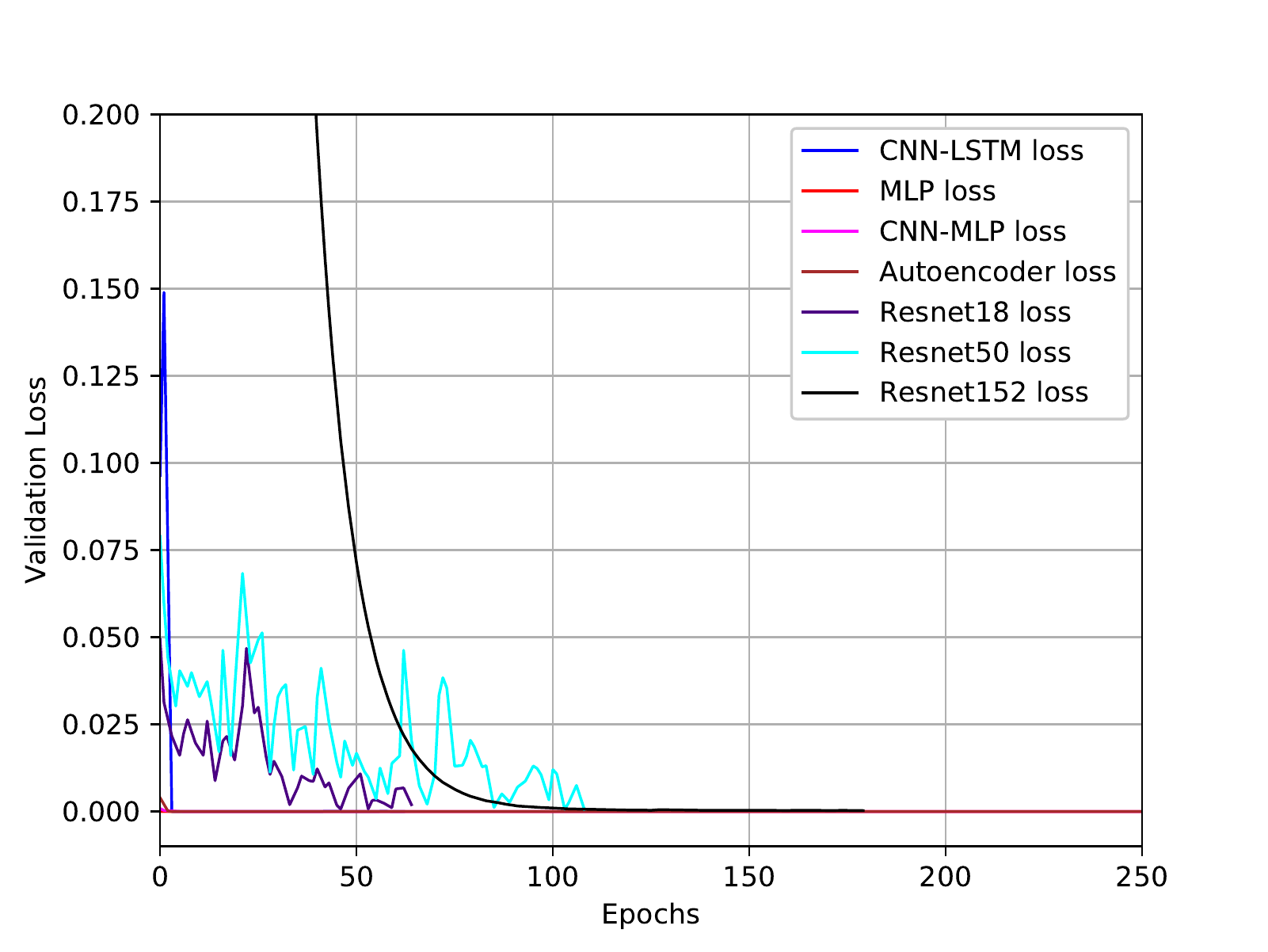}}
    \subfigure[Bitwise Xor]{\label{fig:6c}\includegraphics[width=0.3\linewidth]{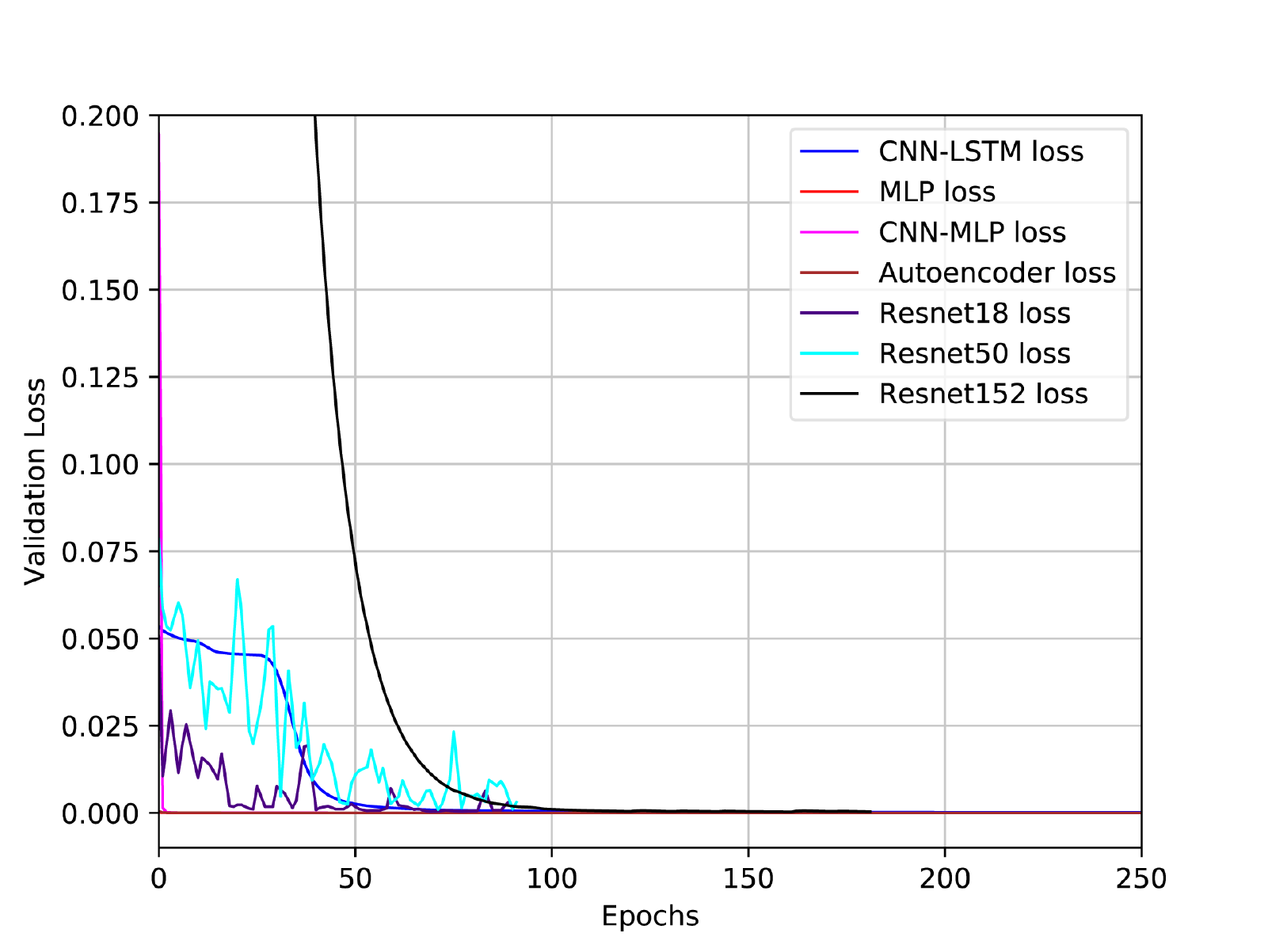}}
	\subfigure[Addition]{\label{fig:6d}\includegraphics[width=0.3\linewidth]{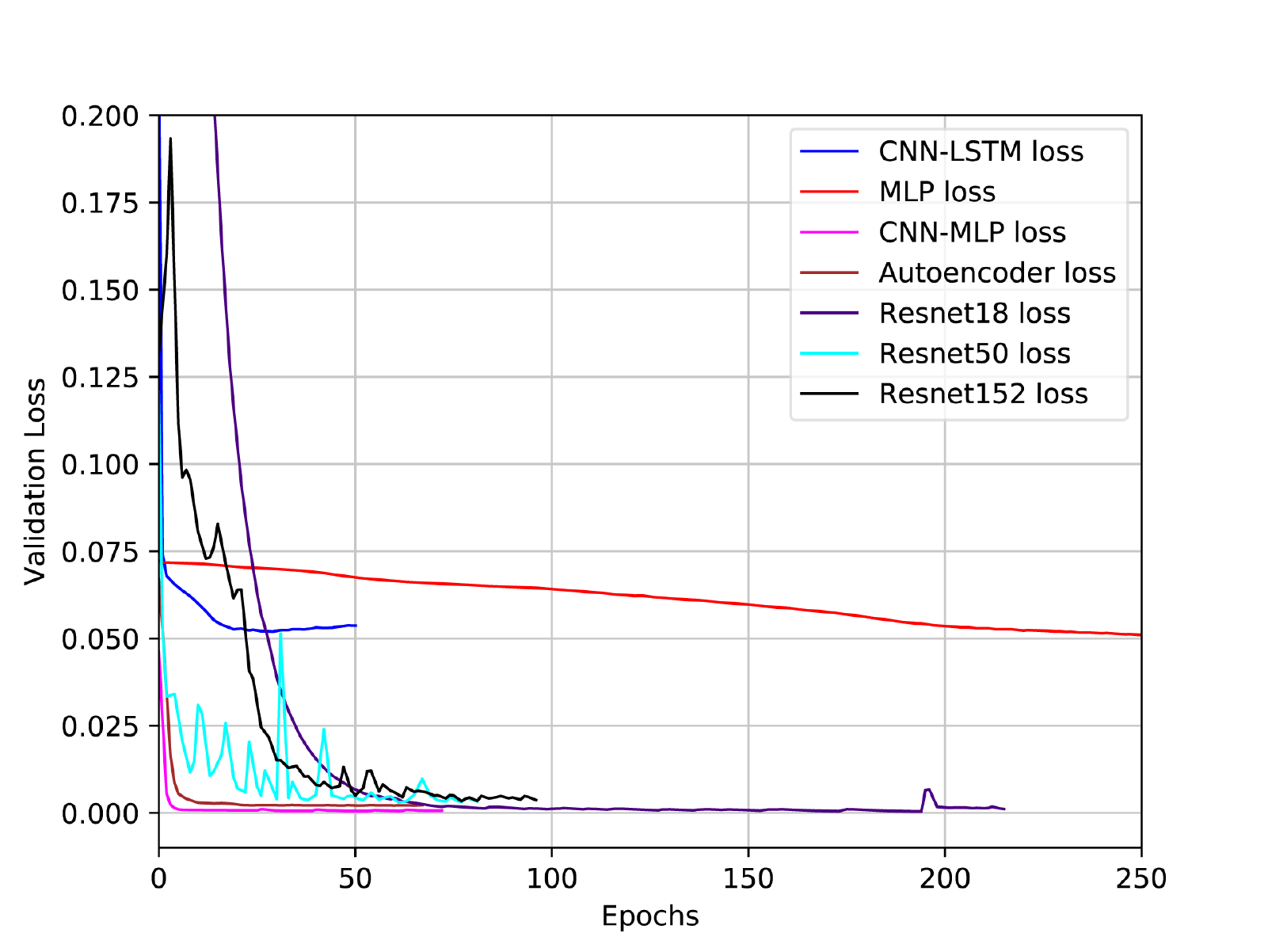}}
    \subfigure[Subtraction]{\label{fig:6e}\includegraphics[width=0.3\linewidth]{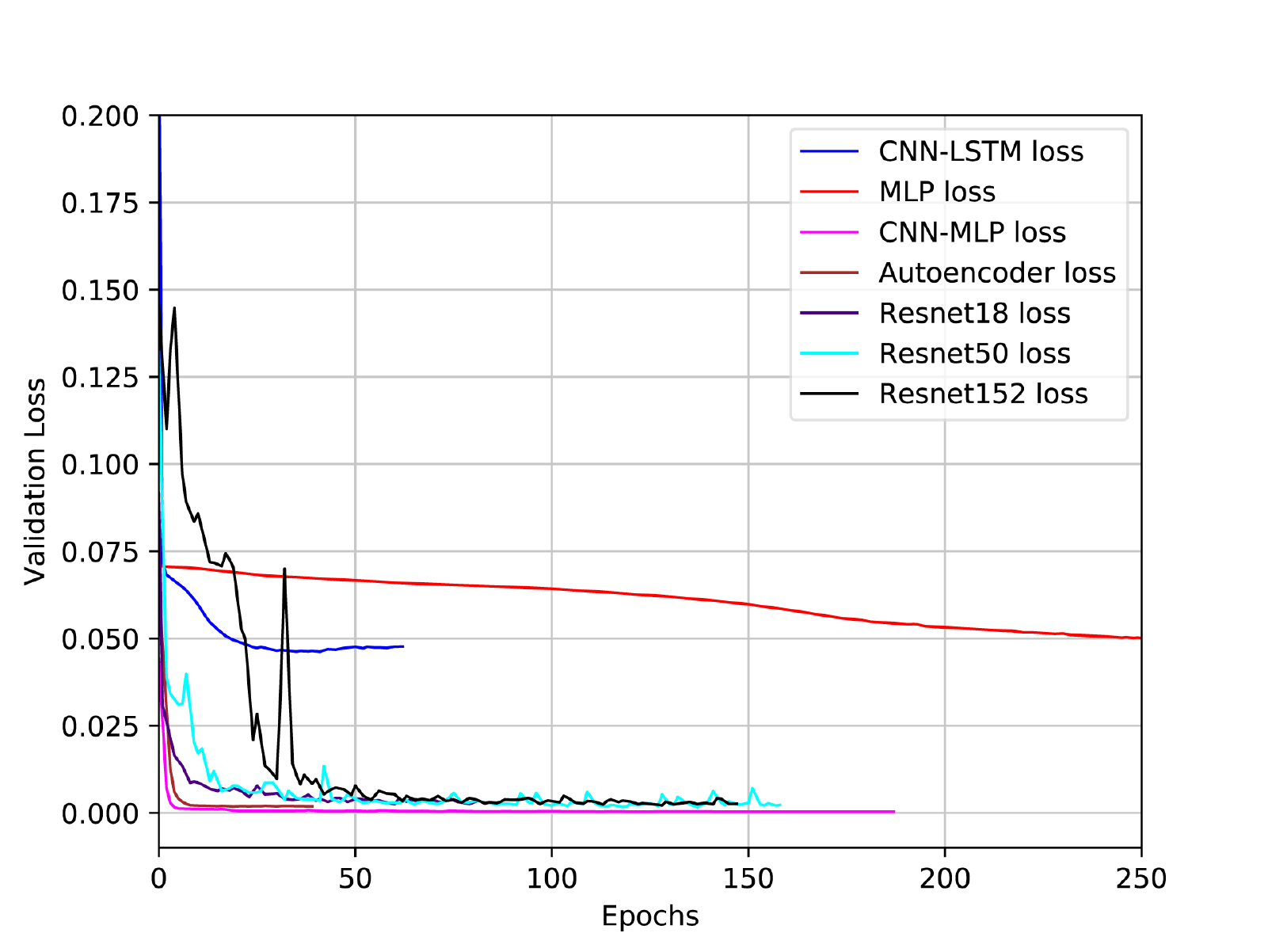}}
	\subfigure[Multiplication]{\label{fig:6f}\includegraphics[width=0.3\linewidth]{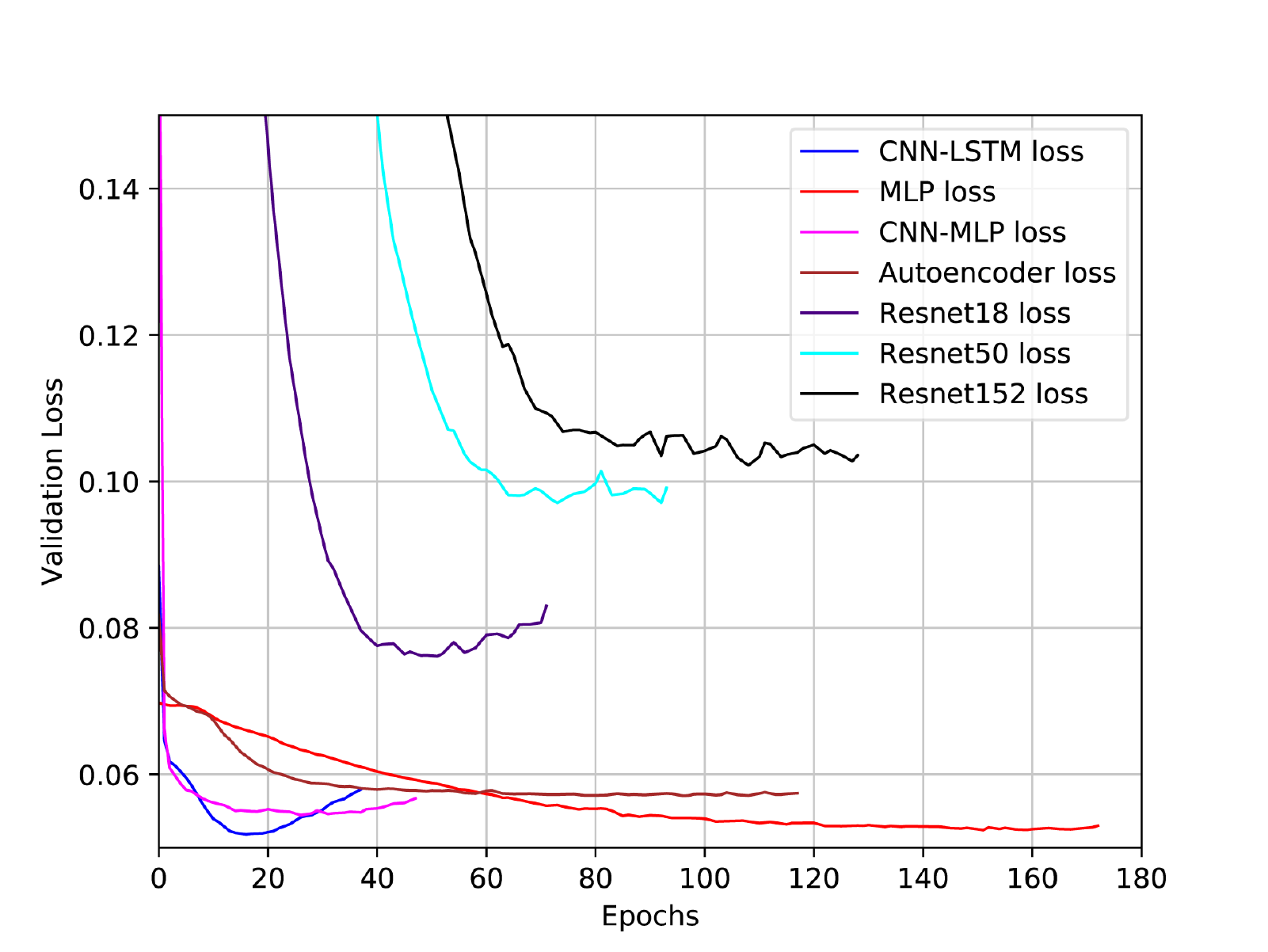}}
\end{center}
	\caption{The validation losses of Bitwise And, Bitwise Or, Bitwise Xor, Addition, Subtraction and Multiplication on 10,000 training data sets.}
	\label{fig:6}
\end{figure*}

In this subsection, we test several typical deep neural networks on these LiLi data sets.
Each data set consists of 10,000 training samples,
10,000 validation samples and 20,000 testing samples.
The testing samples are not included in the training or validation samples.
All models are trained on each training set and stopped when the losses on validation sets no longer decrease.
We use an OCR software \cite{smith2007overview} to recognize
the numbers embedded in the predicted images,
and then compare them with the ground truth numbers.
For one predicted image, it is right when all digits are equal to the ground truth digits.
The accuracies of Bitwise And, Bitwise Or, Bitwise Xor, Addition, Subtraction and Multiplication data sets are shown in Table~\ref{tab:1w_accuracy}.

From Table~\ref{tab:1w_accuracy}, one observes that
all models get the good performances on
Bitwise And, Bitwise Or and Bitwise Xor data sets.
Only CNN-MLP, Autoencoder and ResNets get the good performances on Addition and Subtraction data sets.
Unfortunately, all models fail on Multiplication data set.

\begin{table*}
  \caption{The test accuracies of Bitwise And, Bitwise Or, Bitwise Xor, Addition, Subtraction and Multiplication
  on 10,000 training data sets.}\label{tab:1w_accuracy}
  \centering
  \begin{tabular}{l|ccc|cc|c}
  \toprule
   \multirow{3}{*}{Model}  & \multicolumn{6}{c}{Operations}\\
  \cline{2-7}
               &\multicolumn{3}{c|}{$\star$}& \multicolumn{2}{c|}{$\star \star$}&  $\star \star \star$\\
  \cline{2-7}
               & Bitwise And & Bitwise Or & Bitwise Xor & Addition & Subtraction & Multiplication \\

  \hline
    CNN-LSTM        & 100\% & 100\% & 100\% & 0.07\% & 0.38\% & 0.10\% \\
    MLP             & 100\% & 100\% & 100\%& 0.21\% & 0.21\% & 0.08\% \\
    CNN-MLP         & 100\% & 100\% & 100\% & 96.33\% & 98.69\% & 0.07\% \\
    Autoencoder     & 100\% & 100\% & 100\% & 96.78\% & 97.34\% & 0.08\% \\
    ResNet18        & 99.96\% & 98.52\% & 99.80\% & 99.86\% & 99.49\% & 0.10\% \\
    ResNet50        & 99.92\% & 99.86\% & 99.69\% & 99.14\% & 99.64\% & 0.10\% \\
    ResNet152       & 100\% & 100\% & 100\% & 98.74\% & 98.93\% & 0.14\% \\
  \bottomrule
\end{tabular}
\end{table*}

\begin{figure*}[t]
\begin{center}
	\subfigure[Bitwise And]{\label{fig:7a}\includegraphics[width=0.3\linewidth]{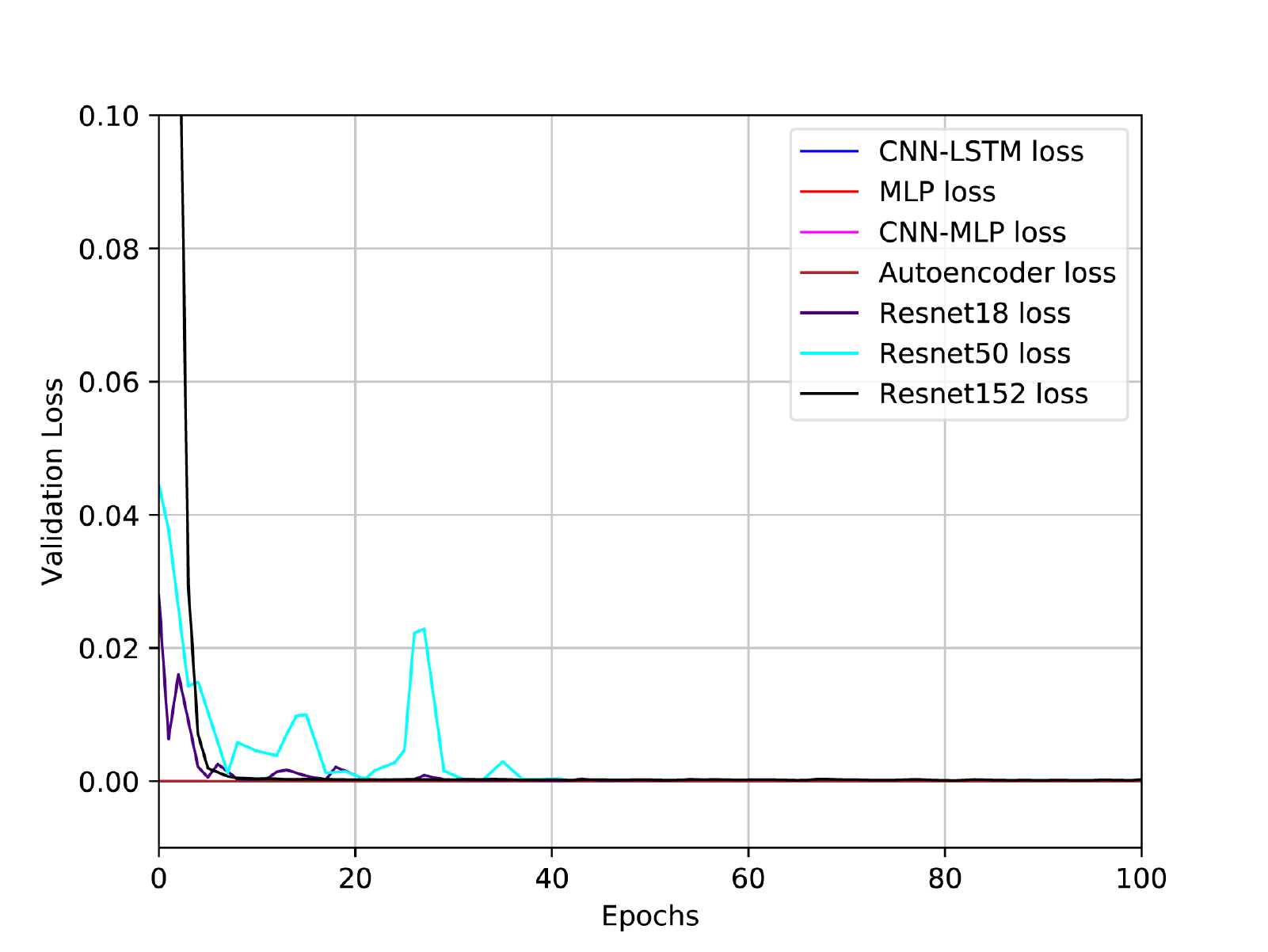}}
	\subfigure[Bitwise Or]{\label{fig:7b}\includegraphics[width=0.3\linewidth]{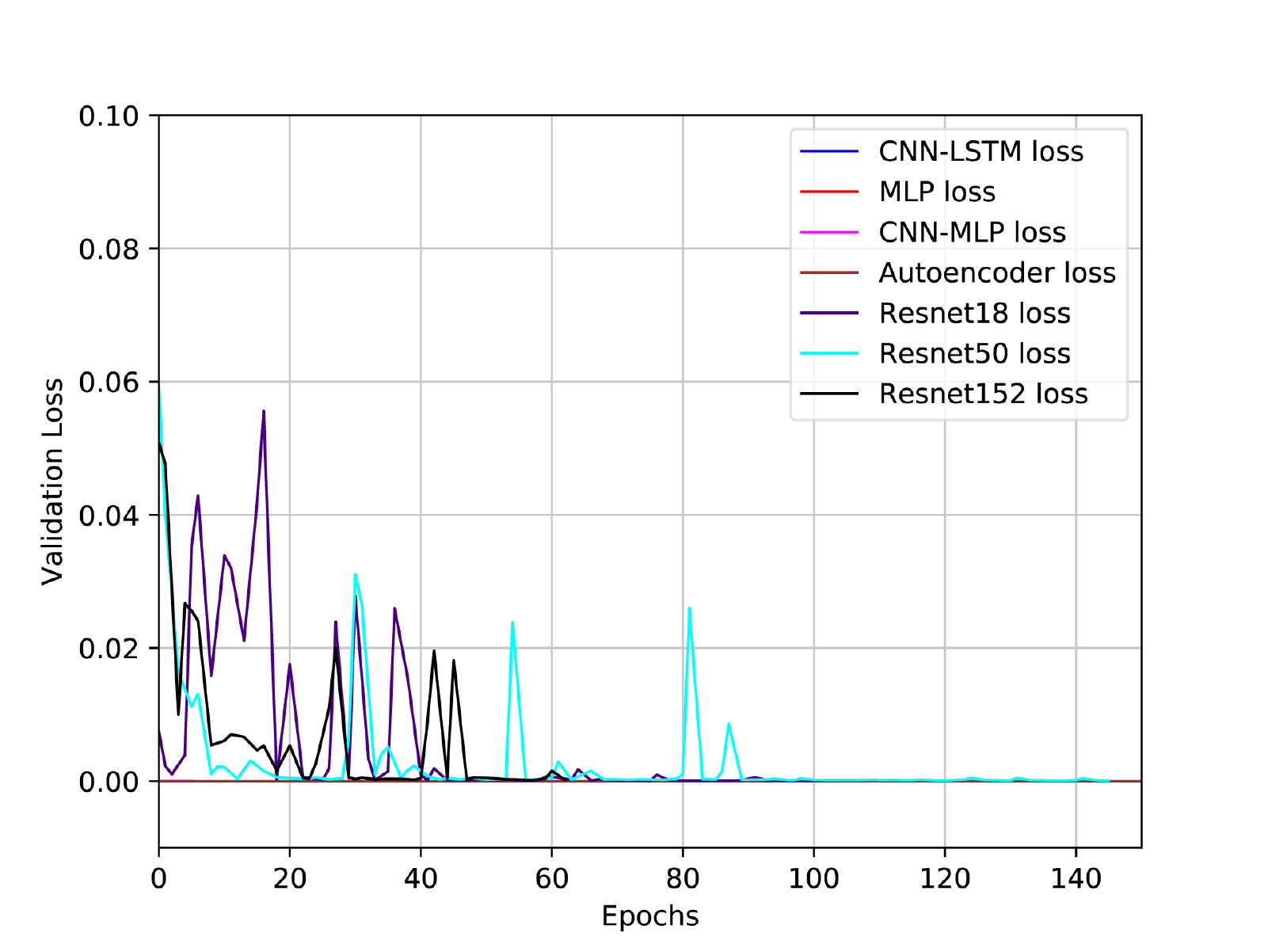}}
    \subfigure[Bitwise Xor]{\label{fig:7c}\includegraphics[width=0.3\linewidth]{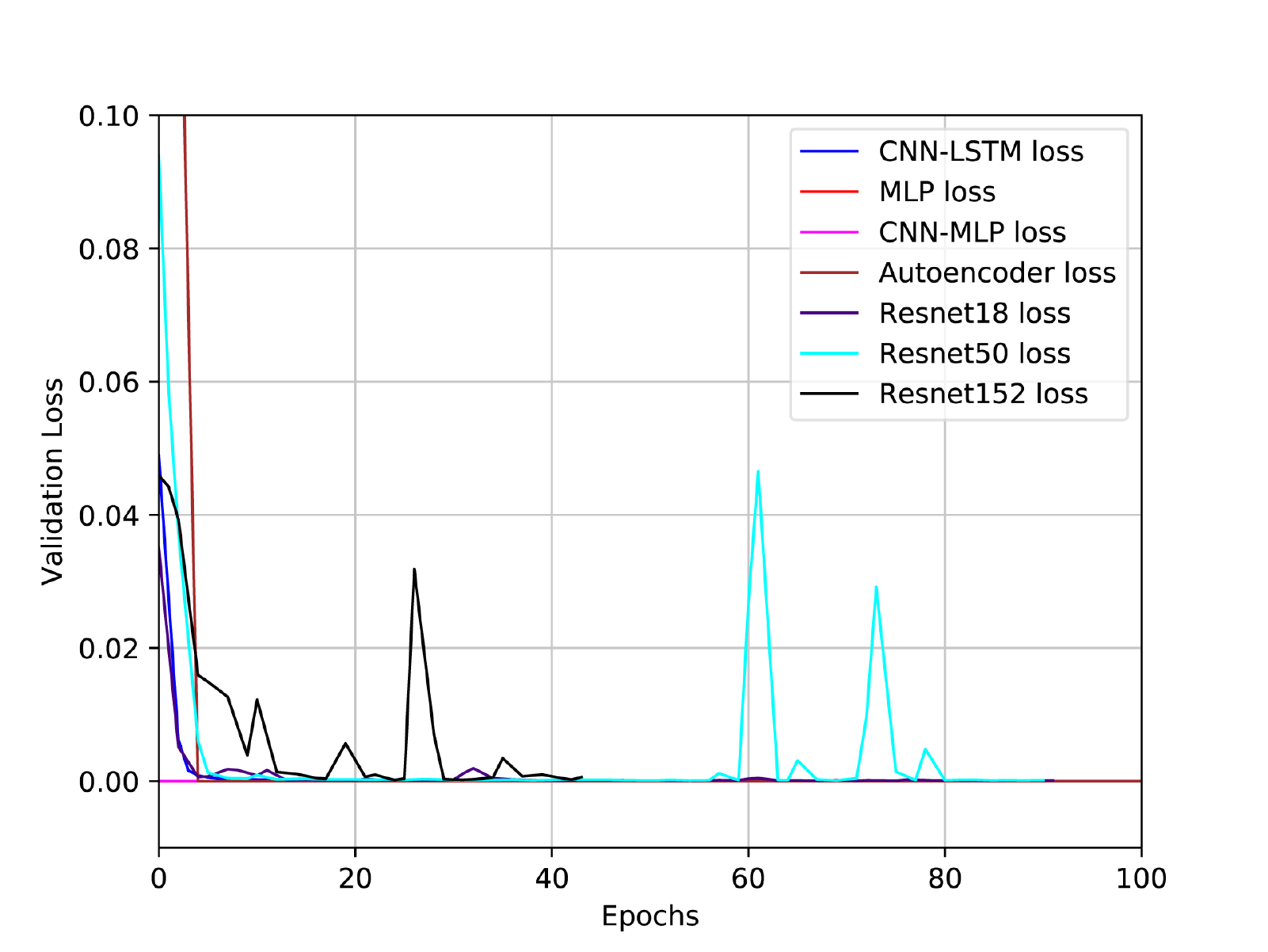}}
	\subfigure[Addition]{\label{fig:7d}\includegraphics[width=0.3\linewidth]{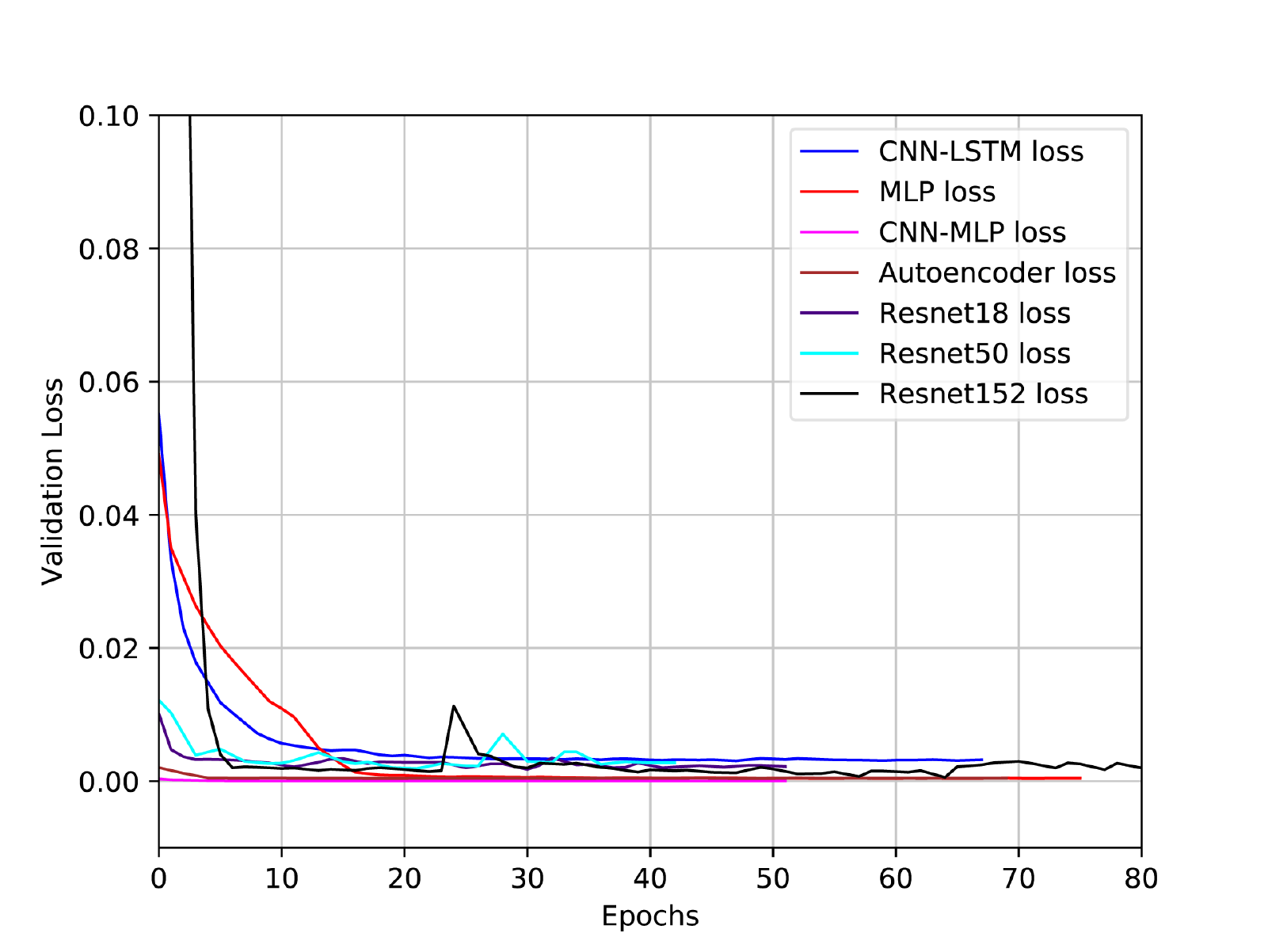}}
    \subfigure[Subtraction]{\label{fig:7e}\includegraphics[width=0.3\linewidth]{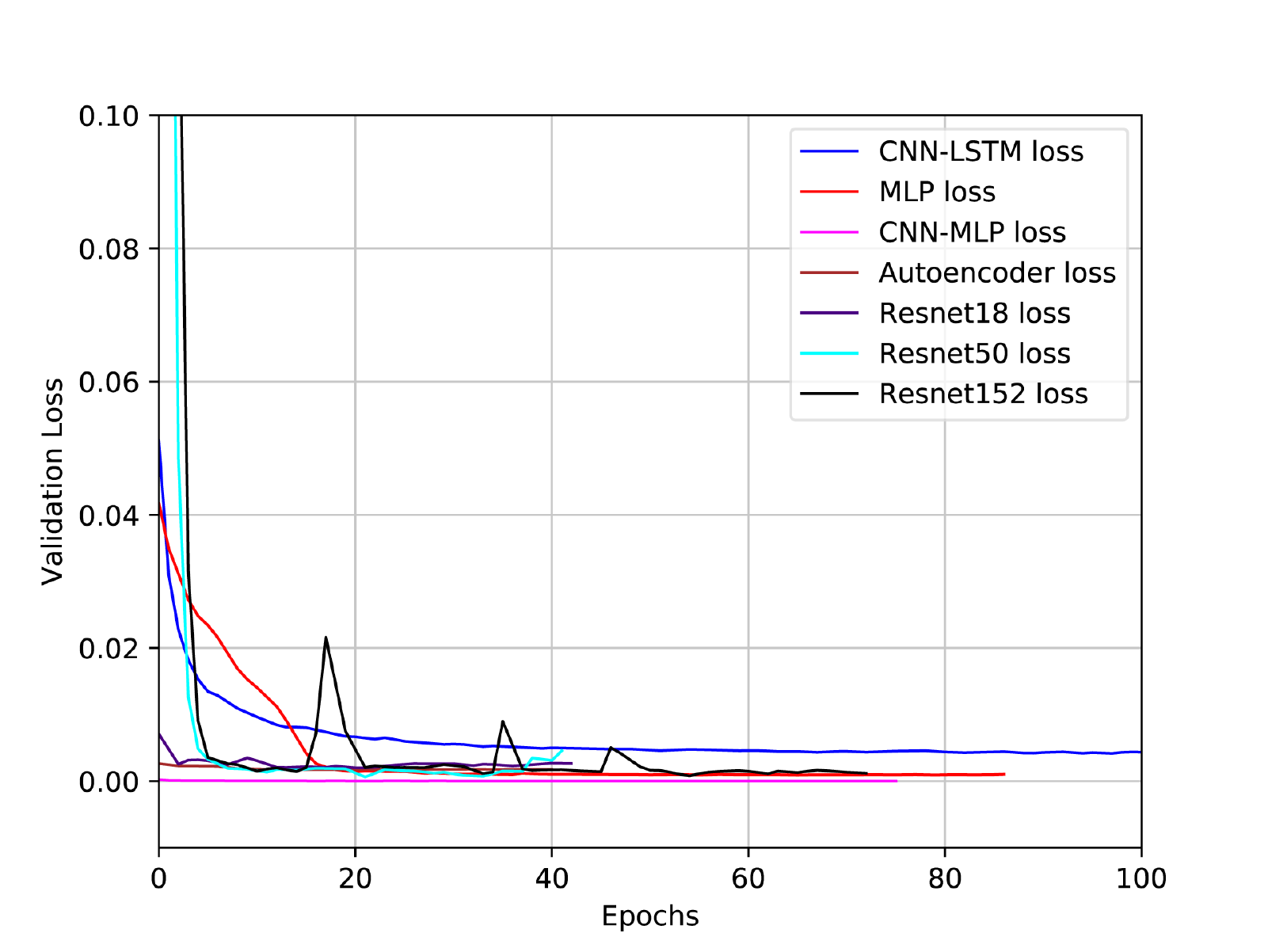}}
	\subfigure[Multiplication]{\label{fig:7f}\includegraphics[width=0.3\linewidth]{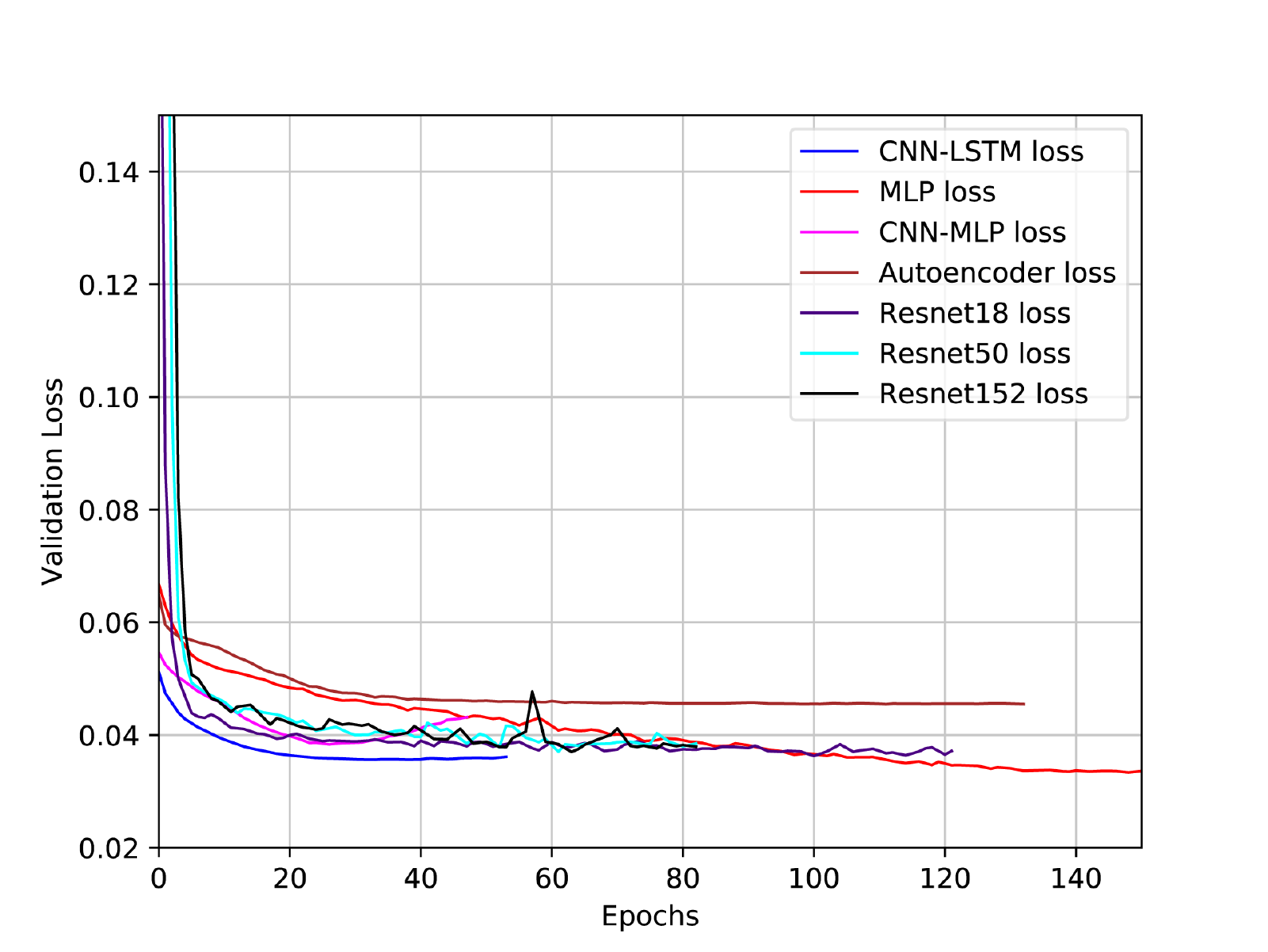}}
\end{center}
	\caption{The validation losses of Bitwise And, Bitwise Or, Bitwise Xor, Addition, Subtraction and Multiplication on 150,000 training data sets.}
	\label{fig:7}
\end{figure*}

The validation loss curves on Bitwise And, Bitwise Or and Bitwise Xor data sets
are shown in Fig.\ref{fig:6a}, Fig.\ref{fig:6b} and Fig.\ref{fig:6c}.
Because of the early-stopping, the epochs of these models are different.
From these figures, one finds that all models converge to small losses.
In addition, the MLP, CNN-MLP and Autoencoder converge faster than
the CNN-LSTM and ResNets.
The validation loss curves on Addition and Subtraction data sets
are shown in Fig.\ref{fig:6d} and Fig.\ref{fig:6e}.
From these figures, one observes that
the losses of the CNN-MLP, Autoencoder and ResNets are smaller than other models.
Moreover, both of CNN-MLP and Autoencoder converge faster than the ResNets.
The validation loss curve on Multiplication data set
is shown in Fig.\ref{fig:6f}.
One can see, from it, that all models have very large losses when they converge.

Next, we try to see if increasing data set size could improve model performances.
In this scene, all models are trained on 150,000 training data sets
and stopped when the losses on validation data sets no longer decrease.
The accuracies of all models on six LiLi data sets
are shown in Table~\ref{tab:15w_accuracy}.

From Table~\ref{tab:15w_accuracy}, one observes that
most models get the good performances on
Bitwise And, Bitwise Or, Bitwise Xor, Addition and Subtraction data sets.
It means the performances of models can be improved by increasing the size of data sets.
This provides a strategy to solve difficult logic learning problems.

\begin{table*}
  \caption{The test accuracies of Bitwise And, Bitwise Or, Bitwise Xor, Addition, Subtraction and Multiplication
  on 150,000 training data sets.}\label{tab:15w_accuracy}
  \centering
  \begin{tabular}{l|ccc|cc|c}
  \toprule
   \multirow{3}{*}{Model}  & \multicolumn{6}{c}{Operations}\\
  \cline{2-7}
               &\multicolumn{3}{c|}{$\star$}& \multicolumn{2}{c|}{$\star \star$}&  $\star \star \star$\\
  \cline{2-7}
               & Bitwise And & Bitwise Or & Bitwise Xor & Addition & Subtraction & Multiplication \\

  \hline
    CNN-LSTM        & 100\% & 100\% & 100\% & 84.21\% & 79.22\% & 0.20\% \\
    MLP             & 100\% & 100\% & 100\% & 98.79\% & 97.39\% & 0.16\% \\
    CNN-MLP         & 100\% & 100\% & 100\% & 99.96\% & 99.96\% & 0.35\% \\
    Autoencoder     & 100\% & 100\% & 100\% & 98.17\% & 98.66\% & 0.16\% \\
    ResNet18        & 100\% & 100\% & 100\% & 99.50\% & 99.50\% & 0.24\% \\
    ResNet50        & 100\% & 100\% & 100\% & 99.56\% & 99.79\% & 0.26\% \\
    ResNet152       & 100\% & 100\% & 100\% & 99.98\% & 99.87\% & 0.24\% \\
  \bottomrule
\end{tabular}
\end{table*}

The validation loss curves are shown in Fig.\ref{fig:7}.
From Fig.\ref{fig:7}, one observes that
the most of the models converge to smaller losses than before.
The validation loss curves on Bitwise And, Bitwise Or and Bitwise Xor data sets
are shown in Fig.\ref{fig:7a} to \ref{fig:7c}.
From these figures, one finds that the CNN-LSTM and ResNets converge faster than before.
The validation loss curves on Addition, Subtraction and Multiplication data sets
are shown in Fig.\ref{fig:7d}, Fig.\ref{fig:7e}
and Fig.\ref{fig:7f}, respectively.
From Fig.\ref{fig:7d} and Fig.\ref{fig:7e},
one observes that the losses of all models are smaller than before.
But, from Fig.\ref{fig:7f}, we observe that
all models still have very large losses when they converge on Multiplication data set.
A good phenomenon is that the losses of all models are smaller than before.

\begin{figure*}[t]
\begin{center}
    \subfigure[CNN-MLP]{\label{fig:8a}\includegraphics[width=0.95\linewidth]{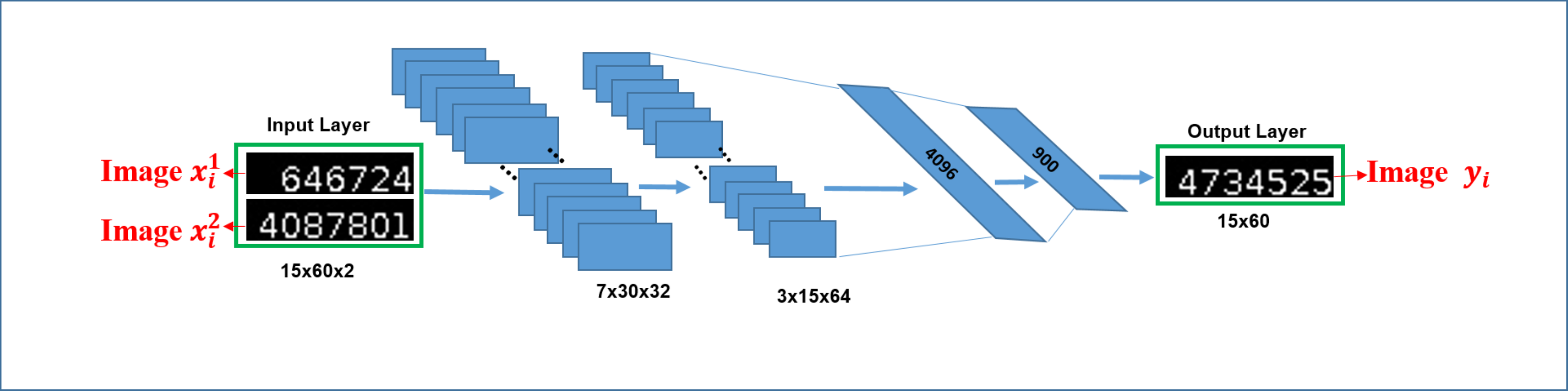}}
	\subfigure[CNN2-MLP]{\label{fig:8b}\includegraphics[width=0.95\linewidth]{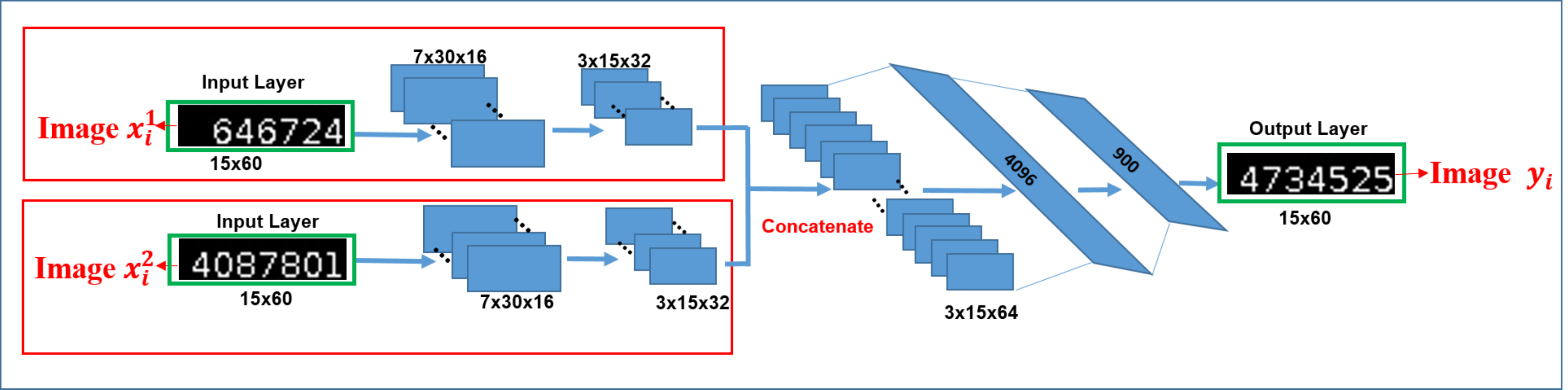}}
\end{center}
    \caption{The architectures of CNN-MLP and CNN2-MLP.}
\label{fig:8}
\end{figure*}

\begin{figure*}[t]
\begin{center}
	\subfigure[Bitwise And]{\label{fig:9a}\includegraphics[width=0.3\linewidth]{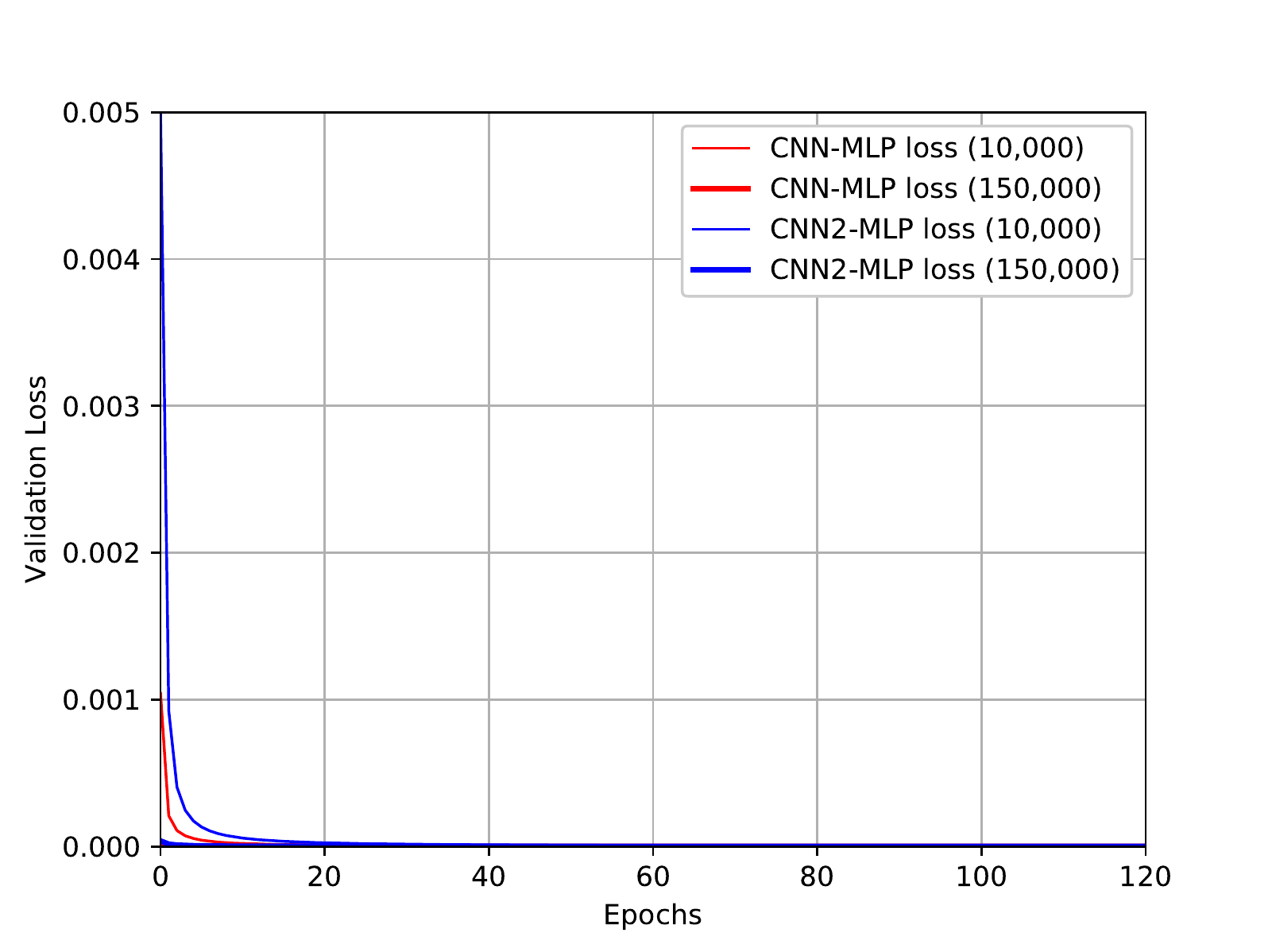}}
	\subfigure[Bitwise Or]{\label{fig:9b}\includegraphics[width=0.3\linewidth]{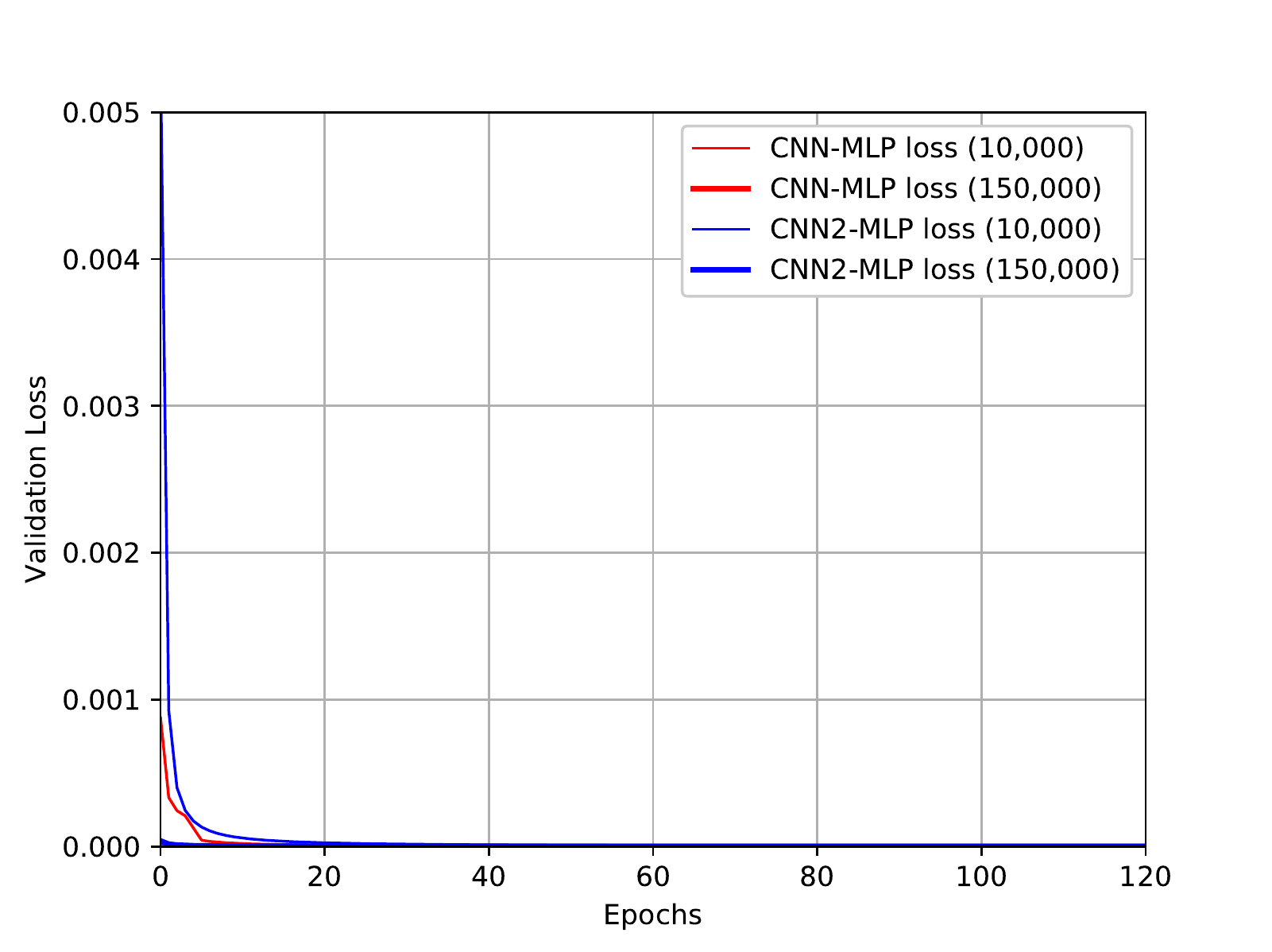}}
    \subfigure[Bitwise Xor]{\label{fig:9c}\includegraphics[width=0.3\linewidth]{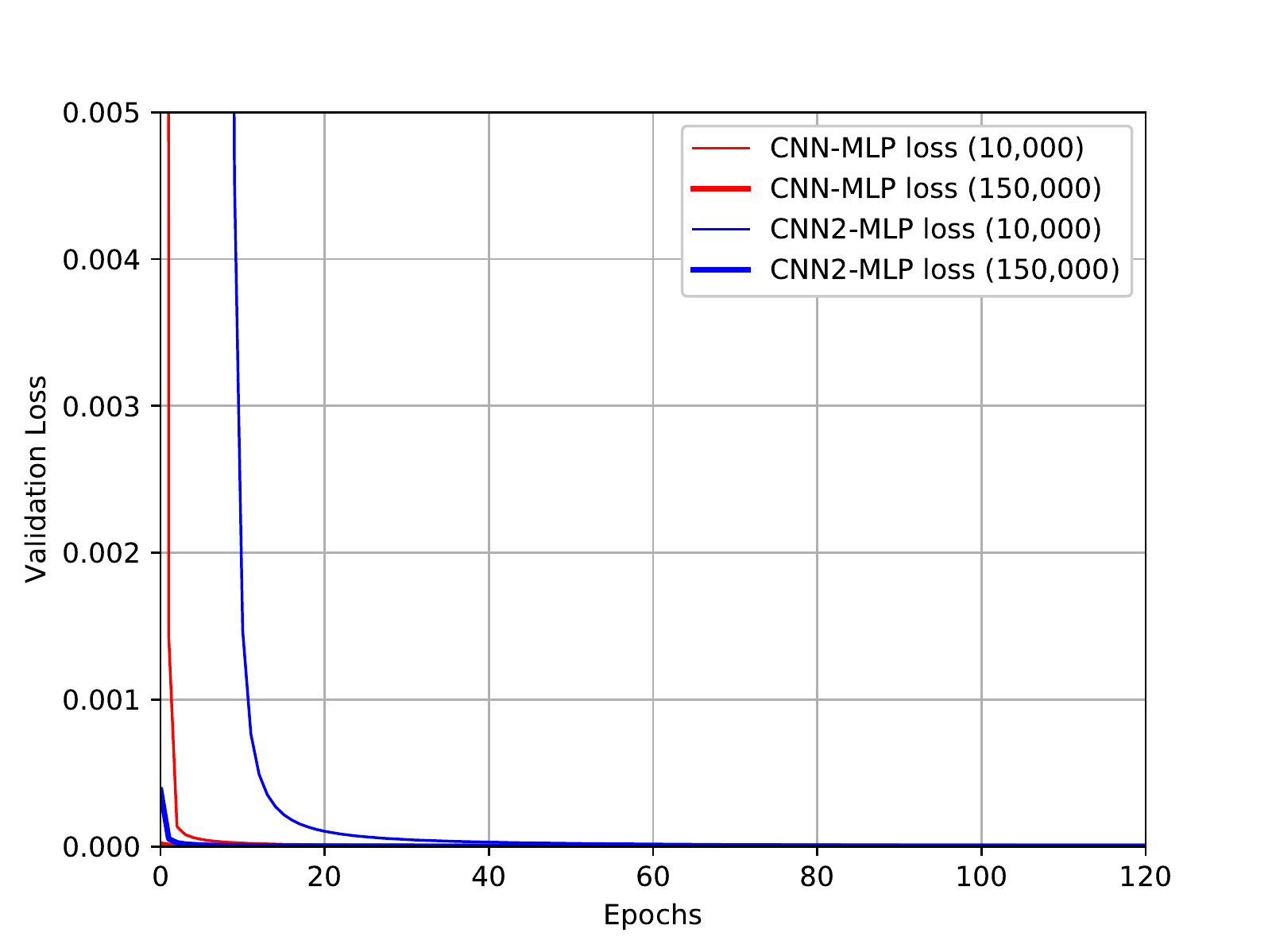}}
	\subfigure[Addition]{\label{fig:9d}\includegraphics[width=0.3\linewidth]{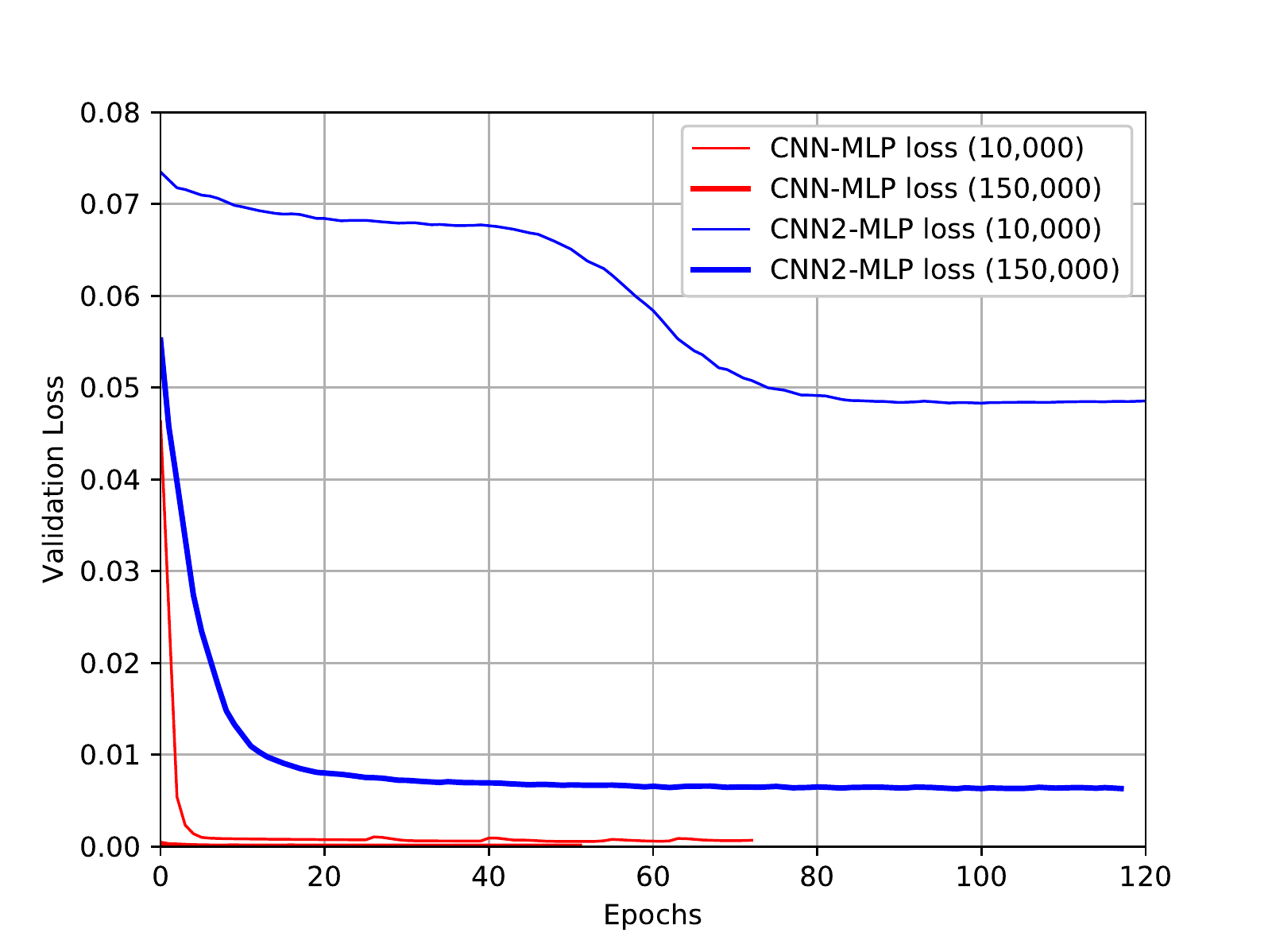}}
    \subfigure[Subtraction]{\label{fig:9e}\includegraphics[width=0.3\linewidth]{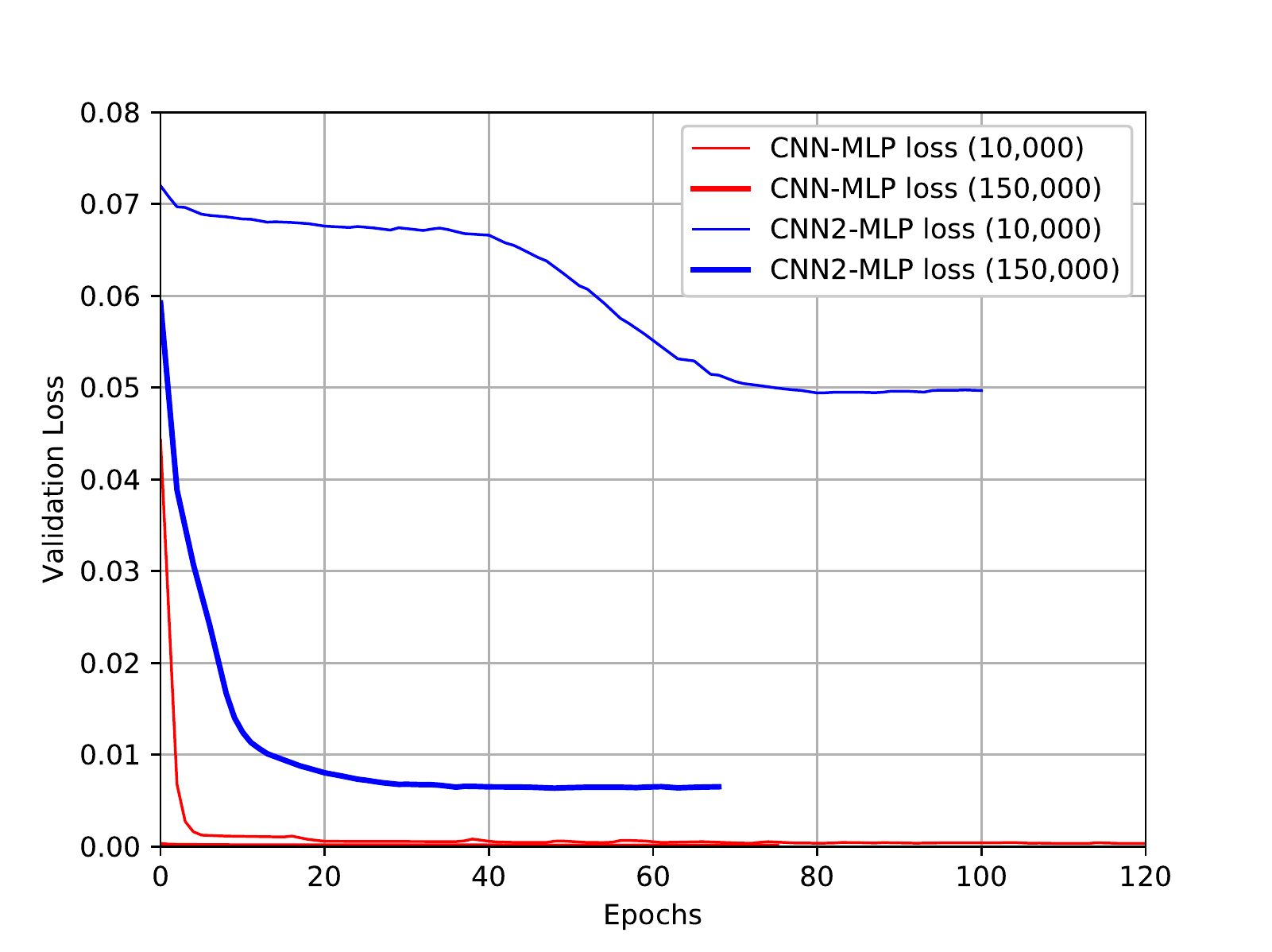}}
	\subfigure[Multiplication]{\label{fig:9f}\includegraphics[width=0.3\linewidth]{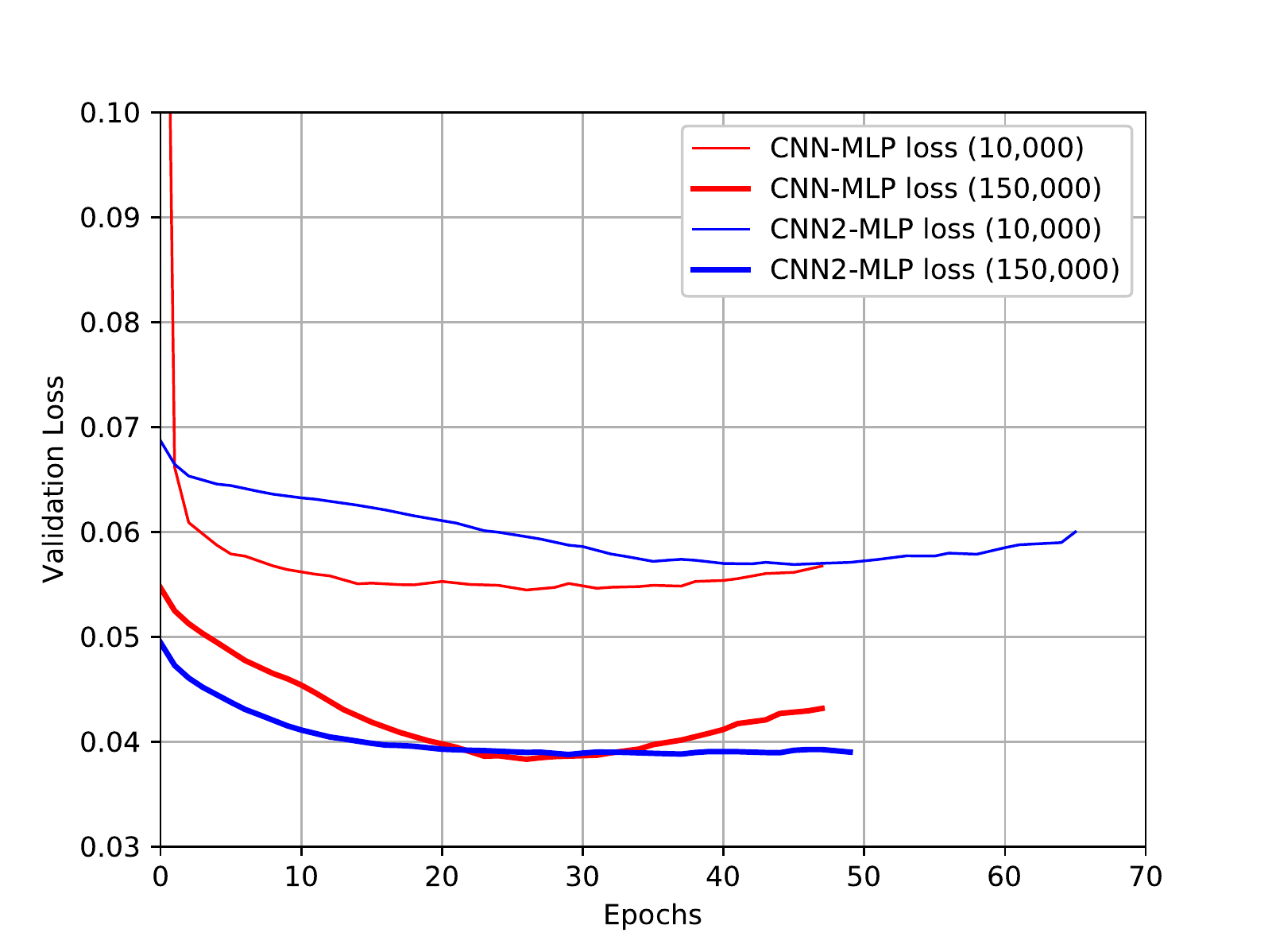}}
\end{center}
	\caption{The validation losses of CNN-MLP and CNN2-MLP on Bitwise And, Bitwise Or, Bitwise Xor, Addition, Subtraction and Multiplication data sets.}
	\label{fig:9}
\end{figure*}

\textbf{One guess: the space position
plays a significant role in the process of learning logical patterns.}
It is worth noting that the CNN-LSTM only gets about 80\% accuracies
on Addition and Subtraction data sets
even increasing the size of data sets.
However, it get 100\% accuracy on
Bitwise And, Bitwise Or and Bitwise Xor data sets.
The reason is that the CNN-LSTM is fed the input images one by one,
learn the features of the images separately so that
they almost do not consider the carry or borrow case
on addition or subtraction.
Each digit of the result of addition and subtraction is
affected by the adjacent positions (the influences from carry or borrow),
while each digit of the result of bitwise and, bitwise or and bitwise xor is not.
If the models want to get high accuracies, they should dispose 2 input images \emph{a} and \emph{b} simultaneously on Addition and Subtraction data sets.
In order to verify this idea, we develop a model called CNN2-MLP
that is similar to CNN-MLP.
These two models have same structure and hyper-parameter settings
except CNN2-MLP learns features of each of two input images separately.
And their structures are shown in Fig.\ref{fig:8}.

The validation loss curves of CNN-MLP and CNN2-MLP
on the six LiLi data sets are shown in Fig.~\ref{fig:9}.
For Bitwise And, Bitwise Or and Bitwise Xor data sets,
both of them converge to the small losses.
For Addition and Subtraction data sets,
the validation loss of CNN2-MLP is large on 10,000 training data sets.
When the size of training data set increasing,
the validation loss of CNN2-MLP is smaller than before but
still larger than the validation loss of CNN-NLP.
For Multiplication data set,
both of them converge to the large losses.
The test accuracies of CNN2-MLP on Bitwise And, Bitwise Or, Bitwise Xor, Addition, Subtraction and Multiplication
data sets are shown in Table~\ref{tab:CNN2_accuracy}.
CNN2-MLP can not get the good performances on Addition and Subtraction data sets,
but still work well on Bitwise And, Bitwise Or and Bitwise Xor data sets.
These experiment results verify that
the space position plays a significant role in the process of learning logical patterns.

\begin{table*}
  \caption{The test accuracies of CNN2-MLP on Bitwise And, Bitwise Or, Bitwise Xor, Addition, Subtraction and Multiplication data sets}
  \label{tab:CNN2_accuracy}
\centering
\begin{tabular}{l|ccc|cc|c}
  \toprule
   \multirow{3}{*}{\# training samples}  & \multicolumn{6}{c}{Operations}\\
  \cline{2-7}
               &\multicolumn{3}{c|}{$\star$}& \multicolumn{2}{c|}{$\star \star$}&  $\star \star \star$\\
  \cline{2-7}
               & Bitwise And & Bitwise Or & Bitwise Xor & Addition & Subtraction & Multiplication \\
  \hline
    150,000    & 100\% & 100\% & 100\% & 67.47\% & 62.92\% & 0.28\% \\
  \hline
    10,000     & 100\% & 100\% & 100\% & 0.24\% & 0.20\% & 0.05\% \\
  \bottomrule
\end{tabular}
\end{table*}

As the size of the given data increases, the MLP tends to have good performances
on Addition and Subtraction data sets.
This is because each digit of the result of the addition and subtraction is
affected by the adjacent positions in both input images.
In particular, for the MLP, the relation between two images at their arbitrary positions,
when data set size is small,
it can not focus on the exact relation on their adjacent positions.
As soon as the data set gets larger,
the defect can be made up.

From what has been discussed above, we can divide these models into three categories:
\begin{enumerate}[(1)]
  \item CNN-LSTM: This model is appropriate for this type of task
  where each digit of the result is only affected
  by the same position of the input numbers
  (e.g. Bitwise And, Bitwise Or and Bitwise Xor data sets).

  \item MLP: The model is appropriate for this type of task
  where each digit of the result is affected
  by all the positions of the input numbers
  (MLP is more appropriate than other models on Multiplication data sets).
  If the size of data set is large enough, MLP can focus on
  the same or adjacent positions of the input numbers
  (e.g. Bitwise And, Bitwise Or, Bitwise Xor, Addition and Subtraction data sets).

  \item CNN-MLP, Autoencoder and ResNets: These models are appropriate for this type of task
  where each digit of the result is affected by the same or adjacent positions of the input numbers (e.g. Bitwise And, Bitwise Or, Bitwise Xor, Addition and Subtraction data sets).
\end{enumerate}

Next, from the standpoint of the visual effects,
these models are compared.
These predicted results output by the models with the poor performances
are shown.
For Addition and Subtraction data sets,
only the CNN-LSTM and MLP get the poor performances;
for Multiplication data sets,
all models get the poor performances.

\begin{figure*}[t]
\begin{center}
	\subfigure[10,000 training data set]{\label{fig:10a}\includegraphics[width=0.45\linewidth]{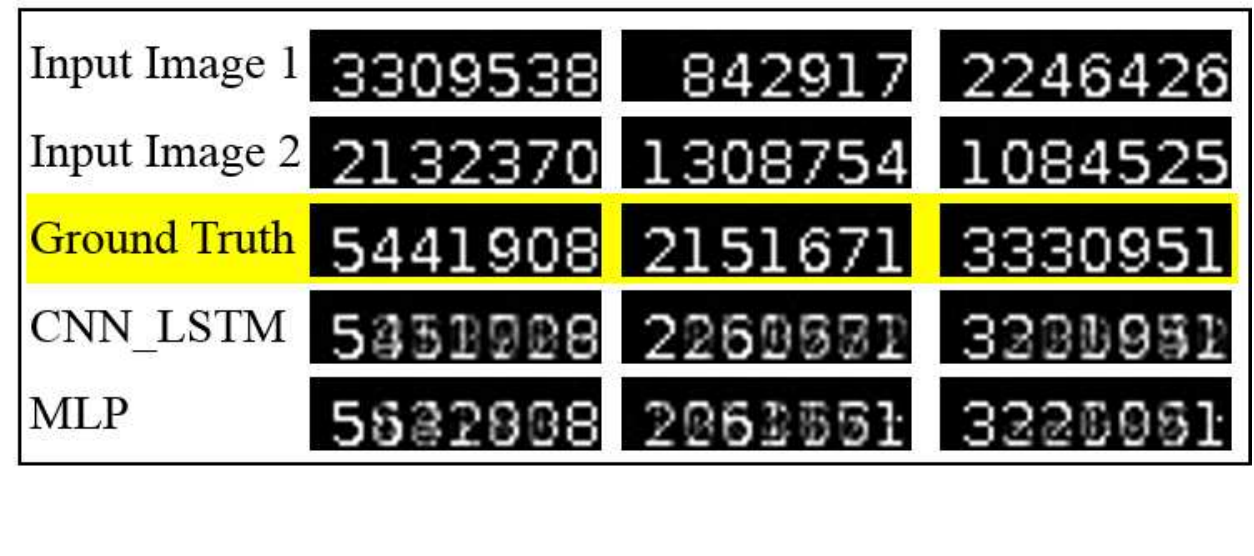}}
	\subfigure[150,000 training data set]{\label{fig:10b}\includegraphics[width=0.45\linewidth]{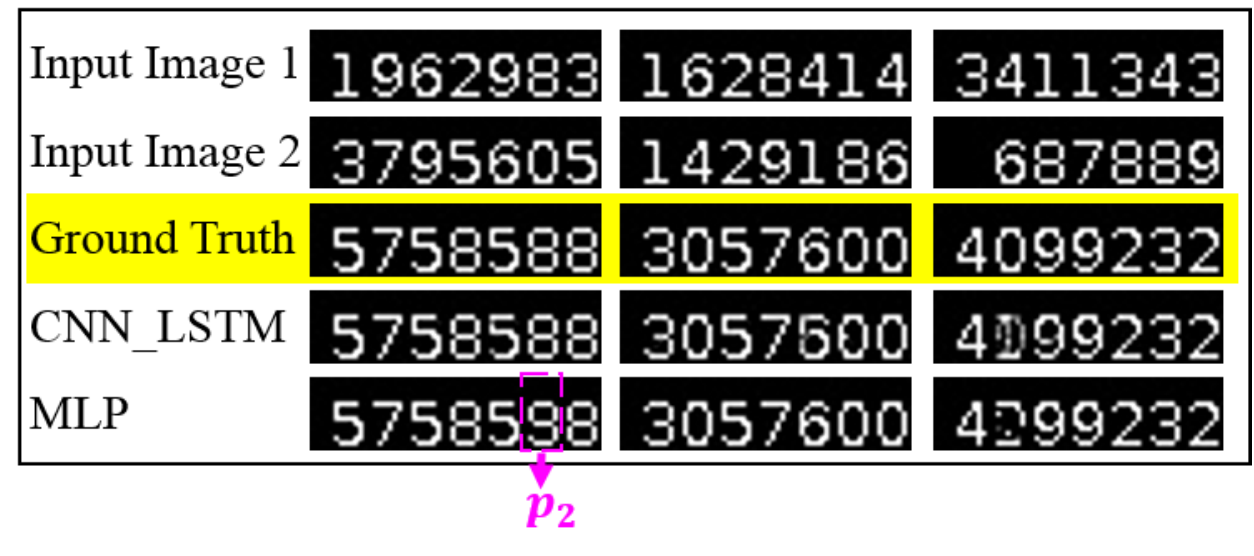}}
\end{center}
	\caption{The test visual effects of Addition on 10,000 training data set and 150,000 training data set.}
	\label{fig:10}
\end{figure*}

\begin{figure*}[t]
\begin{center}
	\subfigure[10,000 training data set]{\label{fig:11a}\includegraphics[width=0.45\linewidth]{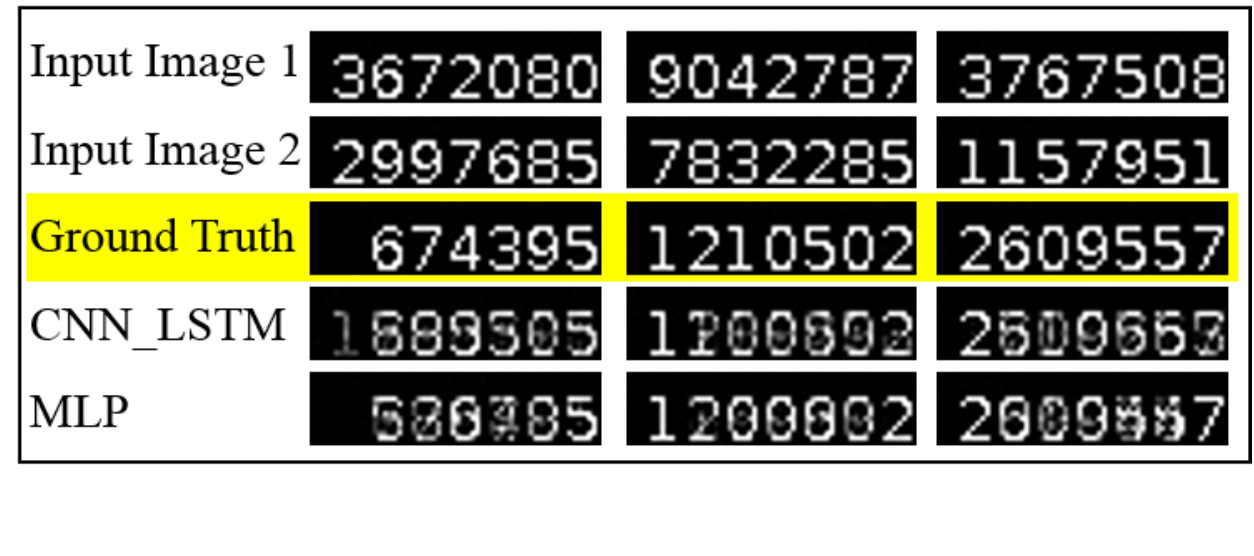}}
	\subfigure[150,000 training data set]{\label{fig:11b}\includegraphics[width=0.45\linewidth]{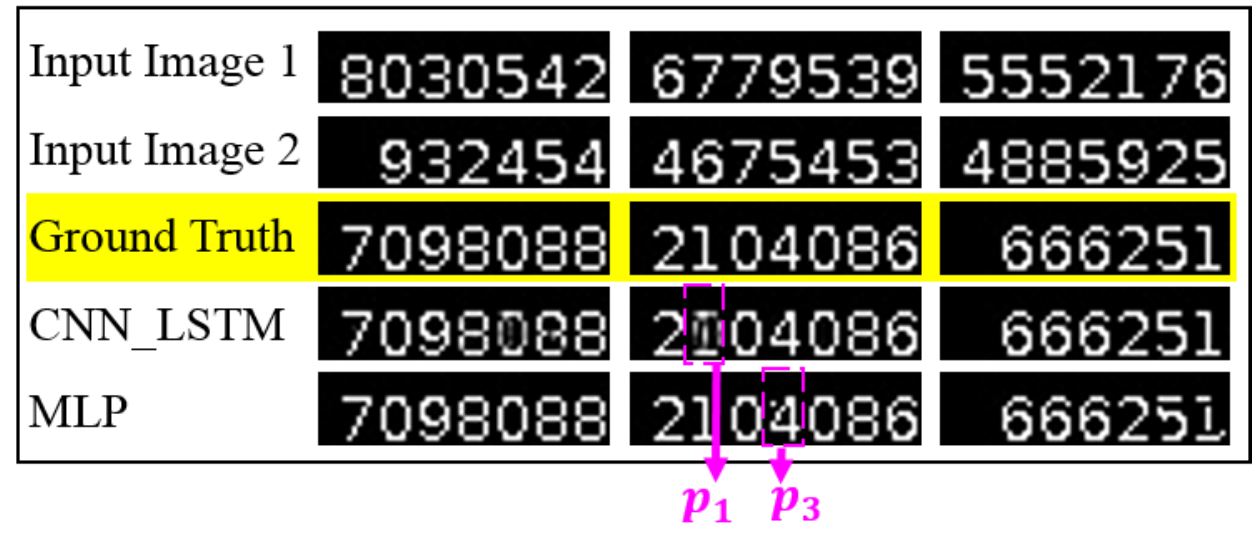}}
\end{center}
	\caption{The test visual effects of Subtraction on 10,000 training data set and 150,000 training data set.}
	\label{fig:11}
\end{figure*}

\begin{figure*}[t]
\begin{center}
	\subfigure[10,000 training data set]{\label{fig:12a}\includegraphics[width=0.45\linewidth]{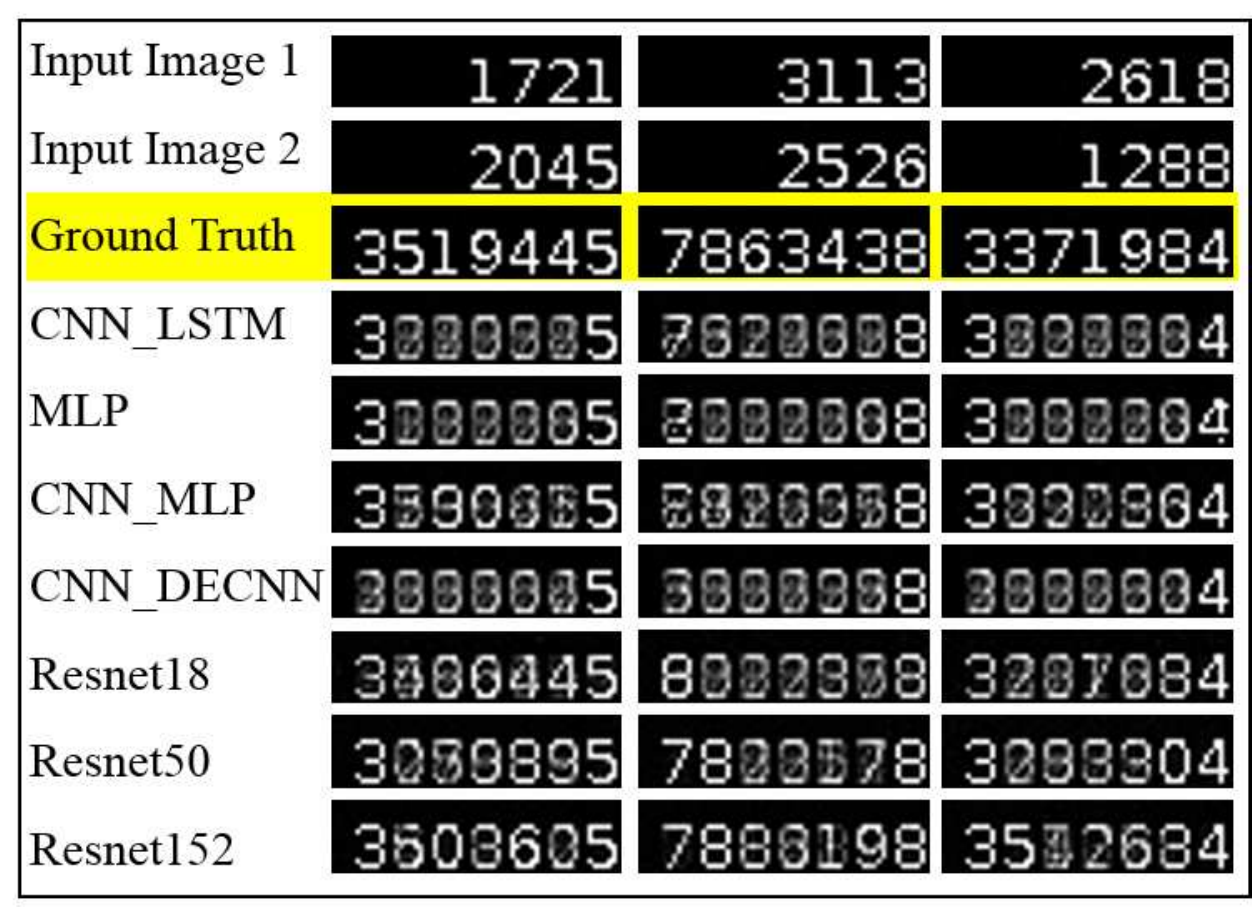}}
	\subfigure[150,000 training data set]{\label{fig:12b}\includegraphics[width=0.45\linewidth]{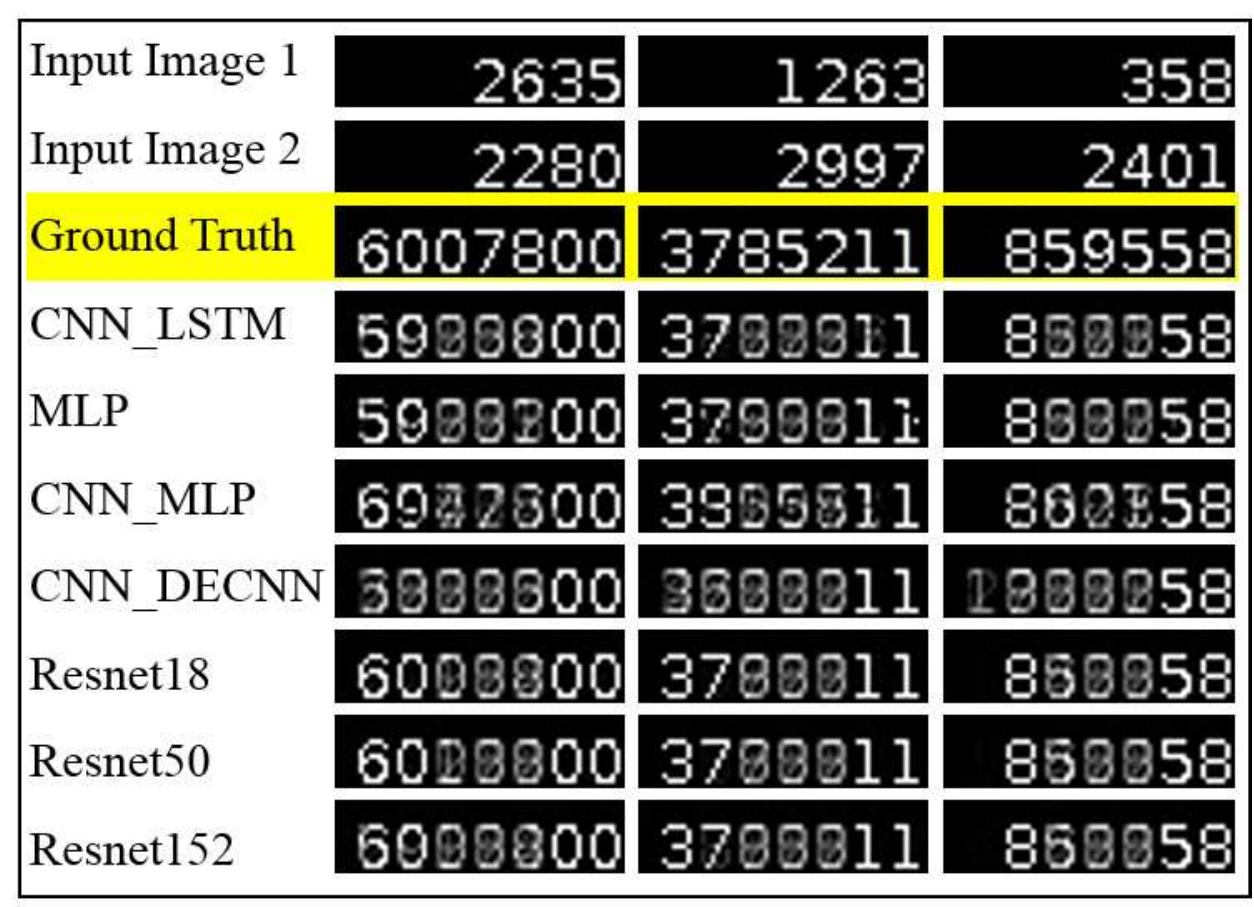}}
\end{center}
	\caption{The test visual effects of Multiplication on 10,000 training data set and 150,000 training data set.}
	\label{fig:12}
\end{figure*}

For Addition and Subtraction data sets,
the visual effects are shown in Fig.\ref{fig:10} and Fig.\ref{fig:11}.
From Fig.\ref{fig:10a} and Fig.\ref{fig:11a},
one observes that most models can clearly learn the first and last digits,
and other digits obscurely in output images.
As the size of the training data set increases,
from Fig.\ref{fig:10b} and Fig.\ref{fig:11b},
one observes that most models can clearly learn most digits
in output images.
For Multiplication data set,
the visual effects are shown in Fig.\ref{fig:12}.
From Fig.\ref{fig:12a},
we observe that most models can only clearly learn the first and last digits
and other digits obscurely in output images.
As the size of the training data set increases, from Fig.\ref{fig:12b},
one sees that most models can clearly learn more digits than before,
but still obscurely for most digits in output images.
There are many reasons why the performances of the predicted result on the digits is poor.
Some predicted digits are very obscure, e.g. the $p_{1}$ is shown
in Fig.\ref{fig:11b}).
Some are similar to other digits, e.g.
the $p_{2}$ is shown in Fig.\ref{fig:10b}).
Some are right but OCR can not recognize them, e.g. the $p_{3}$ is shown in Fig.\ref{fig:11b}.
Hence the accuracies can be higher in fact.

From above experimental results, one observes that
these models can not solve
the difficult LiLi task: Multiplication.
In the next section, an effective solution is provided
by dividing this task into a few easier subtasks.

\section{Divide and conquer model for Multiplication data set} \label{sec:DCM}

Although increasing the size of data set has effects on
solving the difficult logic learning problems,
all models still get the poor performances on Multiplication data set.
To our knowledge, many problems are complex and difficult to solve directly,
but it becomes easier when decomposed \cite{Qian2010MGRS, Ke2014Hybridization, NIPS20166285, Qian2010Positive, Tan2019Granulation}.
Artificial algorithm decomposition can effectively reduce the difficulty of learning \cite{Chen2020Edge}.
Inspired by this, we propose the DCM to address complex task
adopting the decomposition strategy.

We decompose a complex task into $k$ subtasks through the DCM,
and the decomposition criterion is that the combination difficulty of subtasks
is lower than the complex task.

\begin{equation}\label{F:DCMdecompose}
  H > f(h_{1}, h_{2}, ..., h_{k}),
\end{equation}
where $H$ is the difficulty of this complex task,
$h_{i}$ is the difficulty of the $i^{th}$ subtask,
$f$ is the combination difficulty of subtasks and it is determined by all subtasks.

As one sees from Fig.\ref{fig:6f} and Fig.\ref{fig:7f},
the MLP is more robust and can converge to a smaller loss than other models.
For Multiplication,
the value at a given position of E is determined by the values at the given
position in A and B and all positions in A and B before that given position.
MLP is exactly more appropriate this scene
than other models.
So we select the MLP as the decomposition module of the DCM.

\begin{figure*}[t]
\begin{center}
	\includegraphics[width=0.8\linewidth, height=0.27\linewidth]{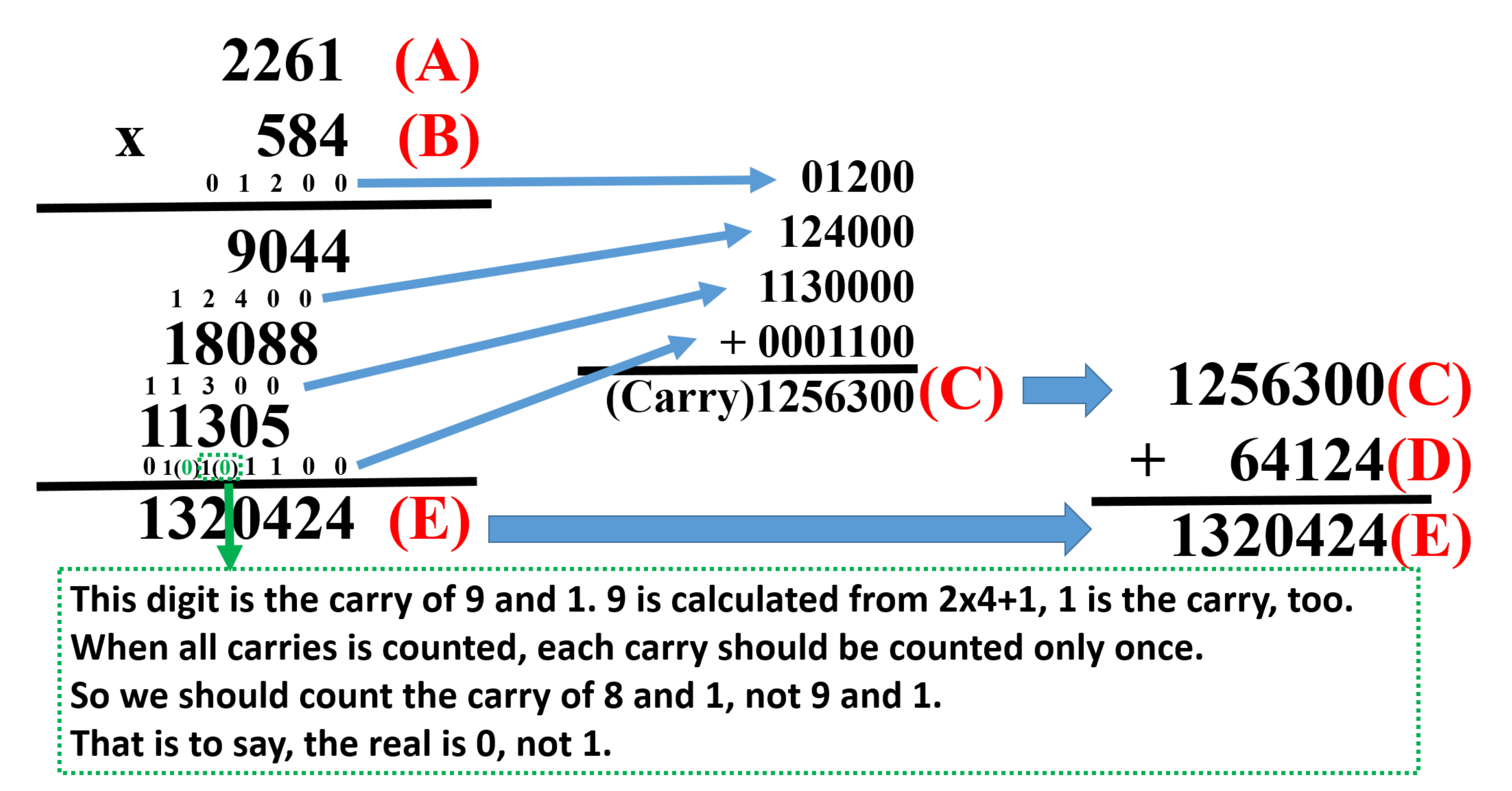}
\end{center}
	\caption{The procedure of multiplication.}
	\label{fig:13}
\end{figure*}

In this experiment, Multiplication data set
is regenerated by adding some information.
For training set, each of these samples consists of 4 input
images each containing a single integer number.
The input images are marked \emph{a}, \emph{b}, \emph{c} and \emph{d}.
The output image marked \emph{e}
is generated by the result of the multiplication operation.
The numbers embedded in images $a,b,c,d$ and $e$
are $A,B,C,D$ and $E$.
For testing set, only generate image $a,b$ and $e$.
For per sample, the ranges of \emph{A} and \emph{B} are 0$\sim$3160.
\emph{E} is the product of \emph{A} and \emph{B}.
The carry operation occurs when the product of two numbers on one digit
is more than ten,
and \emph{C} is used to record the value of carry part,
while \emph{D} is used to record the value of non-carry part.
So, the multiplication is divided into the carry part and non-carry part,
in other words, the sum of \emph{C} and \emph{D} is equal to \emph{E}.
For example, let \emph{A} and \emph{B} be  ``\emph{2261}'' and ``\emph{584}'', respectively,
and then, \emph{C}, \emph{D} and \emph{E}
equal to ``\emph{1256300}'', ``\emph{64124}'' and ``\emph{1320424}'', respectively.
The calculation procedure is shown in Fig.~\ref{fig:13}.

\begin{figure*}[t]
\begin{center}
	\subfigure[training procedure]{\label{fig:14a}\includegraphics[width=0.45\linewidth]{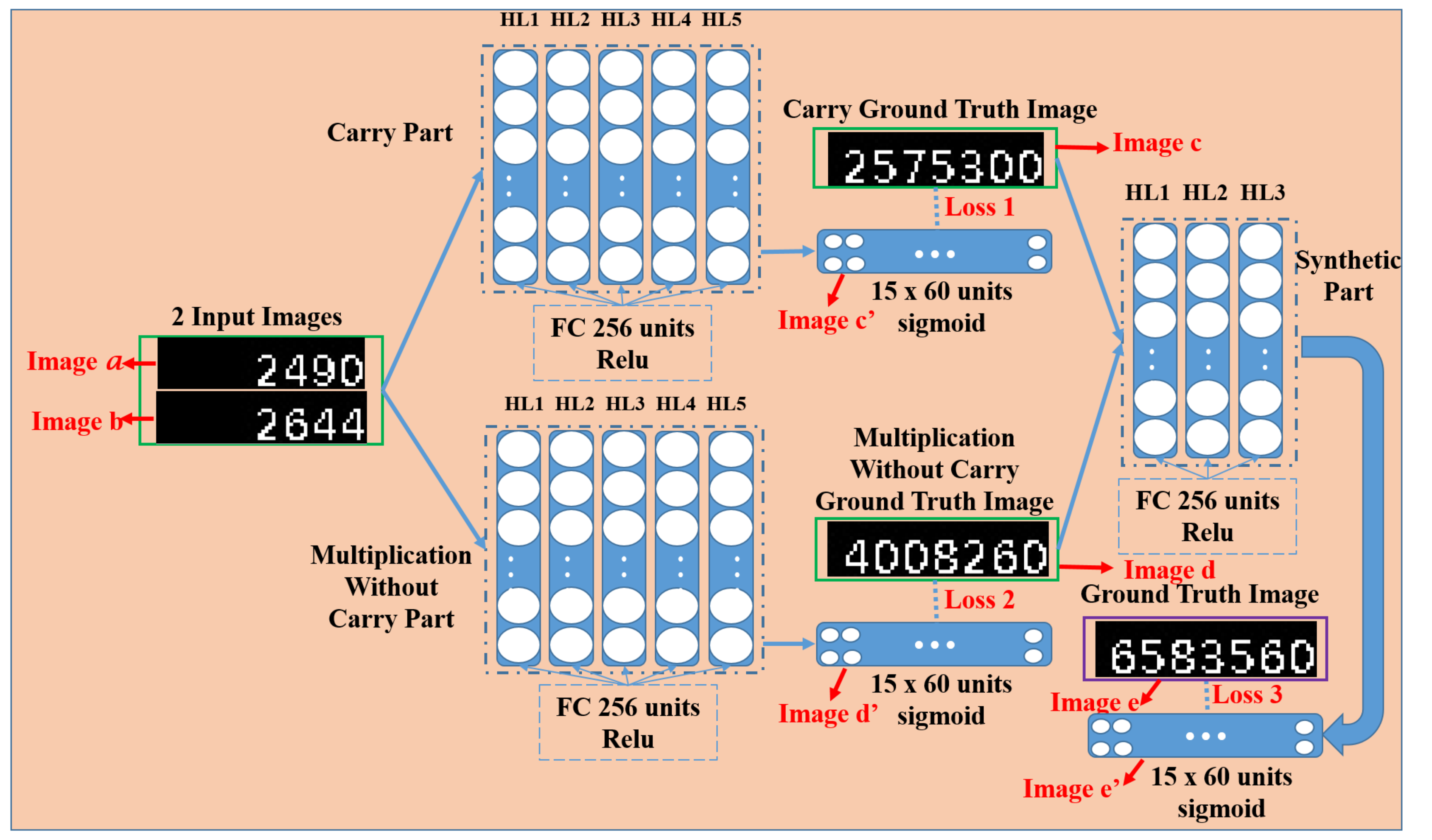}}
	\subfigure[testing procedure]{\label{fig:14b}\includegraphics[width=0.45\linewidth]{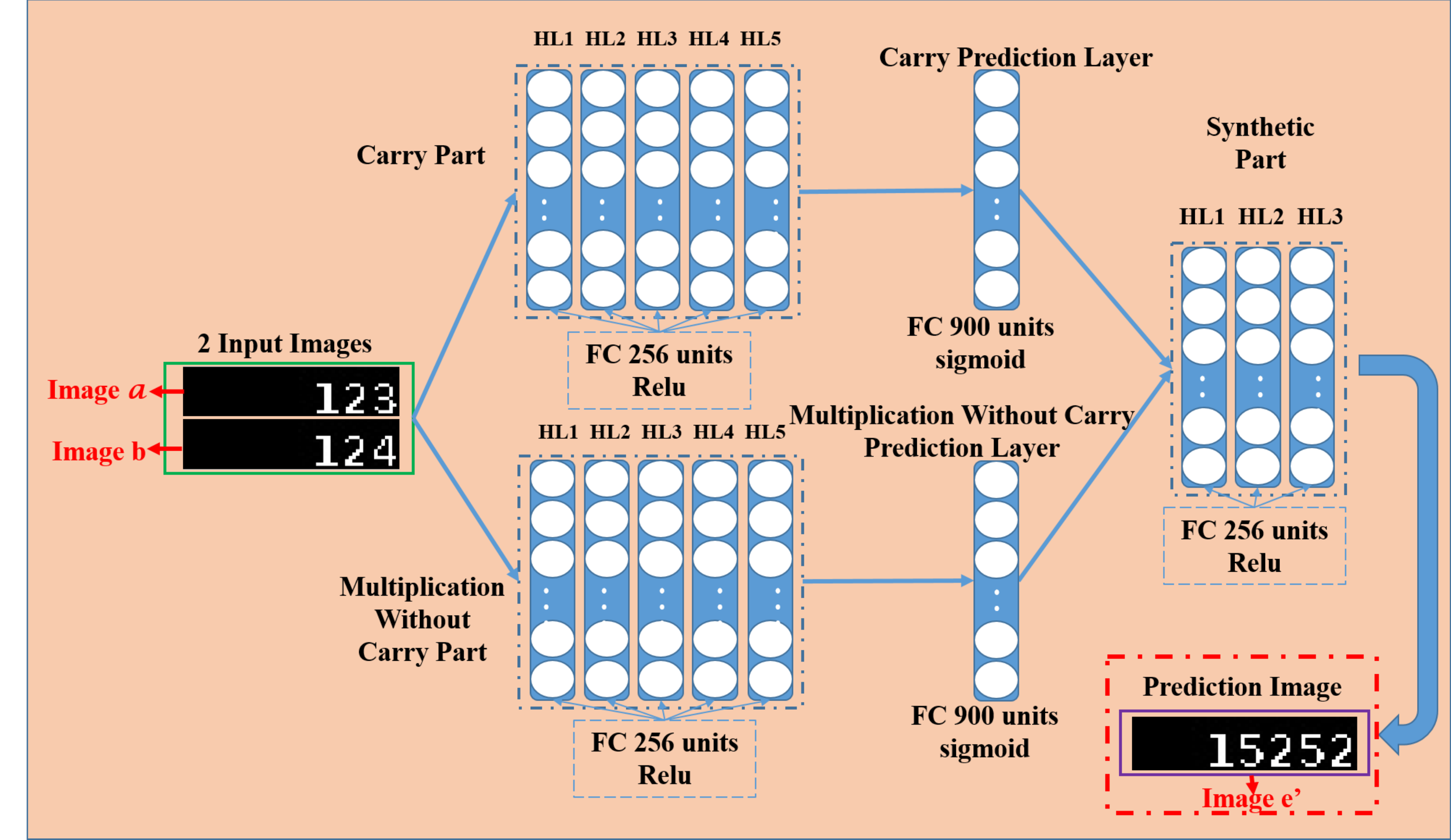}}
\end{center}
	\caption{Training and testing procedure.}
	\label{fig:14}
\end{figure*}

The DCM is divided into three subtasks:
carry subtask, non-carry subtask and synthetic subtask.
First, the carry subtask and non-carry subtask are used to
learn the carries of multiplication and multiplication without carry, respectively.
And then, the synthetic subtask is used to learn the synthetic pattern
of the carry subtask and non-carry subtask.
The network structures of these three subtasks are similar,
but the network parameters are different.
\begin{enumerate}[1)]
  \item \emph{Carry subtask}:
  During training, the images \emph{a} and \emph{b} are used as the input,
  image \emph{c} as the ground-truth result.
  The network of the carry subtask is fully-connected layers
  and uses the ReLU as the activation functions in the hidden layers
  and the sigmoid in the output layer.
  The carry subtask has 5 hidden layers,
  and each layer has 256 units.

  \item \emph{Non-carry subtask}:
  During training, the images \emph{a} and \emph{b} are used as the input,
  the image \emph{d} as the ground-truth result.
  The network of the non-carry subtask is fully-connected layers
  and uses the ReLU as the activation functions in the hidden layers
  and the sigmoid in the output layer.
  The non-carry subtask has 5 hidden layers,
  and each layer has 256 units.

  \item \emph{Synthetic subtask}:
  During training, the images \emph{c} and \emph{d} are used as the input,
  the image \emph{e} as the ground-truth result.
  The network of the synthetic subtask is fully-connected layers
  and uses the ReLU as the activation functions in the hidden layers
  and the sigmoid in the output layer.
  The synthetic subtask has 3 hidden layers, and each layer has 256 units.
\end{enumerate}

The ground-truth image is named as \emph{x} (\emph{x} can be \emph{c}, \emph{d} and \emph{e}),
and the predicted image is named as \emph{x'}.
We hope the number embedded in predicted image \emph{e'}
is equal to the number embedded in ground truth image \emph{e}, i.e., \emph{E'} = \emph{E}.

\begin{enumerate}[a)]
  \item \emph{Training}:
  During training procedure, the images \emph{a} and \emph{b} are used as the input,
  \emph{e} as the ground truth result and \emph{e'} as the output.
  It is interesting that the images \emph{c} and \emph{d} are both the input and ground truth results.
  For the carry subtask and non-carry subtask,
  the images \emph{c} and \emph{d} are the ground truth images, however,
  for the synthetic subtask, the image \emph{c} and \emph{d} are the input images.
  Taking the multiplication formula ``\emph{2490 $\times$ 2644 = 6583560}'' for example
  explains the training procedure which is shown in Fig.~\ref{fig:14a}.
  \emph{A}, \emph{B}, \emph{C}, \emph{D} and \emph{E}
  are ``\emph{2490}'', ``\emph{2644}'', ``\emph{2575300}'', ``\emph{4008260}'' and ``\emph{6583560}'', respectively.
  The carry subtask, non-carry subtask and synthetic subtask are trained separately.
  For the carry subtask and non-carry subtask,
  the images \emph{a} and \emph{b} are used as the inputs,
  the images \emph{c} and \emph{d} as the ground truth images
  and the image \emph{c'} and \emph{d'} as the outputs, respectively.
  For the synthetic subtask, the images \emph{c} and \emph{d} are used as input,
  the image \emph{e} as the ground truth image and image \emph{e'} as output.
  The smaller the differences between predicted image \emph{c'}, \emph{d'} and \emph{e'}
  as well as ground-truth image \emph{c}, \emph{d} and \emph{e} are, the better the performance of DCM is.

  \item \emph{Testing}:
  In the testing procedure, DCM is an end-to-end model.
  We take the multiplication formula ``\emph{123 $\times$ 124 = 15252}'' for example
  to explain the testing procedure which is shown in Fig.~\ref{fig:14b}.
  \emph{A} and \emph{B} are ``\emph{123}'' and ``\emph{124}'', respectively.
  In the testing procedure, the DCM only takes images \emph{a} and \emph{b} as the inputs,
  and then directly gets a predicted image \emph{e'} at the output of the synthetic subtask.
  Specifically, the inputs are firstly passed through the carry subtask and non-carry subtask to get a carry prediction layer and a non-carry prediction layer, respectively.
  Then, the two prediction layers are concatenated and passed through the synthetic subtask to get the final prediction result \emph{E'}.
  \emph{E'} is ``\emph{15252}'' and  equals to \emph{E}
  which shows that the DCM correctly found the relation between the images \emph{a} and \emph{b}
  only using the pure visual information.
\end{enumerate}

The DCM is trained using the stochastic gradient descent with momentum 0.9,
optimising a mean square error (mse) loss
and batch size is fixed 256.
The learning rate starts with 0.8, and reduces slowly when the loss plateaus.
The training on the carry subtask, non-carry subtask
and synthetic subtask terminates when the loss no longer reduces.

\begin{table*}
\begin{center}
\caption{The test accuracy of each subtask of DCM using 150,000 training examples.}\label{tab:threeparts}
\begin{tabular}{l|c|c|c}
\toprule
\multirow{2}{*}{Operation} & \multicolumn{3}{c}{Network branches}\\
\cline{2-4}
 & Carry subtask & Operation without carry subtask & Synthetic subtask \\
\hline
     Multiplication & 86.25\% & 98.38\% & 84.46\%\\
\bottomrule
\end{tabular}
\end{center}
\end{table*}

\begin{figure*}[t]
\begin{center}
	\subfigure[Both get correct predicted images.]{\label{fig:15a}\includegraphics[width=0.3\linewidth]{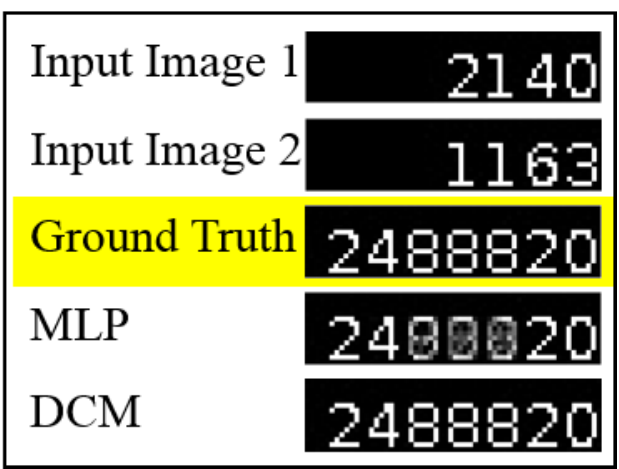}}
	\subfigure[The DCM gets correct predicted images, but MLP does not.]{\label{fig:15b}\includegraphics[width=0.3\linewidth]{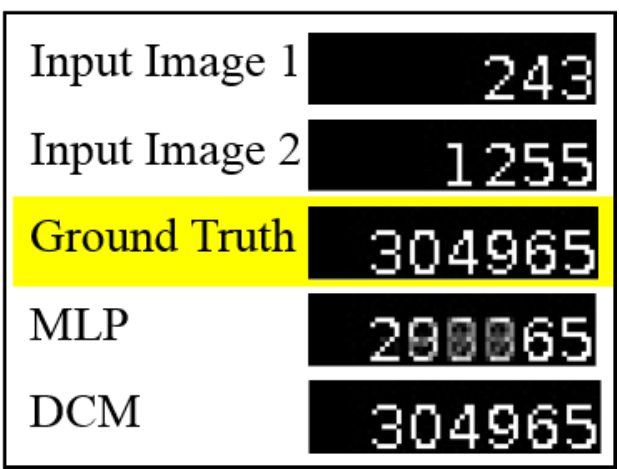}}
	\subfigure[Both get wrong predicted images.]{\label{fig:15c}\includegraphics[width=0.3\linewidth]{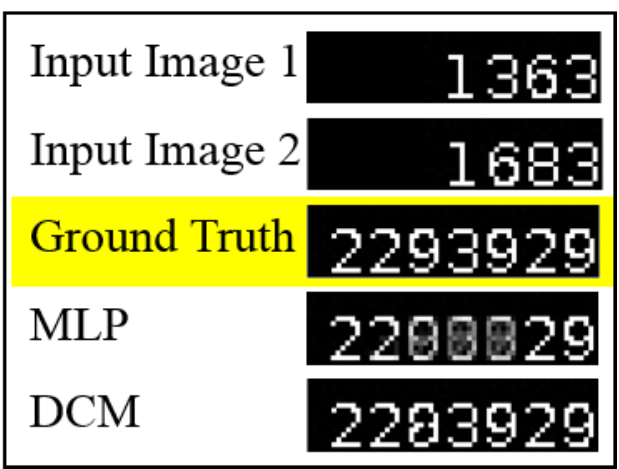}}
\end{center}
	\caption{The visual effects of multiplication task on 150,000 training set.}
	\label{fig:15}	
\end{figure*}

The accuracy of each subtask of DCM is shown in Table~\ref{tab:threeparts}.
In contrast, the DCM achieves the surprising accuracy $84.5\%$ which is higher than the MLP
on Multiplication data set.
Some visual effects from the testing are shown in Fig.~\ref{fig:15}.
In Fig.~\ref{fig:15a}, both DCM and MLP get correct predicted images.
In Fig.~\ref{fig:15b}, the DCM gets the correct predicted image, but the MLP does not.
In Fig.~\ref{fig:15c}, both DCM and MLP predict wrong images.
It can be seen that the last two digits and first two digits
in the image of the MLP are predicted correctly,
but the rest central 3 digits are uncertain.
However, for the DCM, only one digit of the number embedded in the predicted image is uncertain.
That is to say, the DCM can confirm more digits than the MLP.

This owns to the special structure of the DCM.
DCM divides a complex task into three simple subtasks,
carry subtask, non-carry subtask and synthetic subtask,
each subtask only learns one aspect of the task.
This helps to reduce uncertainty of each predicted digit embedded in the image \emph{e'}.
In order to explain the reason for the effectiveness of the DCM conveniently,
we employ some symbols in advance.
The goal of the visual logic learning of the arithmetic operations
is to compute the value of number 3 in a formula like ``number 1 operation number 2 = number 3''.
We call the digit of number n at the m$^{th}$ position (the rightmost position is 1$^{th}$) ``\emph{d$_n^{m}$}''.
The complexity of the task is determined by the degree of uncertainty (the amount of possibilities of each digit)
in the process of learning logical relation between the input images and output image.
For Addition,
``\emph{$d_3^m$}'' only has two possibilities, ``\emph{$(d_1^{m} + d_2^{m})mod10$}'' or ``\emph{$(d_1^{m} + d_2^{m} + 1)mod10$}''.
The case of ``\emph{$d_3^m$}'' on Subtraction is similar to addition.
However, the degree of uncertainty of multiplication is stronger than that of addition and subtraction, where ``\emph{$d_3^m$}'' has ten possibilities.

We assume a formula such as ``\emph{$d_1^2d_1^1 \times d_2^2d_2^1 = d_3^4d_3^3d_3^2d_3^1$}''
or ``\emph{$d_1^2d_1^1 \times d_2^2d_2^1 = d_3^3d_3^2d_3^1$}'' (if \emph{$d_3^4$}=0).
The scope of each digit ``\emph{$d_3^m$}'' (except the digit at rightmost position)
is very big, the digit at rightmost position is always an unique
and determined value ``\emph{$(d_1^1 \times d_2^1)mod10$}''.
The DCM can reduce the degree of uncertainty of predicted number 3.
For example, ``\emph{$d_3^2$}'' is determined
by the carry and non-carry part during multiplication.
In the MLP, the scope of ``\emph{$d_3^2$}'' is 0$\sim$9,
and the scope of the carry at the 2$^{th}$ position is 0$\sim$8.
So the carry at the 2$^{th}$ position is to choose one value in 0$\sim$8 out of the range 0$\sim$9.
The non-carry at the 2$^{th}$ position is to choose one value in 0$\sim$9.
So, there are 900 possibilities (\emph{$C_{10}^{9}C_9^{1}C_{10}^{1}$}) for ``\emph{$d_3^2$}'' in fact.
In the MLP, ``\emph{$d_3^2$}'' is directly computed.
In contrast, our method is first to compute carry and non-carry respectively,
and then synthetic these two subtasks.
The scope of the carry at the 2$^{th}$ position is 0$\sim$8,
so the carry at the 2$^{th}$ position only needs to determine which one is right in 0$\sim$8.
The non-carry at the 2$^{th}$ position is to choose one of 0$\sim$9.
Hence, there are 90 possibilities (\emph{$C_9^{1}C_{10}^{1}$}) for ``\emph{$d_3^2$}''.
The DCM largely reduces the number of possible values from 900 to 90.
Therefore,
the DCM confirms more digits than that of the MLP,
when the predictions of two models are all wrong.

\section{Conclusion}\label{sec:conclusion}

In this study, we have explored an interesting and important
research topic: can logic reasoning patterns be directly learned from given data?
As a preliminary exploration, the topic has been investigated
through a called LiLi task:
directly learning logic from a training image set.
In this work, many typical neural network models
have been used to solve the LiLi task
with the good performances on easy and intermediate logic data sets.
In order to further solve the difficult task,
a new network framework called DCM has been developed
using a decompose strategy and adding some label information.
This idea also can be applied to other complex logic learning tasks.
For example, it is difficult to compute decimal bit operation directly,
we can convert the decimal to binary first,
and then compute binary bit operations.
The DCM provides a strategy to solve
some difficult logic reasoning tasks through
combing the domain expert knowledge with data-driven model.

This work is only a preliminary exploration towards learning logic from data.
Several issues are worthwhile investigating along this direction,
such as mining visual functional relations among multiple variables
and directly learning rules from data.
These issues are very challenging and meaningful.
To this end, more logic reasoning data sets containing complex formulas embedded in the images
and more effective models for solving logical reasoning tasks should be specially designed.

\section*{Acknowledgments}

This work was supported by National Key R\&D Program of China (No. 2018YFB1004300), National Natural Science Fund of China (No. 61672332, 61432011, 61502289),
Key R\&D program (International Science and Technology Cooperation Project) of Shanxi Province, China (No. 201903D421003),
Program for the Young San Jin Scholars of Shanxi (No. 2016769), Young Scientists Fund of the National Natural Science Foundation of China (No. 61802238, 61906115, 61603228, 62006146, 61906114), Shanxi Province Science Foundation for Youths (No. 201901D211169, 201901D211170, 201901D211171), Research Project Supported by Shanxi Scholarship Council
of China (No. HGKY2019001), and Scientific and Technologial Innovation Programs of Higher Education Institutions in
Shanxi (No. 2020L0036).

\section*{References}
\bibliography{LiLi}

\begin{thebibliography}{10}
\expandafter\ifx\csname url\endcsname\relax
  \def\url#1{\texttt{#1}}\fi
\expandafter\ifx\csname urlprefix\endcsname\relax\def\urlprefix{URL }\fi
\expandafter\ifx\csname href\endcsname\relax
  \def\href#1#2{#2} \def\path#1{#1}\fi

\bibitem{colom2010human}
R.~Colom, S.~Karama, R.~E. Jung, R.~J. Haier, Human intelligence and brain
  networks, Dialogues in clinical neuroscience 12~(4) (2010) 489.

\bibitem{Johnson2016CLEVR}
J.~Johnson, B.~Hariharan, L.~V.~D. Maaten, F.~F. Li, C.~L. Zitnick,
  R.~Girshick, {CLEVR}: A diagnostic dataset for compositional language and
  elementary visual reasoning, in: IEEE Conference on Computer Vision and
  Pattern Recognition, Honolulu, USA, 2017, pp. 1988--1997.

\bibitem{Wang1997Fuzzy}
G.~Wang, Fuzzy reasoning and fuzzy logic, in: Soft Computing in Intelligent
  Systems and Information Processing. Proceedings of the 1996 Asian Fuzzy
  Systems Symposium, Kenting, China, 1996, pp. 478--483.

\bibitem{Mizumoto1982Comparison}
M.~Mizumoto, Comparison of fuzzy reasoning methods, Fuzzy Sets and Systems
  8~(3) (1982) 253--283.

\bibitem{Yen1999Fuzzy}
J.~{Yen}, Fuzzy logic-a modern perspective, IEEE Transactions on Knowledge and
  Data Engineering 11~(1) (1999) 153--165.

\bibitem{Pei2004On}
D.~W. Pei, On the strict logic foundation of fuzzy reasoning, Soft Computing
  8~(8) (2004) 539--545.

\bibitem{wille1982restructuring}
R.~Wille, Restructuring lattice theory: an approach based on hierarchies of
  concepts, in: I.~Rival (Ed.), Ordered sets, Springer, 1982, pp. 445--470.

\bibitem{Tadrat2012A}
J.~Tadrat, V.~Boonjing, P.~Pattaraintakorn, A new similarity measure in formal
  concept analysis for case-based reasoning, Expert Systems with Applications
  39~(1) (2012) 967--972.

\bibitem{Golinskapilarek2007Relational}
J.~Golinskapilarek, E.~Orlowska, Relational reasoning in formal concept
  analysis, in: IEEE International Fuzzy Systems Conference, London, UK, 2007.

\bibitem{Shao2020The}
M.~W. Shao, M.~M. Lv, K.~W. Li, C.~Z. Wang, The construction of attribute
  (object)-oriented multi-granularity concept lattices, International Journal
  of Machine Learning and Cybernetics 11~(4) (2020) 1017--1032.

\bibitem{nilsson1986probabilistic}
N.~J. Nilsson, Probabilistic logic, Artificial Intelligence 28~(1) (1986)
  71--87.

\bibitem{nilsson1993probabilistic}
N.~J. Nilsson, Probabilistic logic revisited, Artificial Intelligence 59~(1-2)
  (1993) 39--42.

\bibitem{She2018A}
Y.~She, X.~He, Y.~Qian, W.~Xu, J.~Li, A quantitative approach to reasoning
  about incomplete knowledge, Information Sciences 451-452 (2018) 100--111.

\bibitem{Li2020A}
S.~Y. Li, L.~M. Tam, H.~K. Chen, C.~S. Chen, A novel-designed fuzzy logic
  control structure for control of distinct chaotic systems, International
  Journal of Machine Learning and Cybernetics~(11) (2020) 2391--2406.

\bibitem{pearl1987evidential}
J.~Pearl, Evidential reasoning using stochastic simulation of causal models,
  Artificial Intelligence 32~(2) (1987) 245--257.

\bibitem{chen2016fuzzy}
S.-M. Chen, S.-H. Cheng, C.-H. Chiou, Fuzzy multiattribute group decision
  making based on intuitionistic fuzzy sets and evidential reasoning
  methodology, Information Fusion 27 (2016) 215--227.

\bibitem{yang2008fuzzy}
Z.~Yang, S.~Bonsall, J.~Wang, Fuzzy rule-based bayesian reasoning approach for
  prioritization of failures in fmea, IEEE Transactions on Reliability 57~(3)
  (2008) 517--528.

\bibitem{tenenbaum2006theory}
J.~B. Tenenbaum, T.~L. Griffiths, C.~Kemp, Theory-based bayesian models of
  inductive learning and reasoning, Trends in Cognitive Sciences 10~(7) (2006)
  309--318.

\bibitem{Qian2018Local}
Y.~Qian, X.~Liang, W.~Qi, J.~Liang, L.~Bing, A.~Skowron, Y.~Yao, J.~Ma,
  C.~Dang, Local rough set: A solution to rough data analysis in big data,
  International Journal of Approximate Reasoning 97 (2018) 38--63.

\bibitem{sheroughlogic}
Y.~She, X.~He, H.~Shi, Y.~Qian, A multiple-valued logic approach for
  multigranulation rough set model, International Journal of Approximate
  Reasoning 82 (2017) 270--284.

\bibitem{Lin2019Granular}
Y.~Lin, J.~Li, A.~Tan, J.~Zhang, Granular matrix-based knowledge reductions of
  formal fuzzy contexts, International Journal of Machine Learning and
  Cybernetics~(11) (2020) 643--656.

\bibitem{Li2019Double}
M.~Li, M.~Chen, W.~Xu, Double-quantitative multigranulation decision-theoretic
  rough fuzzy set model, International Journal of Machine Learning and
  Cybernetics 10~(5) (2019) 3225--3244.

\bibitem{Guo2019Mining}
Q.~Guo, Y.~Qian, X.~Liang, Mining logic patterns from visual data, in:
  International Conference on Data Mining Workshops, Beijing, China, 2019.

\bibitem{Dai2019Bridging}
W.~Z. Dai, Q.~Xu, Y.~Yu, Z.~H. Zhou, Bridging machine learning and logical
  reasoning by abductive learning, in: Advances in Neural Information
  Processing Systems, Vancouver, Canada, 2019.

\bibitem{huang2016densely}
G.~Huang, Z.~Liu, L.~Van Der~Maaten, K.~Q. Weinberger, Densely connected
  convolutional networks, in: IEEE Conference on Computer Vision and Pattern
  Recognition, Honolulu, USA, 2017, pp. 4700--4708.

\bibitem{He2016Deep}
K.~He, X.~Zhang, S.~Ren, J.~Sun, Deep residual learning for image recognition,
  in: IEEE Conference on Computer Vision and Pattern Recognition, Las Vegas,
  USA, 2016, pp. 770--778.

\bibitem{Liang2021Evolutionary}
X.~{Liang}, Q.~{Guo}, Y.~{Qian}, W.~{Ding}, Q.~{Zhang}, Evolutionary deep
  fusion method and its application in chemical structure recognition, IEEE
  Transactions on Evolutionary Computation (2021) 1--1\href
  {http://dx.doi.org/10.1109/TEVC.2021.3064943}
  {\path{doi:10.1109/TEVC.2021.3064943}}.

\bibitem{Ren2015Faster}
S.~Ren, K.~He, R.~Girshick, J.~Sun, Faster r-cnn: Towards real-time object
  detection with region proposal networks, IEEE Transactions on Pattern
  Analysis and Machine Intelligence 39~(6) (2017) 1137--1149.

\bibitem{He2017Mask}
K.~He, G.~Gkioxari, P.~Doll{\'a}r, R.~Girshick, Mask r-cnn, in: IEEE
  International Conference on Computer Vision, Venice, Italy, 2017, pp.
  2961--2969.

\bibitem{Shelhamer2014Fully}
E.~Shelhamer, J.~Long, T.~Darrell, Fully convolutional networks for semantic
  segmentation, IEEE Transactions on Pattern Analysis and Machine Intelligence
  39~(4) (2017) 640--651.

\bibitem{Chen2016DeepLab}
L.-C. Chen, G.~Papandreou, I.~Kokkinos, K.~Murphy, A.~L. Yuille, Deeplab:
  Semantic image segmentation with deep convolutional nets, atrous convolution,
  and fully connected crfs, IEEE Transactions on Pattern Analysis and Machine
  Intelligence 40~(4) (2018) 834--848.

\bibitem{Vinyals2016Show}
O.~Vinyals, A.~Toshev, S.~Bengio, D.~Erhan, Show and tell: Lessons learned from
  the 2015 mscoco image captioning challenge, IEEE Transactions on Pattern
  Analysis and Machine Intelligence 39~(4) (2016) 652--663.

\bibitem{Johnson2015DenseCap}
J.~Johnson, A.~Karpathy, L.~Fei-Fei, Densecap: Fully convolutional localization
  networks for dense captioning, in: IEEE Conference on Computer Vision and
  Pattern Recognition, Las Vegas, USA, 2016, pp. 4565--4574.

\bibitem{Yang2016Stacked}
Z.~Yang, X.~He, J.~Gao, L.~Deng, A.~Smola, Stacked attention networks for image
  question answering, in: IEEE Conference on Computer Vision and Pattern
  Recognition, Las Vegas, USA, 2016, pp. 21--29.

\bibitem{Wu2017Image}
Q.~Wu, C.~Shen, P.~Wang, A.~Dick, A.~van~den Hengel, Image captioning and
  visual question answering based on attributes and external knowledge, IEEE
  Transactions on Pattern Analysis and Machine Intelligence 40~(6) (2018)
  1367--1381.

\bibitem{Goodfellow2014Generative}
I.~Goodfellow, J.~Pouget-Abadie, M.~Mirza, B.~Xu, D.~Warde-Farley, S.~Ozair,
  A.~Courville, Y.~Bengio, Generative adversarial nets, in: Advances in Neural
  Information Processing Systems, Montr$\acute{e}$al, Canada, 2014, pp.
  2672--2680.

\bibitem{Reed2016Generative}
S.~Reed, Z.~Akata, X.~Yan, L.~Logeswaran, B.~Schiele, H.~Lee, Generative
  adversarial text to image synthesis, in: International Conference on Machine
  Learning, New York City, USA, 2016, pp. 1060--1069.

\bibitem{hu2017learning}
R.~Hu, J.~Andreas, M.~Rohrbach, T.~Darrell, K.~Saenko, Learning to reason:
  End-to-end module networks for visual question answering, in: IEEE
  International Conference on Computer Vision, Venice, Italy, 2017, pp.
  804--813.

\bibitem{Zhang_2016_CVPR}
P.~Zhang, Y.~Goyal, D.~Summers-Stay, D.~Batra, D.~Parikh, Yin and yang:
  Balancing and answering binary visual questions, in: IEEE Conference on
  Computer Vision and Pattern Recognition, Las Vegas, USA, 2016, pp.
  5014--5022.

\bibitem{Hornik1989Multilayer}
K.~Hornik, M.~Stinchcombe, H.~White, Multilayer feedforward networks are
  universal approximators, Neural Networks 2~(5) (1989) 359--366.

\bibitem{Zadeh1973Outline}
Zadeh, L.~A., Outline of a new approach to the analysis of complex systems and
  decision processes, IEEE Transactions on Systems, Man and Cybernetics
  SMC-3~(1) (1973) 28--44.

\bibitem{Antol2015VQA}
S.~Antol, A.~Agrawal, J.~Lu, M.~Mitchell, D.~Parikh, {VQA}: Visual question
  answering, International Journal of Computer Vision 123~(1) (2015) 4--31.

\bibitem{Graves1997Long}
A.~Graves, Long short-term memory, Neural Computation 9~(8) (1997) 1735--1780.

\bibitem{hoshen2016visual}
Y.~Hoshen, S.~Peleg, Visual learning of arithmetic operation, in: Association
  for the Advancement of Artificial Intelligence, Phoenix, USA, 2016, pp.
  3733--3739.

\bibitem{lecun2015deep}
Y.~LeCun, Y.~Bengio, G.~Hinton, Deep learning, Nature 521~(7553) (2015) 436.

\bibitem{hinton2006reducing}
G.~E. Hinton, R.~R. Salakhutdinov, Reducing the dimensionality of data with
  neural networks, Science 313~(5786) (2006) 504--507.

\bibitem{smith2007overview}
R.~Smith, An overview of the tesseract ocr engine, in: Ninth International
  Conference on Document Analysis and Recognition, Vol.~2, Curitiba, Brazil,
  2007, pp. 629--633.

\bibitem{Qian2010MGRS}
Y.~Qian, J.~Liang, Y.~Yao, C.~Dang, Mgrs: A multi-granulation rough set,
  Information Sciences 180~(6) (2010) 949--970.

\bibitem{Ke2014Hybridization}
L.~Ke, Q.~Zhang, R.~Battiti, Hybridization of decomposition and local search
  for multiobjective optimization, IEEE Transactions Cybernetics 44~(10) (2014)
  1808--1820.

\bibitem{NIPS20166285}
J.~Liang, J.~Fadili, G.~Peyr\'{e}, A multi-step inertial forward-backward
  splitting method for non-convex optimization, in: Advances in Neural
  Information Processing Systems, Barcelona, Spain, 2016, pp. 4035--4043.

\bibitem{Qian2010Positive}
Y.~Qian, J.~Liang, W.~Pedrycz, C.~Dang, Positive approximation: An accelerator
  for attribute reduction in rough set theory, Artificial Intelligence 174
  (2010) 597--618.

\bibitem{Tan2019Granulation}
A.~Tan, W.~Z. Wu, S.~Shia, S.~Zhao, Granulation selection and decision making
  with multigranulation rough set over two universes, International journal of
  machine learning and cybernetics 10~(9) (2019) 2501--2513.

\bibitem{Chen2020Edge}
L.~{Chen}, P.~{Huang}, Y.~{Li}, Z.~{Meng}, Edge-dependent efficient grasp
  rectangle search in robotic grasp detection, IEEE/ASME Transactions on
  Mechatronics (2020) 1--1\href {http://dx.doi.org/10.1109/TMECH.2020.3048441}
  {\path{doi:10.1109/TMECH.2020.3048441}}.

\end{thebibliography}

\end{document}